%% file: arxiv.tex
\documentclass[11pt]{article}

\usepackage{tgtermes}
\usepackage{newtxtext}
\usepackage{newtxmath}

\usepackage{bm}
\usepackage{endnotes}
\usepackage{mathtools}

\usepackage[ruled,linesnumbered]{algorithm2e}
\usepackage{tabularx,booktabs,makecell}
\usepackage{amsfonts}
\usepackage{forest}
\usepackage{multirow}
\usepackage{graphicx}
\usepackage{subcaption}
\usepackage{overpic}
\usepackage{tcolorbox}
\usepackage[table]{xcolor}
\usepackage{tikz}
\usetikzlibrary{arrows.meta}
\usetikzlibrary{positioning}

\usepackage{bbm}
\DeclareMathOperator*{\argmax}{arg\,max}
\DeclareMathOperator*{\argmin}{arg\,min}

\usepackage{hyperref}
\usepackage{cleveref}

\usepackage[numbers]{natbib}

\usepackage[margin=1in]{geometry}
\usepackage{setspace}
\definecolor{amaranth}{rgb}{0.9, 0.17, 0.31}
\definecolor{amethyst}{rgb}{0.6, 0.4, 0.8}
\definecolor{candyapplered}{rgb}{1.0, 0.03, 0.0}

\usepackage{pifont}

\newcolumntype{L}{>{\raggedright\arraybackslash}X}

\numberwithin{equation}{section}

\title{
Imitation Learning for Combinatorial Optimisation under Uncertainty
}

\author{
Prakash Gawas\\
Polytechnique Montr\'eal, CIRRELT, GERAD\\
\texttt{prakashgawas@polymtl.ca}
\and
Antoine Legrain\\
Polytechnique Montr\'eal, CIRRELT, GERAD\\
\texttt{antoine.legrain@polymtl.ca}
\and
Louis-Martin Rousseau\\
Polytechnique Montr\'eal, CIRRELT\\
\texttt{louis-martin.rousseau@polymtl.ca}
}

\date{}

\begin{document}

\maketitle


\begin{abstract}

Imitation learning (IL) provides a data-driven framework for approximating policies for large-scale combinatorial optimisation problems formulated as sequential decision problems, where exact solution methods are computationally intractable.

A central but underexplored aspect of IL is the role of the \emph{expert} that generates training demonstrations. Existing studies employ a wide range of expert constructions, yet lack a unifying framework to characterise their modelling assumptions, computational properties, and impact on learning performance.

This paper introduces a systematic taxonomy of experts for IL in combinatorial optimisation under uncertainty, classified along three principal dimensions:(i) treatment of uncertainty, (ii) level of optimality, (iii) interaction mode. Building on this taxonomy, we propose a unified DAgger framework accommodating multiple expert queries, expert aggregation, and flexible interaction strategies.

The framework is evaluated on a dynamic physician-to-patient assignment problem with stochastic arrivals and capacity constraints. Policies learned from two-stage stochastic experts consistently outperform those trained on deterministic or full-information experts. Interactive learning yields meaningful gains primarily when expert demonstrations are generated under full information, while one-shot non-interactive learning already generalises well for stochastic and deterministic experts.

\end{abstract}


\input{Sec1-alternate_introduction}
\input{Sec2-background}
\input{Sec3-ilforcou}
\input{Sec4-taxonomy}

\input{Sec5-methodology}

\input{Sec6-results}
\input{Sec7-conclusion}


\bibliographystyle{plainnat}
\input{references}


\pagebreak
\input{appendix}

\end{document}

%% file: Sec1-alternate_introduction.tex
\section{Introduction}
\label{intro}

Combinatorial optimisation (CO) lies at the core of many real-world decision-making problems, including scheduling, logistics, network design, and resource allocation. These problems are challenging due to the combinatorial explosion in the number of feasible solutions, whereby even relatively small instances can generate an enormous search space. This difficulty is further compounded by complex constraints and the requirement to identify globally optimal solutions rather than merely feasible ones. As problem size increases, naïve or brute-force approaches quickly become impractical, necessitating more sophisticated methods for efficiently exploring the solution space.

The computational burden becomes particularly severe in settings involving uncertainty or temporal dynamics, such as stochastic or sequential decision problems, where multiple future scenarios must be considered. In such cases, exact optimisation methods (such as branch-and-bound or dynamic programming) often become computationally intractable. This challenge is further exacerbated in applications that require fast and reliable decision-making. To address these limitations, practitioners rely on a variety of approximation techniques, including heuristics, metaheuristics, and problem-specific algorithms. These approaches trade off optimality guarantees for computational efficiency, thereby enabling the practical deployment of CO methods in complex operational environments.

Machine learning (ML) has recently emerged as a powerful complementary tool for tackling CO problems, particularly in regimes where traditional exact or heuristic methods scale poorly. In many operational settings, decisions must be made repeatedly and under strict time constraints, making it impractical to solve large optimisation models from scratch at every decision point. In such environments, ML methods can exploit historical data and structural regularities in problem instances to learn approximate decision rules or policies that generalise across scenarios. To illustrate this motivation, consider problems such as assigning incoming patients to physicians, dispatching vehicles to delivery requests, or selecting branching decisions within a mixed-integer solver. Solving a full optimisation model at every decision point can be computationally prohibitive, particularly in real-time applications. Yet, across repeated instances, consistent structural patterns often emerge in how high-quality decisions depend on the system state. Machine learning models can exploit these patterns by learning from historical solutions or expert demonstrations, allowing them to approximate the outcome of the optimisation process efficiently. Once trained, such models can generate near-optimal decisions in milliseconds rather than seconds or minutes, while remaining responsive to evolving system conditions.

Reinforcement learning (RL) \cite{sutton1998reinforcement} has demonstrated strong potential, particularly in learning decision policies through interaction with an environment, utilising reward feedback to improve performance iteratively. Such approaches are well-suited to settings where explicit models are unavailable, incomplete, or prohibitively complex. Alongside RL, Imitation learning (IL) \cite{hussein2017imitation} has gained attention as an attractive alternative, especially in scenarios where defining a suitable reward function is difficult but expert demonstrations are available. Instead of learning through trial and error, IL aims to learn a direct mapping from system states to actions by mimicking expert behaviour. A central challenge in this paradigm is the acquisition of high-quality expert demonstrations, which are often expensive or infeasible to generate due to the computational cost of solving CO problems optimally. This has motivated growing interest in approximate, surrogate, or interactive expert models that can provide informative guidance while remaining computationally tractable. The design and utilisation of such experts remain key open questions at the intersection of ML and combinatorial optimisation.

\subsection*{Contribution}
The idea of imitating expert policies in CO is relatively recent, and existing work employs a wide range of strategies to define or approximate experts. These include using offline optimal solutions, simplified surrogate models, and structured or repeated interactions with problem-specific oracles. We first survey recent literature to illustrate the breadth of approaches to expert construction and utilisation. However, despite this growing body of work, there is limited understanding of what constitutes an expert in IL for CO and how different expert design choices influence learning outcomes.

To address this gap, we introduce a taxonomy of expert design for IL in CO. The proposed taxonomy unifies existing approaches by organising experts along key conceptual dimensions, providing a coherent framework that connects and contextualises a wide range of prior methods. It also serves as a guide for the principled selection and design of experts across different problem settings.

In the second part of the paper, we propose an adaptive two-stage learning pipeline in which a CO-based expert layer generates demonstrations that are then used to train an ML-based decision model. The framework exposes several design choices that correspond directly to the dimensions of the expert taxonomy. We conduct a comparative empirical study using a physician scheduling problem to examine how these choices affect learning performance. The results offer three concrete practical insights, illustrated through a 
controlled case study on a physician-to-patient assignment problem: (i)  two-stage stochastic experts consistently produce stronger learned policies 
than deterministic or full-information experts, as they generate demonstrations 
that better reflect the uncertainty faced at deployment; (ii) interactive 
learning via DAgger yields meaningful gains primarily when expert demonstrations 
suffer from information asymmetry, while one-shot learning already generalises 
well for stochastic and deterministic experts; and (iii) a MIP gap tolerance 
of 2\% identifies a practical sweet spot, achieving solution quality comparable 
to near-optimal expert demonstrations at roughly half the computational cost.

\subsection*{Scope}
Several existing surveys examine the integration of ML and CO \cite{bengio2021machine, kotary2021end, mandi2024decision, sadana2025survey, scavuzzo2024machine}. In contrast, this article focuses specifically on the role and design of the expert in IL. We concentrate on settings involving sequential decision making under uncertainty, where expert queries are costly and real-time decisions are required. Nevertheless, we also consider studies involving static and two-stage stochastic problems for completeness. Experts may take the form of human decision makers, problem-specific heuristics, or exact optimisation solvers. The overarching objective is to develop learning-based models that approximate or replace expert decision making, thereby enabling scalable and timely decisions without substantial loss in solution quality. No assumptions are made regarding specific structural properties of the underlying optimisation problems.

\subsection*{Organisation}
The remainder of the paper is organised as follows. Section~\ref{background} reviews key concepts in ML and CO for sequential decision making. Section~\ref{il} introduces the fundamentals of IL. Section~\ref{exptaxo} presents the proposed taxonomy of expert policies and situates it within the existing literature.  Section~\ref{meth} details the proposed methodology for applying IL to sequential decision making under uncertainty.  Section~\ref{pdes} describes the problem setting considered in this study and presents the experimental results and analysis. We discuss possible research directions in \Cref{ord} and conclude the paper in \Cref{conclusion}.

%% file: Sec2-background.tex
\section{Background}
\label{background}
Many real-world CO problems can be naturally formulated as
sequential decision problems (SDPs) \cite{powell2007approximate}. At each decision
epoch, an agent selects an action based on the current system state and available
information, while accounting for both immediate consequences and future system
evolution. The process is inherently dynamic, as new information is revealed and the system transitions over time. The objective is to design policies that minimise
cumulative cost or maximise cumulative reward over a finite or infinite planning horizon, making SDPs well-suited to problems involving uncertainty, timing, and
adaptive decision-making.

\subsection{Markov Decision Process}

SDPs are commonly modelled using the Markov decision
process (MDP) framework, which provides a formal representation of decision-making
under uncertainty \cite{powell2022reinforcement}. An MDP models the interaction
between controllable actions and stochastic system dynamics over time. In \Cref{tab:mdp_components}, we
summarise the core components of an MDP. Additional details are provided in the appendix.

\begin{table}[ht]
\centering
\caption{Key components of a Markov decision process}
\label{tab:mdp_components}
\renewcommand{\arraystretch}{1.2}
\begin{tabular}{p{4cm} p{10cm}}
\hline
\textbf{Component} & \textbf{Description} \\
\hline
Decision epoch ($k$) &
A time index at which the decision maker observes the system state and selects an action. We focus on finite-horizon problems with final epoch $K< \infty$. \\

State ($x_k$) &
A complete description of the system at epoch $k$ that is sufficient for future decision making. The set of all states defines the state space $\mathcal{X}$. \\

Action ($a_k$) &
The decision taken at epoch $k$, which may be discrete, continuous, or vector-valued, and must satisfy feasibility constraints $a_k \in \mathcal{A}_k$. \\

Exogenous information ($\xi_k$) &
Random information revealed after action selection, representing uncertainty external to the system. \\

Cost/Reward function ($C_k(x_k,a_k)$) &
The immediate cost incurred or reward received when an action is taken in a given state at epoch $k$. \\

Transition function ($X^M(x_k,a_k,\xi_k)$) &
The mapping that governs the evolution of the system from one state to the next, capturing the system dynamics. \\

Policy ($\pi$) &
A mapping from states to admissible actions. Policies may be deterministic or parameterised functions. \\
\hline
\end{tabular}
\end{table}

A particularly important special case arises when $K = 1$. When uncertainty is present, this setting corresponds to a two-stage stochastic programming problem with a single realisation of uncertainty. In the absence of uncertainty, it reduces to a deterministic optimisation problem. Although such problems are static in nature, they often arise repeatedly in practice. For example, a company may solve an independent delivery routing problem each day, with no direct dependence across days. When $K > 1$, the problem becomes an SDP.

\subsection{Machine Learning}

Machine learning (ML) \cite{hastie2009elements, bishop2006pattern} is a subfield of artificial intelligence concerned with
learning patterns from data to make predictions or decisions without
explicit programming. At a high level, the objective is to learn a mapping from
inputs to outputs. Let $X$ denote the input (feature) space and $Y$ the output
space. We consider parameterised models of the form
$f_\theta: X \rightarrow Y$, where $\theta \in \Theta$ denotes the model
parameters. Learning consists of selecting parameters that yield an effective
input–output mapping based on observed data and a chosen performance criterion.
Depending on the type of feedback available, learning problems are commonly
classified into supervised, unsupervised, and reinforcement learning.

\textbf{Supervised learning} considers settings in which the learner is given a
finite collection of labelled examples $\{(x_i,y_i)\}_{i=1}^N$, with
$x_i \in X$ and $y_i \in Y$. The objective is to approximate the underlying
input–output relationship by minimising a loss function
$\ell : Y \times Y \rightarrow \mathbb{R}$ that measures prediction error.
A common example is the mean squared error,
\[
\frac{1}{N} \sum_{i=1}^{N} \left( f_\theta(x_i) - y_i \right)^2.
\]
Since the true data-generating distribution is unknown, optimisation is
performed empirically over the observed samples.

\textbf{Unsupervised learning} addresses scenarios in which only input samples
$\{x_i\}_{i=1}^N$ are available, without labels. The goal is to uncover latent
structure or informative representations of the data. Typical tasks include
clustering, dimensionality reduction, and representation learning. Evaluation
is generally based on internal metrics or qualitative assessment, as ground
truth labels are unavailable.

\textbf{Reinforcement learning} focuses on sequential decision-making problems
in which an agent interacts with an environment over time. At each step, the
agent observes a state, selects an action, and receives scalar feedback in the
form of a reward, together with a new state. Unlike supervised learning, explicit
target outputs are not provided; instead, the agent must infer desirable
behaviour from delayed and possibly sparse rewards. The objective is to learn a
policy that maximises expected cumulative reward, typically formalised using
the MDP framework.

The learning process involves several interconnected components. First, a model class is selected, such as linear predictors, decision trees, kernel methods, or neural networks, each offering different expressive capabilities. Learning proceeds by specifying an appropriate \emph{loss function} and employing an \emph{optimisation algorithm}, such as gradient descent or its variants, to adjust model parameters to minimise the loss on the training data. In addition to model parameters, learning algorithms involve \emph{hyperparameters}, such as learning rates, regularisation strengths, or architectural choices, which govern the learning dynamics and model complexity. To evaluate generalisation performance and mitigate overfitting, data are typically partitioned into a \emph{training set}, a \emph{validation set} for hyperparameter tuning, and a \emph{test set} for final evaluation. \emph{Overfitting} occurs when a model fits the training data well but fails to generalise to unseen instances. We refer the reader to \cite{hastie2009elements} for more information of these elements.

\subsection{Machine Learning for MDPs}

Machine learning provides several paradigms for addressing MDPs in settings
where system dynamics are partially known or entirely unknown. The state of the MDP directly corresponds to the input feature representation used by the ML model, we use the same notation ($x$) to denote the two. Two prominent
approaches are reinforcement learning and approximate dynamic programming
(ADP). RL is typically applied in model-free settings, where transition dynamics
are unknown, and policies are learned directly through interaction with the
environment. Methods such as Q-learning and policy gradient algorithms rely on
sampled trajectories to estimate value functions or policy gradients and to
iteratively improve decision policies.

Approximate dynamic programming, which has its roots in operations research and
control, often assumes access to a simulator or generative model of the system.
ADP mitigates the intractability of exact dynamic programming by approximating
value functions or policies using function approximation techniques, including
linear basis functions, regression trees, and neural networks. These
approximations enable iterative solution methods such as fitted value iteration
and approximate policy iteration. Although RL and ADP differ in historical
origins and terminology, the distinction between the two has become increasingly
blurred, particularly with the adoption of deep learning–based methods.

IL provides an alternative paradigm for solving MDPs,
especially in settings where specifying a reward function is difficult,
misaligned with operational objectives, or undesirable from a modelling perspective. Rather than learning through trial-and-error interaction, IL seeks
to learn a policy directly from expert-generated state–action trajectories.
This expert-driven formulation is particularly appealing in combinatorial
optimisation settings, where high-quality decision policies may be available
through optimisation solvers or heuristics, but reward design and exploration
are costly.

The effectiveness of IL depends critically on the quality, availability, and
structure of the expert demonstrations. While simple approaches such as
behavioural cloning treat imitation as a supervised learning problem, more
advanced methods incorporate expert interaction or infer latent objectives.
The design and use of experts, therefore, play a central role in applying IL to
MDPs arising from complex optimisation problems.

The focus of this work is the application of IL to CO problems under uncertainty, with particular emphasis on the role and design of the expert. Building on the formulation of MDPs and the use of expert demonstrations, the following section provides a detailed overview of IL techniques.

%% file: Sec3-ilforcou.tex
\section{Imitation Learning: Models and Challenges}\label{il}

IL ~\cite{hussein2017imitation} is a specialised paradigm within supervised ML in which an agent learns to perform a task by mimicking behaviour demonstrated by an expert. Unlike classical supervised learning, where data points are typically assumed to be independent and identically distributed, IL commonly arises in sequential decision-making settings. In such problems, observations are temporally correlated and depend on previously executed actions.

The term \emph{\textbf{expert}} refers to the source of demonstrations used by the learning agent and may correspond to a human decision maker, a heuristic, or an exact optimisation algorithm. \emph{\textbf{Demonstrations}} are typically provided as trajectories $\tau = \{x_k, a^*_{k}\}_{i=1}^K$, where $K$ denotes the trajectory length.  An expert dataset is then defined as $\mathcal{D} = \{\tau_j\}_{j=1}^N$, with each trajectory being generated under the expert policy ($\tau_j \sim \pi^*$) where an action is given by $a_k^* = \pi^*(x_k)$. In an ML context, actions are also referred to as \emph{targets} or \emph{labels}. Standard supervised learning methods can be applied to $\mathcal{D}$ to train a model that predicts expert-like actions from observed states. Here, each state–action pair is treated as an independent training example, effectively flattening trajectories, without explicitly uncoupling or destroying the underlying trajectory structure. When $K=1$, the setting translates to traditional supervised learning.

IL provides a data-driven approach for solving complex SDPs with combinatorial action spaces. However, its effectiveness depends critically on access to high-quality expert demonstrations, which are often expensive to obtain when optimisation solvers generate expert actions. In addition, IL must scale to high-dimensional decision spaces and address \emph{covariate shift}, whereby the learnt policy encounters states at deployment that differ from those observed during training, leading to compounding errors. This section introduces two core IL paradigms, discusses these challenges, and motivates the use of approximate or imperfect experts.

\subsection{Learning Paradigms}

We describe two principal learning paradigms within IL: behavioural cloning and inverse RL, followed by a discussion of practical challenges and common mitigation strategies.

\subsubsection{\textbf{Behavioural Cloning} (BC)}

Behavioural cloning \cite{pomerleau1988alvinn} formulates imitation as a supervised learning problem that maps states or observations directly to expert actions. The objective is to learn the parameters $\theta$ of a policy $\pi_\theta$ by solving
\begin{equation}
\label{learneqn}
\theta = \argmin_{\theta} \mathbb{E}_{(x,a^*) \in \mathcal{D}} \big[\mathcal{L}(x, a^*, \pi_\theta)\big].
\end{equation}
Equation~\eqref{learneqn} minimises the one-step deviation between the learner and the expert along demonstrated trajectories. In the context of CO, two behavioural cloning formulations are commonly employed.

\begin{itemize}
    \item \textbf{Parameter Prediction Task} (PPT) -  
    In this approach, an ML model predicts parameters that define a CO problem, which is subsequently solved by an external optimiser. The idea is to predict parameters that produce solutions close to the true ones in an imitation learning setting, where the expert knows the true parameters. The training objective is typically expressed in terms of \emph{regret} (also referred to as task loss in \cite{donti2017task}) when the predicted parameters appear in the objective of the problem. Regret is defined as the difference between the objective value achieved by the predicted decision and that of the expert decision under a realisation of uncertainty $\xi$. Following \cite{sadana2025survey}, the regret loss is given by
    \begin{equation}\label{regretpeqn}
    \mathcal{L}_{R}(x, a^*, \pi_\theta) = C(x, \pi_\theta(x), \xi) - C(x, a^*, \xi),
    \end{equation}
    where  $C$ denotes the cost function. The loss in \eqref{regretpeqn} is non-negative and vanishes only when the predicted decision coincides with the expert action. This paradigm is also referred to as the regret minimisation task (RMT) \cite{sadana2025survey}.
    When predicted parameters appear in the constraints of the problem, regret alone is not sufficient, and one also needs to minimise the likelihood of infeasible outcomes. In such cases, the loss function involves a combination of a penalty for having infeasibility and losing optimality \cite{mandi2025feasibility}.  

    \item \textbf{Action Imitation Task} (AIT) -  
    In contrast, AIT trains a model $\pi_\theta$ to directly output an action that approximates the expert decision, thereby eliminating the need for an optimisation solver at inference time. A common imitation loss is
    \begin{equation}\label{imilosspeqn}
    \mathcal{L}_{I}(x, a^*, \pi_\theta) = \ell(\pi_\theta(x), a^*),
    \end{equation}
    where $a = \pi_\theta(x)$ and $\ell(\cdot,\cdot)$ may be mean squared error or binary cross-entropy. When regret is used as the distance metric, AIT closely aligns with PPT. This approach is also known as learning to optimise (LtO) \cite{kotary2024learning}.
\end{itemize}

\subsubsection*{Comparing PPT and AIT}
Under the PPT paradigm, the ML model predicts parameters of an easier CO problem that is subsequently solved by an external optimiser. By contrast, AIT eliminates this two-stage inference pipeline by training the model to directly output decisions, avoiding solver calls at deployment time. Nevertheless, an optimisation solver remains necessary during data generation to compute expert actions $a^*$. As a result, both paradigms rely on expert optimisation, but differ in when optimisation is used.

\begin{figure}
    \centering
    
    \caption{PPT}
    \label{pptnet}
\scalebox{0.75}{
    \begin{tikzpicture}[node distance=2cm, >=Stealth]
    \centering 
    \tikzstyle{box} = [rectangle, draw=black, rounded corners, minimum height=0.5cm, minimum width=0.5cm, align=center, line width=0.5mm]
     \tikzstyle{solverbox} = [rectangle, draw=black, double, sharp corners, minimum height=1cm, minimum width=0.5cm, align=center, line width=0.5mm]
    \tikzstyle{dottedleft} = [rectangle, draw=candyapplered, dashed, sharp corners,minimum height=1cm, line width=0.8mm,inner sep=0.05cm,inner xsep=5mm]
    \tikzstyle{dottedright} = [rectangle, draw=amethyst, dashed, sharp corners,minimum height=1cm, line width=0.8mm,inner sep=0.5cm]
    
    \tikzstyle{input neuron} = [circle, draw=black, fill=orange!80, minimum size=4mm, inner sep=0pt]
    \tikzstyle{hidden neuron} = [circle, draw=black, fill=cyan!50, minimum size=4mm, inner sep=0pt]
    \tikzstyle{output neuron} = [circle, draw=black, fill=orange!30, minimum size=4mm, inner sep=0pt]
    \tikzstyle{connection} = [-, line width=0.4mm]

    \node[box] (A) {$\boldsymbol{x}$};
    \node[right=1.2cm of A, minimum height=1cm, minimum width=1cm] (B) { 
        \begin{tikzpicture}[x=1cm, y=1cm, scale=0.8]
            \node[input neuron] (I1) at (0,0) {};
         
            \node[hidden neuron] (H1) at (0.8,0.8) {};
            \node[hidden neuron] (H2) at (0.8,0) {};
            \node[hidden neuron] (H3) at (0.8,-0.8) {};

            \node[output neuron] (O1) at (1.6,0.4) {};
            \node[output neuron] (O2) at (1.6,-0.4) {};

            \foreach \j in {1,2,3}
                \draw[connection] (I1) -- (H\j);

            \foreach \i in {1,2,3}
            {
                \draw[connection] (H\i) -- (O1);
                \draw[connection] (H\i) -- (O2);
            }
        \end{tikzpicture}
    };
    
    \node at (B) [xshift=-0.7cm,yshift=0.5cm] {$\varphi_\theta$}; 

    \node[box, right=1.2cm of B] (C) {$\boldsymbol{\hat{\xi}}$};
    \node[solverbox, right=2cm of C] (D) {$\min_{a\in\mathcal{F}} C(x,a,\hat{\xi})$};
    \node[box, right=1.2cm of D] (E) {$\boldsymbol{a^*(\hat{\xi}})$};
    \node[below left=-0.5cm and 0.3cm of D.south west, anchor=north west] (gear) {\includegraphics[width=0.6cm]{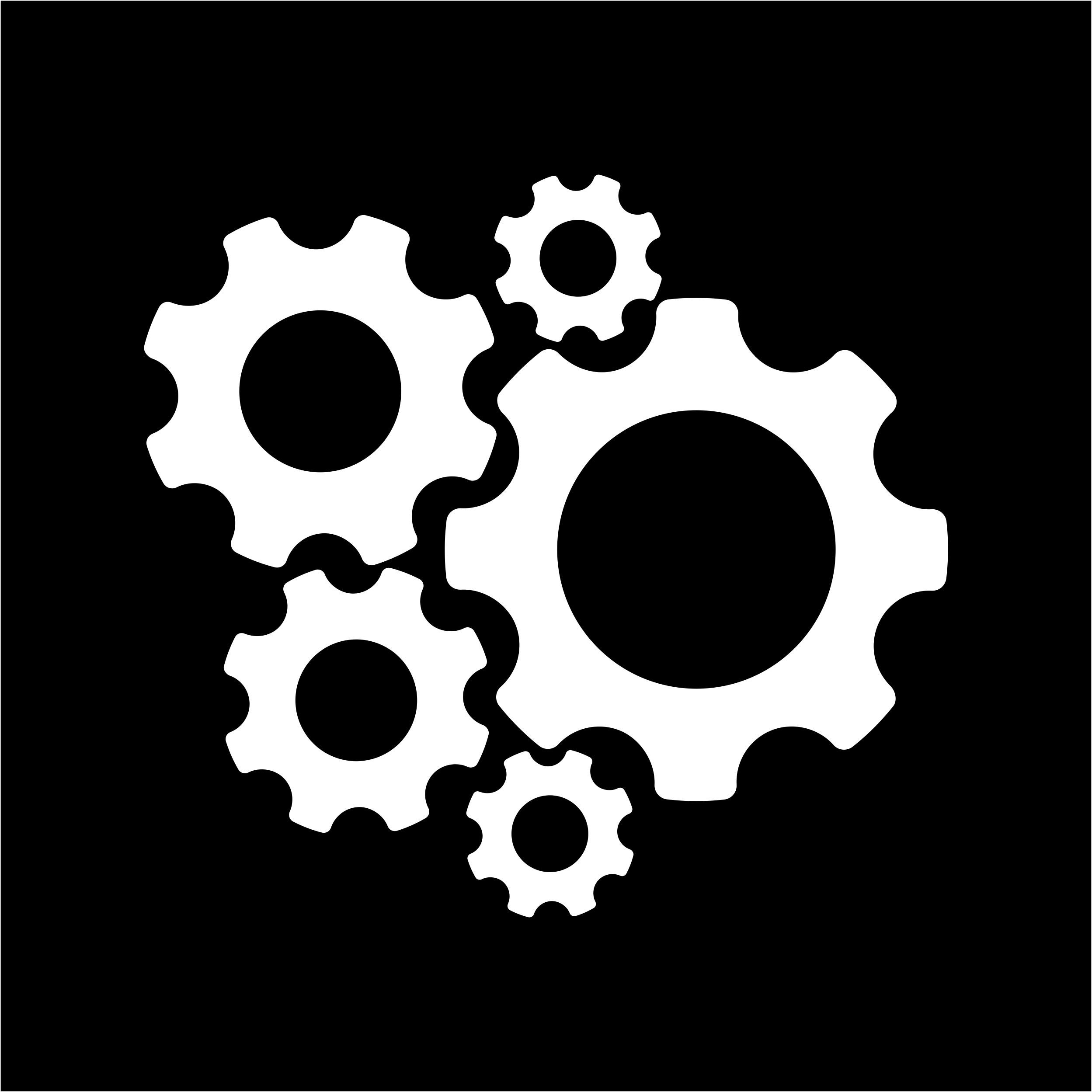}}; 

    \node[dottedleft, fit=(A)(B)(C)] (G) {}; 
    \node[dottedright, fit=(D)(E)] (H) {}; 
 \node[anchor=south] at (G.north) {Prediction Phase};
  \node[anchor=south] at (H.north) {Optimization Phase};
    \draw[<->, line width=0.5mm,  draw=black] (A) -- (B);
    \draw[<->, line width=0.5mm,  draw=black] (B) -- (C);
    \draw[<->, line width=0.5mm,  draw=blue] (C) -- (D);
    \draw[<->, line width=0.5mm,  draw=black] (D) -- (E);

\end{tikzpicture}}

\end{figure}

\begin{figure}
    \centering
        \caption{AIT}
    \label{aitnet}
\scalebox{0.75}{
    \begin{tikzpicture}[node distance=2cm, >=Stealth, xshift=-50cm]
    \centering 
    \tikzstyle{box} = [rectangle, draw=black, rounded corners, minimum height=0.5cm, minimum width=0.5cm, align=center, line width=0.5mm]
\tikzstyle{solverbox} = [rectangle, draw=black, double, sharp corners, minimum height=1cm, minimum width=0.5cm, align=center, line width=0.5mm]
    \tikzstyle{dottedleft} = [rectangle, draw=candyapplered, dashed, sharp corners,minimum height=1cm, line width=0.8mm,inner sep=0.05cm,inner xsep=5mm]
    \tikzstyle{dottedright} = [rectangle, draw=amethyst, dashed, sharp corners,minimum height=1cm, line width=0.8mm,inner sep=0.5cm]
    
    \tikzstyle{input neuron} = [circle, draw=black, fill=orange!80, minimum size=4mm, inner sep=0pt]
    \tikzstyle{hidden neuron} = [circle, draw=black, fill=cyan!50, minimum size=4mm, inner sep=0pt]
    \tikzstyle{output neuron} = [circle, draw=black, fill=orange!30, minimum size=4mm, inner sep=0pt]
    \tikzstyle{connection} = [-, line width=0.4mm]

    \node[box] (A) {$\boldsymbol{x}$};
    \node[solverbox, right=1.2cm of A] (B) {$\min_{a\in\mathcal{F}} C(x,a,\bar{\xi})$};
    \node[right=2cm of B, minimum height=1cm, minimum width=1cm] (C) { 
        \begin{tikzpicture}[x=1cm, y=1cm, scale=0.8]
            \node[input neuron] (I1) at (0,0) {};
         
            \node[hidden neuron] (H1) at (0.8,0.8) {};
            \node[hidden neuron] (H2) at (0.8,0) {};
            \node[hidden neuron] (H3) at (0.8,-0.8) {};

            \node[output neuron] (O1) at (1.6,0.4) {};
            \node[output neuron] (O2) at (1.6,-0.4) {};

            \foreach \j in {1,2,3}
                \draw[connection] (I1) -- (H\j);

            \foreach \i in {1,2,3}
            {
                \draw[connection] (H\i) -- (O1);
                \draw[connection] (H\i) -- (O2);
            }
        \end{tikzpicture}
    };
    
    \node at (C) [xshift=-0.7cm,yshift=0.5cm] {$\pi_\theta$}; 

    \node[solverbox, right=1.2cm of C] (D) {\hspace{0.3cm}$\boldsymbol{\hat{a}}$\hspace{0.3cm}};
    \node[box, right=1.2cm of D] (E) {$\boldsymbol{\hat{a}}$};
    \node[below left=-0.5cm and 0.3cm of B.south west, anchor=north west] (gear) {\includegraphics[width=0.6cm]{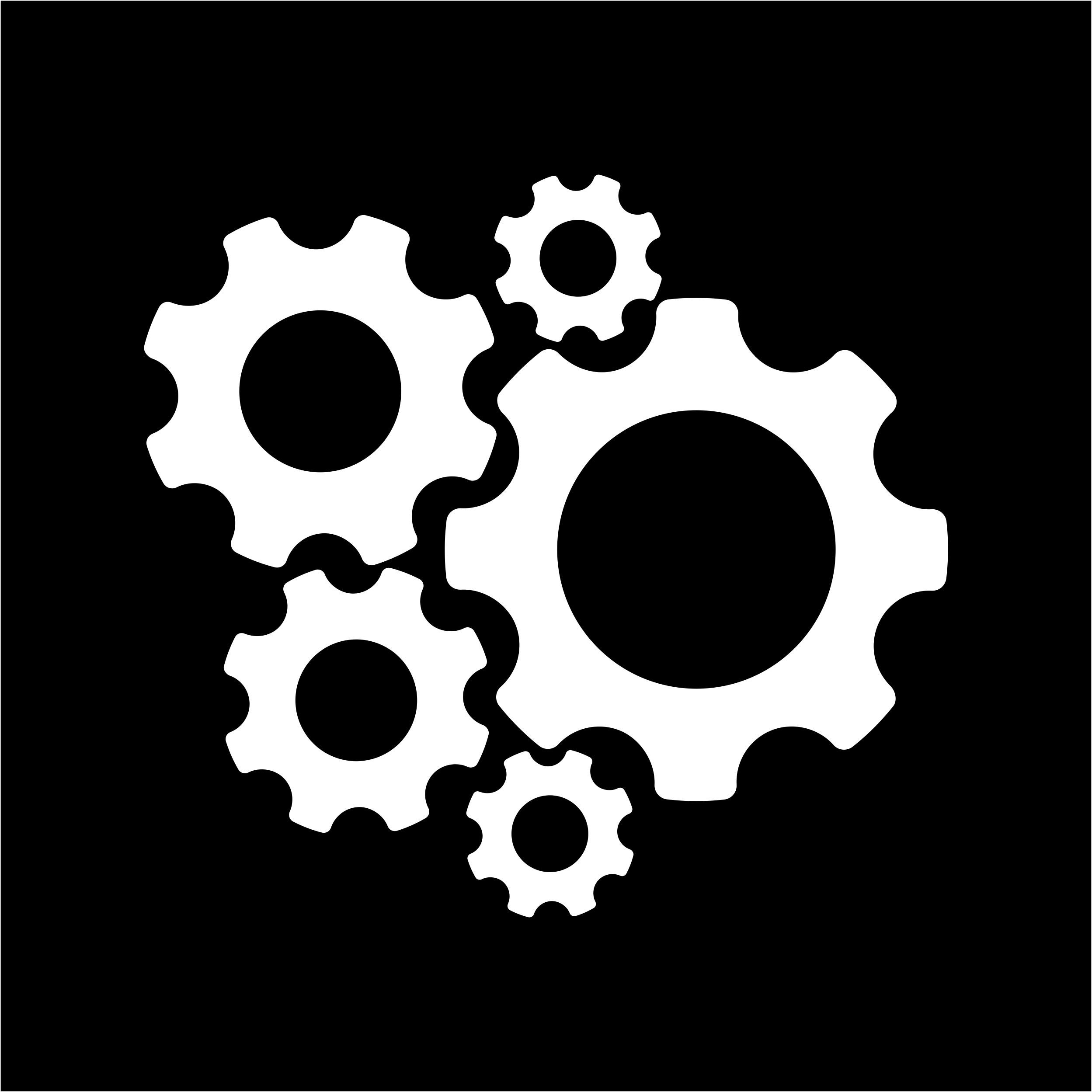}}; 
    \node[below left=-0.5cm and 0.3cm of D.south west, anchor=north west] (gear) {\includegraphics[width=0.6cm]{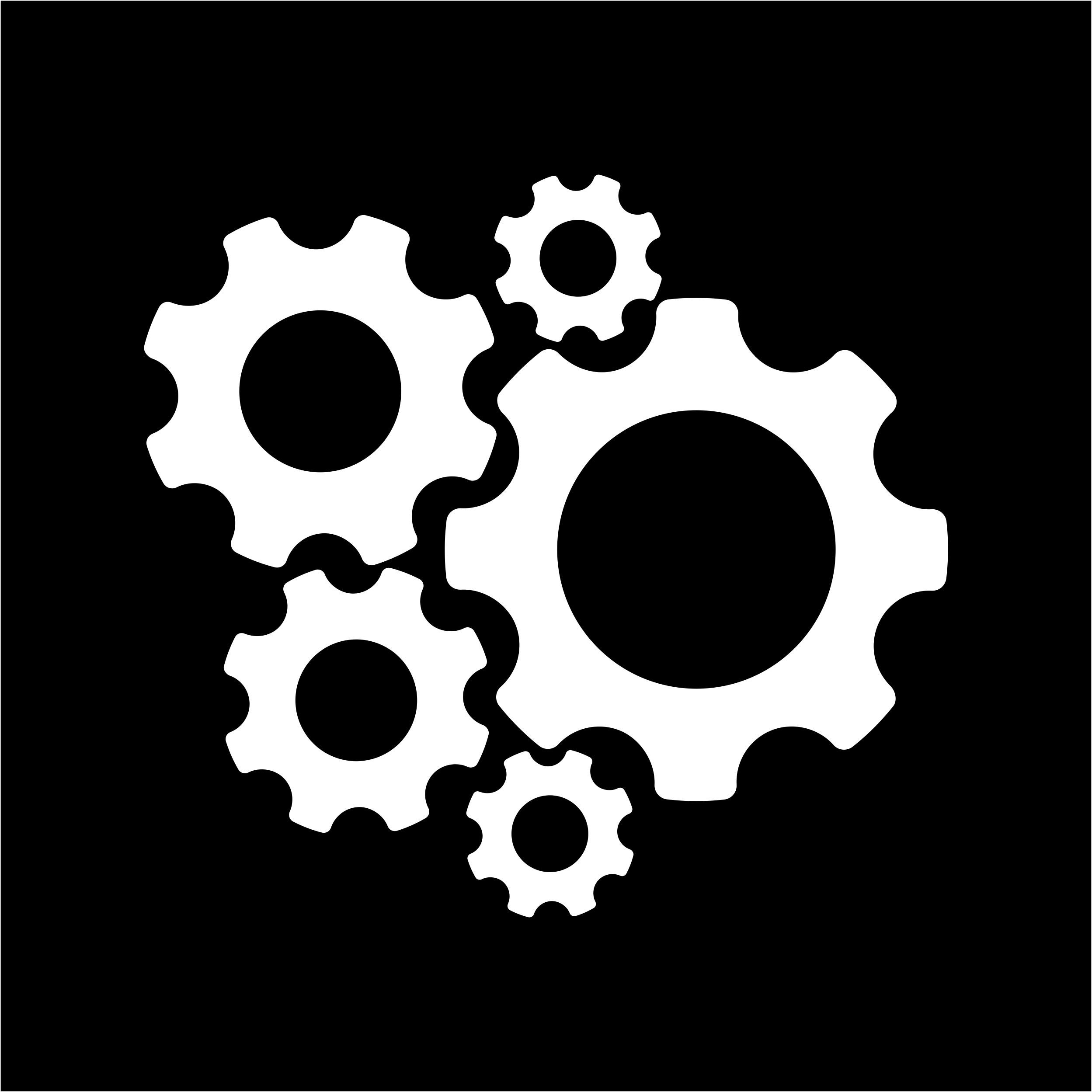}}; 

    \node[dottedright, fit=(A)(B)] (G) {}; 
   \node[dottedleft, fit=(C)(D)(E)] (H) {}; 
 \node[anchor=south] at (G.north) { Optimization Phase};
  \node[anchor=south] at (H.north) {Prediction Phase};
    \draw[<->, line width=0.5mm,  draw=black] (A) -- (B);
    \draw[<->, line width=0.5mm,  draw=blue] (B) -- (C);
    \draw[<->, line width=0.5mm,  draw=black] (C) -- (D);
    \draw[->, line width=0.5mm,  draw=black] (D) -- (E);
\end{tikzpicture}}
\end{figure}

IL with AIT models in CO settings presents several challenges, primarily due to large and complex decision spaces and the presence of ambiguous demonstrations. Without explicitly structured output spaces, standard models may generate infeasible or inconsistent actions. Feasibility can be restored through post-processing techniques such as projection or heuristic correction, although this may result in a modest increase in regret \cite{kotary2024learning}. 

The PPT approach can mitigate infeasibility when the predicted parameters appear only in the objective function. However, when predicted parameters enter the constraints, the resulting solution may violate feasibility with respect to the constraints defined by the true parameters. Similar to AIT, \cite{hu2023predict+} generate an initial solution using predicted parameters and subsequently apply corrective actions, if necessary, at an additional cost.

A key challenge in PPT lies in differentiating the loss function, as it depends on the optimisation outcome. Common strategies to address this include regularisation, perturbation techniques, and the use of surrogate loss functions \cite{mandi2024decision, sadana2025survey}. Both PPT and AIT therefore offer promising approaches for decision making in CO under uncertainty; however, a systematic understanding of their relative strengths and limitations remains limited.

\subsubsection*{Distribution Shift} - 
Given an offline training dataset $\mathcal{D} $ generated by an expert policy $\pi^*$, behavioural cloning seeks to solve \Cref{learneqn}. This formulation gives rise to a distribution shift, or distribution mismatch, a phenomenon well documented \cite{ross2011reduction} in the supervised learning literature for sequential decision-making problems. During deployment, deviations from expert behaviour may lead the learnt policy to encounter states that were absent from the training data. Such deviations can compound over time, causing the policy to drift into unfamiliar regions of the state space where reliable recovery is difficult.

To achieve stronger generalisation, one would instead aim to solve
\begin{equation}
\label{actlearneqn}
   \theta = \argmin_{\theta} \mathbb{E}_{x \sim \pi_\theta} \big[\mathcal{L}(x, a^*, \pi_\theta)\big],
\end{equation}
thereby training the model under the state distribution induced by its own actions. This formulation exposes the learner to states arising from suboptimal decisions and enables exploration of a broader portion of the transition dynamics, improving robustness at deployment.

A standard methodology for addressing this mismatch is dataset aggregation (DAgger) \Cref{alg:dagger}, introduced by \cite{ross2011reduction}. DAgger iteratively generates trajectories using the current learnt policy $\pi_i$ and labels the visited states with actions provided by the expert (steps 4 and 5). The action to be executed at any epoch with an episode ultimately decides if we see new states or not. It is given by a function called the decision rule ($DR(\cdot)$) in step 7. An example of the Vanilla DR is given in \Cref{alg:vandag3}. Vanilla DAgger executes the expert action with a decaying probability and the learner action otherwise, gradually shifting control from the expert to the learnt policy over training iterations. At the end of the episode, a new policy denoted by $\hat{\pi}_{i+1}$ is trained on the updated dataset. By repeatedly collecting expert labels on states visited by the learner, DAgger aligns the training distribution with the state distribution induced at deployment. However, a key limitation in CO settings is the high computational cost of step 5, as expert labelling typically involves repeatedly solving CO problems at each decision epoch. We revisit the DAgger algorithm in \Cref{meth}.

\begin{algorithm}[!ht]
\footnotesize
\SetKwInOut{Input}{Input}
\SetKwInOut{Output}{Output}
\SetKwFunction{DAgger}{DAgger}
\SetKwProg{Fn}{Function}{:}{end}
\caption{DAgger (Dataset Aggregation)}
\label{alg:dagger}

\Fn{\DAgger{}}{
Initial policy $\hat{\pi}_{0}$; 
expert $\pi^{*}$; 
iterations $I$; 
horizon $K$; 
learning routine $\textsc{Train}(\cdot)$;
$\mathcal{D} \leftarrow \emptyset$
\For{$i \leftarrow 0$ \KwTo $I-1$}{
    Observe initial state $x_{0}$\;
    \tcc{Collect a K-step trajectory}
    \For{$k \leftarrow 0$ \KwTo $K-1$}{
        $a^{*}_{k} \leftarrow \pi^{*}(x_{k})$\quad\quad \quad\quad\quad\tcp{Query expert action}
        
        $\mathcal{D} \leftarrow \mathcal{D} \cup \{(x_{k}, a^{*}_{k})\}$\quad \quad\tcp{Append demonstration}
        
        $a_{k} \leftarrow DR(x_{k}, a^{*}_{k}, \hat{\pi}_i)$\quad \quad\tcp{Select action using decision rule}
        
        $x_{k+1} \leftarrow X^M(x_{k}, a_{k}, \xi_{k})$\quad\tcp{Environment transition}
    }

    \tcc{Retrain policy on aggregated dataset}
    $\hat{\pi}_{i+1} \leftarrow \textsc{Train}(\mathcal{D})$\;
}

}
\end{algorithm}

\begin{algorithm}
\footnotesize
   \SetKwFunction{Function}{VanillaDAgger}
   \SetKwProg{Fn}{Function}{:}{end}
\caption{VANILLA DAgger Decision Rule}
\label{alg:vandag3}
\Fn{\Function{$x_k,{a}_{k}, \hat{\pi}_i $}}{  

$\beta_{i} \gets \lambda^{i-1} \beta_0$\;
$z \sim \text{Uniform}(0, 1)$\;

\If{$z \leq \beta_i$}{
    \Return ${a}_{k}$\;
}
    \Return $\hat{\pi}_{i}(x_k)$\;
}
\end{algorithm}

\subsubsection{\textbf{Inverse Reinforcement Learning} (IRL)}

IRL, first introduced by \cite{russell1998learning}, aims to recover the underlying reward function that explains expert behaviour, rather than directly learning a mapping from states to actions. Following standard practice in the IRL literature, we adopt reward-based notation, in contrast to the cost-based formulation used for behavioural cloning. Once the reward function is inferred, a policy can be obtained by solving the resulting RL problem.

Formally, IRL seeks a reward function parameterised by $\theta$ such that the optimal policy induced by this reward coincides with the expert policy $\pi^*$:
\begin{equation}
\label{recovereqn}
\pi^* = \argmax_{\pi} \mathbb{E}_{\pi} \Bigg[\sum_{k = 1}^{K} r_\theta(x_k, \pi(x_k))\Bigg].
\end{equation}
Given an expert dataset $\mathcal{D}$, the reward parameters are estimated by solving
\begin{equation}
\label{irleqn}
\theta = \argmax_{\theta} \; \mathcal{J}(\mathcal{D}, r_\theta).
\end{equation}

Because policy optimisation in IRL is typically performed using RL, IRL methods naturally account for long-term consequences of actions and are therefore less sensitive to covariate shift than behavioural cloning. However, this robustness comes at a high computational cost. IRL is generally sample inefficient, requires repeated interaction with the environment, and alternates between reward estimation and policy optimisation. Moreover, IRL is inherently ill-posed, since multiple reward functions may induce the same expert behaviour, making the unique recovery of the true reward function impossible in general.

A variety of IRL methods have been proposed, differing primarily in how the objective function $\mathcal{J}$ is defined and optimised. Notable approaches include maximum margin methods, which separate expert and non-expert trajectories; maximum entropy IRL, which matches expert behaviour while encouraging stochasticity; Bayesian IRL, which infers a distribution over reward functions; and generative adversarial imitation learning (GAIL), which frames imitation as an adversarial learning problem. 

In practice, IRL is most suitable when expert demonstrations are limited and 
accurate modelling of long-term behaviour is essential. A key advantage is that 
once a reward function is learnt, policies can be re-optimised under changing 
system dynamics without additional demonstrations. By contrast, behavioural 
cloning is preferable when expert data are abundant and sufficiently cover the 
relevant state-action space, and when small deviations from expert behaviour 
do not lead to severe performance degradation. As the experts considered in 
this work is algorithmic rather than human, and the reward specification is not 
the primary challenge, IRL falls outside the scope of the proposed taxonomy; 
we refer the reader to \cite{zare2024survey} and \cite{zheng2022imitation} for comprehensive 
treatments.

\subsection{Learning from Noisy or Suboptimal Experts}

Although IL relies on expert demonstrations, fully optimal and noise-free data are rarely available in practice. Human experts may be inconsistent, and computing optimal solutions can be prohibitively expensive. Consequently, most real-world IL applications rely on \emph{imperfect} but readily available demonstrations. In CO, human experts are generally unsuitable due to problem complexity, and optimisation solvers are used instead. However, solving SDPs under uncertainty is computationally demanding \cite{powell2022reinforcement}, motivating the use of static approximations or simplified scenarios to generate demonstrations.

Several studies in classical IL literature adopt this approach by solving static or simplified CO problems offline to obtain training targets \cite{pham2023prediction, baty2024combinatorial}. Although such solutions are suboptimal, they are informative and significantly cheaper to compute. In many cases, these demonstrations are treated as optimal and used directly within classical IL pipelines. Nevertheless, given the $\mathcal{NP}$-hard nature of many CO problems, even static instances often require heuristics, time limits, or prescribed optimality gaps, introducing additional noise into the demonstrations. Despite its prevalence, the impact of such noise on IL performance in CO settings remains poorly understood, and systematic comparative studies are limited.

Recent work in IL has begun to explicitly address learning under noisy supervision. For example, \cite{xu2022discriminator, kim2021demodice} propose offline IL frameworks that leverage supplementary datasets generated by suboptimal policies. \cite{sekhari2024selective} develop an IL algorithm that reduces the number of expert queries while achieving low regret under noisy demonstrations. \cite{wang2023imitation} introduce a two-stage approach that denoises demonstrations using diffusion models before imitation. Additional contributions in this area include \cite{wu2019imitation, sasaki2020behavioral, wang2021learning}, as summarised in \cite{zare2024survey}.

An additional challenge arises when multiple labels are available for the same state, often generated by different imperfect experts. Such situations may lead to conflicting demonstrations—for instance, when different drivers adopt distinct strategies under identical conditions. While this diversity can enrich the learning signal, it necessitates aggregation or filtering mechanisms to avoid degraded policy performance. Early work by \cite{raykar2009supervised, raykar2010learning} addresses this issue in pre-labelled datasets. More recently, \cite{sun2023mega} propose MEGA-DAgger, an interactive framework for learning from multiple imperfect experts that filters unsafe demonstrations and evaluates experts using scenario-specific criteria. \cite{sekhari2024selective} further extend this line of work to settings in which different experts specialise in distinct regions of the state space. However, this line of research on handling imperfect experts has been developed primarily within the field of robotics. Its application to IL for CO problems remains largely unexplored.

\subsection{Applications of IL in SDPs}

Recent applications of ML to SDPs in CO broadly fall into two categories. The first,
\emph{ML-augmented optimisation}, uses learning models to enhance classical
optimisation algorithms. The second, known as \emph{end-to-end learning} (E2EL),
integrates optimisation components directly within learning architectures. This
distinction follows the taxonomy of \cite{kotary2021end}. While both IL and RL are used in these settings, we focus on
work that applies IL to guide decision making within these two paradigms.

\subsubsection{ML-Augmented Optimisation}

ML-augmented optimisation embeds learning models within optimisation algorithms
to guide or accelerate key decisions. Learning may be used to initialise
algorithms, tune parameters, or repeatedly inform decisions during execution,
such as branching, node selection, or column generation
\cite{bengio2021machine}. Existing approaches either predict discrete decisions
directly or estimate surrogate scores that rank candidate actions and guide the
search process.

\begin{enumerate}
   
    \item \textbf{Branch-and-Bound}  -   We refer the reader to \cite{bengio2021machine} and \cite{scavuzzo2024machine} for a basic introduction to Branch-and-Bound (B\&B). Branch-and-Bound (B\&B) relies on two core decisions: selecting a branching variable and choosing a node to expand. Several studies use IL to guide these decisions. Works such as \cite{alvarez2017machine, alvarez2016online, balcan2018learning, ding2020accelerating, gasse2019exact, gupta2022lookback, khalil2016learning, lin2022learning, lu-2020, Nair2020SolvingMI, zarpellon2021parameterizing} apply IL to learn branching policies. In parallel, \cite{he2014learning, khalil2022mip, labassi2022learning, yilmaz2021study} focus on node selection strategies. The DAgger algorithm is employed in \cite{he2014learning} and \cite{Nair2020SolvingMI} to incorporate expert interaction. Additional contributions include learning when to invoke primal heuristics \cite{khalil2017learning} and predicting backdoors to accelerate mixed-integer programming \cite{cai2024multi}.

    \item \textbf{Cutting Planes} -   
    Cut generation is a central technique in mixed-integer programming, aimed at tightening relaxations without excluding feasible integer solutions. Several studies use supervised learning to decide which cuts to add, including \cite{deza2023machine, ding2020accelerating, huang2022learning, paulus2022learning}.

    \item \textbf{Column Generation and Branch-and-Price} -  
    Column generation solves large-scale linear programmes by generating variables on demand. When combined with Branch-and-Bound, this yields the Branch-and-Price framework \cite{barnhart1998branch}. Learning-based methods in this setting focus on selecting promising columns or designing effective branching strategies to improve speed. Representative works include \cite{furian2021machine, morabit2021machine, pereira2022learning, sun2022learning, shen2022enhancing}.

    \item \textbf{Large Neighbourhood Search} -   
    Large Neighbourhood Search (LNS) improves solutions by iteratively destroying and repairing partial solutions. ML has been used to generate high-quality initial solutions and to adaptively select neighbourhoods \cite{hizir2025large, huang2023searching, liu2022learning, mossina2019multi, Nair2020SolvingMI, song2020general, sonnerat2021learning}.

    \item \textbf{Search Space Reduction} -  
    Another line of work aims to directly reduce the search space by eliminating decisions unlikely to appear in an optimal solution. For example, in the travelling salesman problem, ML models can predict irrelevant arcs for pruning. Studies following this approach include \cite{fischetti2019machine, fitzpatrick2021learning, grassia2019learning, lauri2019fine, larocca2024combining, li2025learning, spieckermann2025reduce, sun2021generalization, tahir2021improved, tayebi2022learning, vaclavik2018accelerating}.

    \item \textbf{Constraint Programming} -  
    Constraint Programming (CP) solves combinatorial problems through systematic constraint propagation and search. Several works leverage supervised learning to guide CP solvers by predicting constraint satisfaction or search decisions, including \cite{selsam2018learning, sun2024enhancing, xu2018towards}.

    \item \textbf{Algorithm Configuration} -  
    ML is also used to decide whether specific algorithmic components should be applied to a given instance. Examples include predicting when to apply Dantzig–Wolfe decomposition \cite{kruber2017learning}, when to use scaling techniques \cite{berthold2021learning}, whether to apply cutting planes beyond the root node \cite{berthold2206learning}, and whether to linearise the objective in mixed-integer quadratic programmes \cite{bonami2018learning}. In all cases, algorithm runs are used to generate labelled training data.
\end{enumerate}

Problems in this category may be either \emph{single-stage} or \emph{sequential}. Single-stage settings include tasks such as predicting high-quality initial solutions for warm-starting. In contrast, sequential settings require a series of interdependent decisions, such as branching choices along a B\&B tree. \cite{Nair2020SolvingMI} addresses both perspectives. In addition, some approaches rely on a single prediction to inform multiple downstream decisions. For example, \cite{khalil2022mip} predicts likely variable assignments at the root node of a binary linear program and reuses this information for subsequent node selection.

\subsubsection{End-to-End Learning} -  End-to-end learning refers to a modelling paradigm that tightly integrates ML and optimisation within a single framework. In this setting, the learning model either directly predicts solutions to optimisation problems, corresponding to the AIT, or produces partially specified optimisation models whose parameters are inferred from raw or structured input data, corresponding to the PPT.

\subsubsection*{Parameter Prediction Task} -  Early work in this direction includes \cite{donti2017task} and \cite{elmachtoub2022smart}. In particular, \cite{elmachtoub2022smart} introduce the Smart “Predict, then Optimise” (SPO) framework, which trains ML models to predict uncertain coefficients in linear objective functions. As the original SPO loss is non-convex and difficult to optimise, the authors propose a convex surrogate, SPO+, which enables efficient training. Subsequent studies apply the SPO and SPO+ losses across a range of applications, including last-mile delivery, energy systems, and ship management \cite{chu2023data, demirovic2020dynamic, sang2022electricity, tian2023smart}. 

\cite{donti2017task} extend regret minimisation to stochastic convex optimisation problems and rely on implicit differentiation of the regret objective. More recently, \cite{baty2024combinatorial, greif2024combinatorial, jungel2025learning, rautenstrauss2025optimization} propose learning pipelines enriched with optimisation layers and apply them to dynamic vehicle routing problems. These works typically rely on full-information offline solutions as supervision and aim to minimise regret with respect to these benchmarks. Notably, \cite{greif2024combinatorial} and \cite{rautenstrauss2025optimization} employ DAgger-based data collection strategies to improve robustness. 

Related ideas appear in \cite{parmentier2021learning}, which studies a single-machine scheduling problem using structured prediction \cite{nowozin2011structured} to approximate hard instances via easier surrogate problems whose parameters are learnt. \cite{sinclair2023hindsight} leverage hindsight or full-information solutions to estimate expert values and propose a hindsight learning algorithm, demonstrating its effectiveness across multiple problem domains.

\subsubsection*{Action Imitation Task} - 
An early attempt to directly predict solutions to difficult combinatorial problems, such as the Travelling Salesman Problem (TSP), is due to \cite{vinyals2015pointer}, who propose pointer networks trained in a supervised manner on solved instances. Their approach demonstrates limited generalisation to instances of varying sizes. Building on this idea, \cite{li2018combinatorial} and \cite{joshi2019efficient} develop supervised models for static CO problems.

\cite{kotary2024learning} propose a supervised learning framework for predicting optimal solutions based on observable features correlated with unknown parameters of a parametric optimisation problem. \cite{chen2023two} address the Flexible Job Shop Scheduling Problem using a two-stage supervised framework that imitates a constraint programming solver, first predicting machine assignments and then job start times, followed by post-hoc feasibility restoration. 

Several works extend action imitation to dynamic or stochastic settings. \cite{larsen2018predicting} study the Load Planning Problem by imitating second-stage decisions of a two-stage optimisation model using full-information supervision. \cite{ozarik2024machine} learn driver routing behaviour in last-mile delivery from historical data. \cite{qi2023practical} consider dynamic inventory management under demand uncertainty, training supervised models on offline full-information solutions. \cite{pham2023prediction} apply regression-based learning to dynamic radiotherapy appointment scheduling. More recently, \cite{gawas2024ildagger} study the Employee Notification Timing Problem and use DAgger to train policies that imitate a range of offline expert models.

ML-augmented optimisation and end-to-end learning thus represent two dominant paradigms for combining ML with optimisation. While ML-augmented approaches embed learning within specific algorithmic components such as branching, cutting, or search space reduction, end-to-end methods aim to predict decisions or optimisation parameters directly through unified learning pipelines. Across both paradigms, supervision plays a central role, yet the nature of the expert varies substantially, ranging from full-information solutions and heuristics to intermediate solver states. These choices have significant implications for learning performance, generalisability, and computational cost. Understanding how experts are defined and utilised is, therefore, crucial for algorithm design. We next introduce a taxonomy of expert types to clarify these distinctions and to guide the development of IL methods for SDPs.

%% file: Sec4-taxonomy.tex
\section{Expert Taxonomy}\label{exptaxo}

In IL, an \emph{expert} is a policy, or more generally a
decision-making function, that specifies the desired behaviour in a sequential
decision problem. The expert guides the learning agent by providing action
demonstrations for given states, typically assuming access to an optimal or
near-optimal strategy under a known or implicit objective. Related terms used in
the literature include \textit{oracle}, \textit{demonstrator},
\textit{optimal policy}, \textit{ground truth}, \textit{reference policy}, and
\textit{target}. We broadly distinguish between two categories of experts.

\begin{itemize}
    \item \textbf{Human Expert} –  
    Human experts rely on experience and contextual understanding to make
    decisions under uncertainty. In dynamic settings such as vehicle dispatching
    or healthcare scheduling, they implicitly incorporate real-time information,
    operational constraints, and institutional knowledge. For example,
    \cite{ozarik2024machine} study last-mile delivery by imitating driver
    decisions that capture practical insights not explicitly represented in
    optimisation models. While human decisions can be nuanced, they may also be
    inconsistent or subject to cognitive bias. Moreover, collecting large and
    reliable datasets from human experts is often difficult due to limited
    availability and high cost.

    \item \textbf{Algorithmic Expert} –  
    Algorithmic experts generate decisions using mathematical models,
    optimisation solvers, or rule-based and heuristic methods. Such experts can
    typically produce large volumes of consistent demonstrations. Their design
    depends on how the decision objective is specified and how uncertainty,
    denoted by $\boldsymbol{\xi}$, is handled. The vector $\boldsymbol{\xi}$
    captures exogenous factors such as demand, travel times, or arrival
    processes that are not fully known at the time of decision making.
\end{itemize}

Given the complexity of real-world SDPs and the need for timely decisions, algorithmic experts are often more suitable for generating demonstrations in practice. For many real-world business applications, computing optimal actions, even offline, is often impractical due to problem complexity for an SDP. Consequently, it is common in operations research to design algorithms that produce near-optimal solutions—decisions that are not globally optimal but are sufficiently close to be useful in practice. When such solutions can be computed within reasonable time limits, they can serve as expert demonstrations for IL.

Despite the widespread use of approximate or surrogate experts, the literature lacks a clear and systematic taxonomy for defining experts and their roles within IL. To address this gap, we propose a structured taxonomy of expert types to guide the design and application of IL methods in SDPs. The three principal 
dimensions of the taxonomy --- treatment of uncertainty, level of optimality, 
and interaction mode --- are chosen because they directly govern the quality, 
cost, and distributional properties of the demonstrations seen by the learner, 
which are the factors most consequential for learning performance in sequential 
decision problems under uncertainty.

\subsection{Expert Decision Models under Uncertainty}

We further classify algorithmic experts according to how they handle uncertainty $\boldsymbol{\xi}$ at the time of decision making. The treatment of $\boldsymbol{\xi}$ determines whether an expert behaves myopically or anticipates future outcomes, and whether it is relatively simple or highly adaptive. We outline five classes of algorithmic experts, distinguished by how uncertainty is incorporated into their decision models.

\begin{enumerate}
\item \textbf{Myopic Expert} -  
A myopic expert selects actions based solely on immediate costs, effectively ignoring future uncertainty. It treats $\boldsymbol{\xi}$ as unavailable or irrelevant and bases decisions only on the current state. This leads to simple models and fast solution times, but often results in poor long-term performance due to the lack of anticipation.

\item \textbf{Deterministic Expert} -  
The deterministic expert replaces uncertain parameters $\boldsymbol{\xi}$ with fixed point estimates $\hat{\boldsymbol{\xi}}$, such as mean or nominal values, and solves the resulting deterministic optimisation problem. While straightforward to implement, this approach may perform poorly when actual realisations deviate substantially from the assumed estimates.

\item \textbf{Full-Information (Hindsight) Expert} -   
The full-information expert assumes perfect foresight and solves the optimisation problem using the realised values $\tilde{\boldsymbol{\xi}}$ of all uncertain parameters. Although infeasible in real-world operations, this expert serves as a useful benchmark, as it yields the best possible decision in hindsight for each instance.

\item \textbf{Two-Stage Stochastic Expert} -   
In the two-stage stochastic framework, uncertainty $\boldsymbol{\xi}$ is not observed at the initial decision stage. The expert first selects a here-and-now (first stage) action and subsequently adapts through recourse decisions (second stage) once uncertainty is revealed. The cost function can be divided into two parts, where the first part $Q_1(a)$ represents the first stage cost and $Q_2(a, \xi)$ denotes the second stage cost.  This structure captures limited anticipation by committing to an initial action before uncertainty is known and then optimally adjusting with recourse actions after the uncertainty is revealed. Thus, it produces more robust supervision than previous experts. 

\item \textbf{Multi-Stage Stochastic Expert} -   
Multi-stage stochastic experts operate in fully dynamic settings where uncertainty unfolds gradually over time. Decisions are updated at each stage based on newly revealed information, allowing policies to adapt as $\boldsymbol{\xi}$ evolves. While this formulation closely reflects real-world decision making under uncertainty, it is computationally expensive to solve exactly, particularly when uncertainty is high-dimensional or difficult to model.
\end{enumerate}

Table~\ref{sumexpmod} summarises the objective structure and treatment of uncertainty for each expert type. Here, $C$ denotes the total cost function and $\mathcal{A}$ represents the set of all feasible actions.

\begin{table}[!ht]
\caption{Summary of Expert Models}
\label{sumexpmod}
\centering
\resizebox{14cm}{!}{
\begin{tabular}{|l|l|l|l|}
\hline
\textbf{Policy/Action} & \textbf{Model Type} & \textbf{Objective} & \textbf{Uncertainty} \\ \hline
$\pi^{*m}, a^{*m}$ & Myopic & $\argmin\limits_{a \in \mathcal{A}} \{ C(a) \}$ & No uncertainty \\ \hline
$\pi^{*d}, a^{*d}$ & Deterministic & $\argmin\limits_{a \in \mathcal{A}} \{ C(a, \hat{\boldsymbol{\xi}}) \}$ & Point estimate $\hat{\boldsymbol{\xi}}$ \\ \hline
$\pi^{*f}, a^{*f}$ & Full-Information & $\argmin\limits_{a \in \mathcal{A}} \{ C(a, \tilde{\boldsymbol{\xi}}) \}$ & Realised outcomes $\tilde{\boldsymbol{\xi}}$ \\ \hline
$\pi^{*t}, a^{*t}$ & Two-Stage Stochastic & $\argmin\limits_{a \in \mathcal{A}^1} \{  C(a, \xi) = Q_1(a) + \mathbb{E}_{\xi}[Q_2(a, \xi)] \}$ & Uncertainty Set \\ \hline
$\pi^*, a^*$ & Multi-Stage Stochastic & $\argmin\limits_{a \in \mathcal{A}} \mathbb{E}_{\xi} \{ C(a, \xi) \}$ & Full Uncertainty \\ \hline
\end{tabular}}
\end{table}

Most of the literature in SDP relies on either myopic or full-information experts for imitation, depending on the problem structure and availability of future information. In settings where uncertainty is absent—such as static or fully observable optimisation problems—a deterministic expert is typically employed. Since all relevant parameters are known with certainty, the expert can compute a single optimal solution using standard optimisation techniques. 

\subsection{Expert Optimality}

Even when employing the expert types defined above, the underlying optimisation
problem may remain difficult to solve. Many deterministic optimisation problems
are $\mathcal{NP}$-hard, making exact solution approaches computationally
impractical in many real-world settings. As a result, practitioners often rely
on heuristics or impose stopping criteria when using optimisation solvers. We
therefore classify experts according to the optimality guarantees of the models
they solve.

\begin{itemize}
    \item \textbf{Task-Optimal Expert} –  
    A task-optimal expert solves its underlying optimisation model to
    optimality, assuming the problem is tractable and can be solved exactly
    within reasonable computational limits. Such experts are most applicable
    when efficient solution methods are available. For example, a strong branching expert evaluates linear programming relaxations for all fractional
    variables and selects the branching variable that yields the greatest
    objective improvement \cite{Nair2020SolvingMI, alvarez2016online}. Optimal
    solutions may be obtained using specialised solvers, including
    mixed-integer programming solvers \cite{sun2022learning}, SAT solvers
    \cite{sun2024enhancing}, and dedicated solvers such as Concorde for the TSP
    \cite{sun2021generalization}. We denote the policy of a task-optimal expert
    by $\pi^{*\cdot}$ and the corresponding optimal action by $a^{*\cdot}$,
    where $\cdot$ refers to the expert type in \Cref{sumexpmod}.

    \item \textbf{Approximate Expert} –  
    An approximate expert produces actions that trade off solution quality for
    computational efficiency, aiming to generate near-optimal decisions.
    Common approximation strategies include:
    \begin{itemize}
        \item \textbf{Time limits}: terminating the solver after a fixed time
        budget and returning the best solution found
        \cite{li2025learning, baty2024combinatorial, pham2023prediction, sun2021generalization}.
        \item \textbf{Optimality gaps}: stopping once a predefined optimality
        gap is achieved \cite{larsen2018predicting}.
        \item \textbf{Objective-based refinement}: applying alternative
        formulations or local improvements to progressively improve solution
        quality \cite{ding2020accelerating}.
        \item \textbf{Heuristics}: directly using high-quality heuristic
        methods, such as hybrid genetic search for the prize-collecting TSP
        \cite{baty2024combinatorial}, or evaluating a finite set of candidate
        actions and selecting the best-performing one \cite{song2020general}.
    \end{itemize}
    We denote the policy of an approximate expert by $\pi^{\cdot}$ and the
    corresponding action by $a^{\cdot}$.
\end{itemize}

\subsection{Expert Interaction}

Depending on the cost, availability, and reliability of the expert, different strategies can be used to incorporate expert supervision into IL.

\begin{itemize}
    \item \textbf{Learning without interaction} -  
    In this non-iterative setting, the expert is queried once to label a batch of observed states, after which the learning model is trained offline. This approach is appropriate when expert queries are expensive or time-consuming, as it minimises interaction. Most studies in the literature adopt this strategy by generating a fixed dataset of expert-labelled states. However, as discussed in \Cref{il}, one-shot learning is vulnerable to poor generalisation when the learnt policy encounters states that were not present in the expert dataset.  We also refer to this setting as \emph{Isolated learning}.

    \item \textbf{Learning with interaction} -  
    Interactive expert querying is commonly used to enrich training data and mitigate distribution shift by exposing the learner to a broader range of states. \cite{he2014learning} employ the DAgger algorithm to iteratively query an expert when learning node selection policies for branch-and-bound. Similarly, \cite{song2020general} propose a DAgger-based approach to learn search policies for both $A^*$ and branch-and-bound algorithms in integer programming. Several recent works adopt iterative expert interaction in combinatorial optimisation settings, including \cite{gawas2024ildagger, greif2024combinatorial, Nair2020SolvingMI, rautenstrauss2025optimization}. 

    While effective, the standard DAgger algorithm can be computationally expensive due to frequent expert queries. To reduce this cost, alternative interaction schemes have been proposed \cite{labassi2022learning, rautenstrauss2025optimization}. Related work also considers online learning settings, where models are updated incrementally as new data are collected, which likewise requires expert interaction. Examples include \cite{alvarez2016online, khalil2016learning, gasse2019exact, lu-2020}.
\end{itemize}

\Cref{classlit} summarises the literature on expert models, highlighting distinctions between task-optimal and approximate experts, as well as between one-shot and interactive interaction strategies. The figure indicates that stochastic expert models and interactive learning approaches remain relatively uncommon, likely due to the additional methodological and computational complexity they introduce.

\begin{figure}[!ht]
\caption{Classification of literature.}
\label{classlit}
    \centering
\begin{tikzpicture}[
  box/.style={
    draw,
    rounded corners,
    minimum width=40mm,
    minimum height=6mm,
    font=\scriptsize,
    align=center
  },
   box1/.style={
    draw,
    circle,
    minimum size=6mm,
    font=\scriptsize,
    align=center
  },
     box2/.style={
    draw,
    circle,
    minimum size=6mm,
    font=\scriptsize,
    align=center
  },
    boxb/.style={
    draw,
    rounded corners,
    minimum width=10mm,
    minimum height=5mm,
    font=\scriptsize,
    align=center
  },
  boxr/.style={
    draw,
    minimum width=4mm,
    minimum height=4mm,
    font=\scriptsize,
    align=center
  },
    boxs/.style={
    draw,
    minimum width=4mm,
    minimum height=4mm,
    font=\scriptsize,
    align=center
  },
  >=latex,
  boxA/.style={box1, fill=green!25},
  boxB/.style={boxr, fill=blue!25},
  boxC/.style={boxs, fill=yellow!25},
  boxD/.style={box2, fill=red!25},
]

\node[box] (root) at (-1,0) {\large Literature};

\node[box] (A) at (-5,-1) {\large Myopic};
\node[box] (D) at ( 3,-1) {\large Deterministic};

\node[box] (B) at (-3.5,-7) {\large Hindsight};
\node[box] (E) at ( 1.5,-7) {\large{2-Stage Stochastic}};

\draw[->] (root) -- (A);

\draw[->] (root) -- (D);
\coordinate (split) at (-1,-3.5); 
\draw[->] (root) -- (split);      
\draw[->] (split) |- (B);         
\draw[->] (split) |- (E);         

\node[boxA] (A1) at (-6.2,-2) {};
\node[boxD] (A2) at (-3.8,-2) {};
\draw[->] (A) -- (A1);
\draw[->] (A) -- (A2);

\node[boxA] (B1) at (-4.7,-8) {};
\node[boxD] (B2) at (-2.3,-8) {};
\draw[->] (B) -- (B1);
\draw[->] (B) -- (B2);

\node[boxA] (D1) at ( 1.8,-2) {};
\node[boxD] (D2) at ( 4.2,-2) {};
\draw[->] (D) -- (D1);
\draw[->] (D) -- (D2);

\node[boxA] (E1) at ( 0.3,-8) {};
\node[boxD] (E2) at ( 2.7,-8) {};
\draw[->] (E) -- (E1);
\draw[->] (E) -- (E2);

\node[boxB] (A1a) at (-6.8,-3) {};
\node[boxC] (A1b) at (-5.6,-3) {};
\draw[->] (A1) -- (A1a);
\draw[->] (A1) -- (A1b);

\node[boxB] (A2a) at (-4.4,-3) {};
\node[boxC] (A2b) at (-3.2,-3) {};
\draw[->] (A2) -- (A2a);
\draw[->] (A2) -- (A2b);

\node[boxB] (B1a) at (-5.3,-9) {};
\node[boxC] (B1b) at (-4.1,-9) {};
\draw[->] (B1) -- (B1a);
\draw[->] (B1) -- (B1b);

\node[boxB] (B2a) at (-2.9,-9) {};
\node[boxC] (B2b) at (-1.7,-9) {};
\draw[->] (B2) -- (B2a);
\draw[->] (B2) -- (B2b);

\node[boxB] (D1a) at ( 1.2,-3) {};
\node[boxC] (D1b) at ( 2.4,-3) {};
\draw[->] (D1) -- (D1a);
\draw[->] (D1) -- (D1b);

\node[boxB] (D2a) at ( 3.6,-3) {};
\node[boxC] (D2b) at ( 4.8,-3) {};
\draw[->] (D2) -- (D2a);
\draw[->] (D2) -- (D2b);

\node[boxB] (E1a) at (-0.3,-9) {};
\node[boxC] (E1b) at ( 0.9,-9) {};
\draw[->] (E1) -- (E1a);
\draw[->] (E1) -- (E1b);

\node[boxB] (E2a) at ( 2.1,-9) {};
\node[boxC] (E2b) at ( 3.3,-9) {};
\draw[->] (E2) -- (E2a);
\draw[->] (E2) -- (E2b);

\node[boxb] (A1aL) at (-6.8,-5.5) {\cite{alvarez2017machine},
\cite{furian2021machine},
\cite{gupta2022lookback},\\
\cite{khalil2017learning},
\cite{kraul2023machine},
\cite{morabit2021machine},\\
\cite{Nair2020SolvingMI},
\cite{ojha2023optimization},
\cite{paulus2022learning},\\
\cite{shen2022enhancing},
\cite{sonnerat2021learning},
\cite{sun2022learning}}; \draw[->] (A1a) -- (A1aL);
\node[boxb] (A1bL) at (-5.6,-4) {\cite{gasse2019exact}, \cite{khalil2016learning}, \cite{vaclavik2018accelerating}}; \draw[->] (A1b) -- (A1bL);
\node[boxb] (A2aL) at (-4.4,-5.5) {\cite{guaje2024machine},
\cite{huang2023searching},
\cite{lin2022learning},\\
\cite{paul2025data},
\cite{pereira2022learning},
\cite{zarpellon2021parameterizing}}; \draw[->] (A2a) -- (A2aL);
\node[boxb] (A2bL) at (-3.2,-4) {\cite{lu-2020}}; \draw[->] (A2b) -- (A2bL);

\node[boxb] (B1aL) at (-5.3,-11.5) {\cite{demirovic2020dynamic},
\cite{elmachtoub2022smart},
\cite{julien2024machine},\\
\cite{jungel2025learning},
\cite{kotary2024learning},
\cite{niroumandrad2024learning},\\
\cite{qi2023practical},
\cite{sinclair2023hindsight},
\cite{wang2023imitation}}; \draw[->] (B1a) -- (B1aL);
\node[boxb] (B1bL) at (-4.1,-10) {\cite{he2014learning}, \cite{gawas2024ildagger}, \cite{labassi2022learning},\\ \cite{rautenstrauss2025optimization}}; \draw[->] (B1b) -- (B1bL);
\node[boxb] (B2aL) at (-2.9,-11.5) {\cite{baty2024combinatorial},
\cite{cai2024multi},
\cite{chu2023data},\\
\cite{donti2017task},
\cite{hottung2020deep},
\cite{larsen2018predicting},\\
\cite{li2025learning},
\cite{pham2023prediction},
\cite{sang2022electricity},\\
\cite{tian2023smart},
\cite{yilmaz2021study}}; \draw[->] (B2a) -- (B2aL);
\node[boxb] (B2bL) at (-1.7,-10) {\cite{song2020general}}; \draw[->] (B2b) -- (B2bL);

\node[boxb] (D1aL) at ( 1.2,-5.5) {\cite{detassis2021teaching},
\cite{fitzpatrick2021learning},
\cite{joshi2019efficient},\\
\cite{kaempfer2018learning},
\cite{lauri2019fine},
\cite{li2018combinatorial},\\
\cite{liu2022learning},
\cite{lodi2020learning},
\cite{morabit2022machine},\\
\cite{parmentier2021learning},
\cite{selsam2018learning},
\cite{spieckermann2025reduce},\\
\cite{sun2021generalization}}; \draw[->] (D1a) -- (D1aL);
\node[boxb] (D1bL) at ( 2.4,-4) {\cite{gawas2024ildagger}
}; \draw[->] (D1b) -- (D1bL);
\node[boxb] (D2aL) at ( 3.6,-5.5) {\cite{chen2023two},
\cite{ding2020accelerating},
\cite{gerbaux2025machine},\\
\cite{grassia2019learning},
\cite{han2023gnn},
\cite{hizir2025large},\\
\cite{khalil2022mip},
\cite{larocca2024combining},
\cite{matsuoka2019machine},\\
\cite{mossina2019multi},
\cite{tayebi2022learning},
\cite{vinyals2015pointer},\\
\cite{xavier2021learning},
\cite{xu2018towards},
\cite{yaakoubi2020flight},\\
\cite{yang2023learning}}; \draw[->] (D2a) -- (D2aL);
\node[boxb] (D2bL) at ( 4.8,-4) {\cite{greif2024combinatorial},\cite{Nair2020SolvingMI}}; \draw[->] (D2b) -- (D2bL);

\node[boxb] (E1aL) at ( -0.3,-10) {\cite{abbasi2020predicting}}; \draw[->] (E1a) -- (E1aL);
\node[boxb] (E2bL) at ( 3.3,-10) {\cite{gawas2024ildagger}}; \draw[->] (E2b) -- (E2bL);

\node at (-1,-13) {%
\begin{tikzpicture}[box/.style={draw, rounded corners, minimum width=6mm, minimum height=3mm}]
    \node[boxA] (lA) at (0,0) {};    \node[right=4pt of lA] {\scriptsize Optimal};
    \node[boxD] (lB) at (3,0) {};    \node[right=4pt of lB] {\scriptsize Approximate};
    \node[boxB] (lC) at (6,0) {};    \node[right=4pt of lC] {\scriptsize One-Shot};
    \node[boxC] (lD) at (9,0) {};    \node[right=4pt of lD] {\scriptsize Interactive};
\end{tikzpicture}
};

\end{tikzpicture}
\end{figure}

\subsection{Labelling strategy}
Most existing studies rely on an expert to provide a single label for each state. However, for certain types of experts, it is possible to obtain multiple labels for the same state within one or more interactions.

\begin{itemize}
    \item \textbf{Single label per state} -  
    This is the standard approach, in which the expert is queried once to obtain a single target action for a given state.

    \item \textbf{Multiple labels per state} -  
    Expert actions may vary due to approximation methods or inherent stochasticity in the underlying models. To improve robustness, multiple labels can be collected for the same state and combined using aggregation techniques such as majority voting or averaging. This process reduces variability and mitigates the impact of expert error. Multiple labels can be obtained using two mechanisms.
    \begin{enumerate}
        \item \textbf{Single call} -  
        A single interaction with the expert yields multiple candidate actions for a given state. This occurs, for example, when solving an integer program using a mathematical programming solver that returns multiple feasible solutions. In \cite{khalil2022mip}, CPLEX is allowed to run for 60 minutes to generate a solution pool for each instance, terminating after identifying 1{,}000 solutions with objective values within 10\% of the best solution found.
        
        \item \textbf{Multiple calls} -  
        Alternatively, repeated queries to the expert may be used to collect several actions for the same state. This approach is common when using deterministic experts in dynamic stochastic programming, where the optimal action depends on sampled future scenarios \cite{gawas2024ildagger, greif2024combinatorial}.
    \end{enumerate}
\end{itemize}

\subsection{Number of Experts Used}
Algorithms can also be categorised based on the number of experts involved and how their expertise is integrated during learning. We identify two primary categories:
\begin{itemize}
    \item \textbf{Single Expert}- The conventional setting in IL, where a single, typically optimal expert provides guidance. The learner assumes the expert is both reliable and inexpensive to query, consistently supplying accurate actions for all states. This simplifies learning but may limit adaptability in complex or diverse environments. Most literature uses a single expert to build the dataset.
    
    \item \textbf{Multi-Expert}- In some cases, multiple experts with complementary specialisations are available. Their actions can be aggregated into a composite "meta-expert", combining diverse expertise to enhance robustness and flexibility across different aspects of the task. \cite{song2020general} uses two different experts to get demonstrations. First, they use an expensive expert to gather initial expert demonstrations, which are used to train initial IL policies. Next, a simpler retrospective expert is used to augment the dataset. \cite{rautenstrauss2025optimization} uses a set of non-optimal policies to label states for which the true hindsight expert is expensive. These states generally represent initial decision epochs in the horizon. \cite{sun2023mega} introduces MEGA-DAgger, a variant of DAgger designed for interactive learning with multiple imperfect experts.
\end{itemize}

\subsection{Trajectory Control}

When experts are employed within an iterative learning framework, they can also be used to control trajectories and correct errors made by a previously learnt model. This approach ensures that diverse states are seen that are likely to be encountered during deployment. The following control strategies are commonly used:
\begin{itemize}
    \item \textbf{Total Expert Control} - The expert governs all states visited within a learning trajectory. A typical example is pure supervised learning without iteration, where the expert labels all states once without further interaction. 
    \item \textbf{Diminishing Expert Control} - The learner initially relies heavily on the expert, but gradually reduces this dependence as its performance improves. This approach is exemplified by the basic DAgger algorithm, also referred to as Vanilla DAgger \cite{ross2011reduction}. This type of control is most used in interactive experts in \cite{gawas2024ildagger}, \cite{greif2024combinatorial}, \cite{song2020general}.
    \item \textbf{Conditional Expert Control} - The expert is invoked selectively based on performance criteria, such as the confidence of the learnt model action falling below a threshold, as in SAFE DAgger \cite{zhang2016query}. Alternatively, an external policy may guide trajectories when the true expert is expensive or imperfect.
\end{itemize}

    Moreover, to reduce reliance on costly expert calls, the expert is queried only for critical cases, such as novel or poorly handled states. This selective querying improves sample efficiency and focuses learning on high-impact scenarios. \cite{menda2019ensembledagger} and \cite{hoque2021lazydagger} describe the Ensemble DAgger and Lazy DAgger algorithms, where the expert is selectively queried based on the discrepancy measure between the expert and learnt policy actions. However, such variants of the DAgger algorithms have not been experimented with for SDP in operations research.

Using the proposed expert taxonomy, we introduce and evaluate an Adaptive DAgger-based algorithm that maps each dimension of the taxonomy to an explicit design choice in the learning procedure. This framework enables a structured comparison of expert configurations within an action imitation (AIT) learning setting.

%% file: Sec5-methodology.tex
\section{Algorithm}\label{meth}

In this section, we present a unified adaptive algorithm for learning from experts in sequential decision problems. The proposed algorithm, shown in \Cref{gdagger}, explicitly incorporates the expert taxonomy introduced in \Cref{exptaxo}. A summary of the algorithm parameters is provided in \Cref{listofparam}.

\begin{table}[!ht]
\caption{List of parameters for \Cref{gdagger}}
\label{listofparam}
\centering
\resizebox{0.6\textwidth}{!}{
\begin{tabular}{|c|p{11cm}|}
\hline
\textbf{Parameter} & \textbf{Description} \\ \hline
$\mathcal{D}$ & Dataset containing state–expert action pairs \\ \hline
$H$ & Number of times to simulate current learnt policy $\pi_i$ \\ \hline
$\mathcal{E} = \{\pi^*_1, \pi^*_2, \dots\}$ & Set of available experts \\ \hline
$K$ & Episode trajectory length \\ \hline
$F_j$ & Function that determines the number of expert calls at a given state \\ \hline
$J$ & Number of expert calls in a given state \\ \hline
$F_e$ & Function that selects which expert to query \\ \hline
$\kappa$ & Stopping criterion for the expert\\ \hline
$A$ & Aggregation function for combining multiple expert actions \\ \hline
$DR$ & Decision rule for selecting the executed action \\ \hline
$\texttt{STOP()}$ &  Stopping criterion \\ \hline
\end{tabular}}
\end{table}

The algorithm is based on the DAgger framework, which follows an iterative policy learning paradigm inspired by online learning and assumes access to one or more experts. Let $\mathcal{E}$ denote the set of available experts. At each learning iteration, indexed by $i$, the learner retrains its policy using all states encountered so far, together with the corresponding expert-labelled actions. This iterative process allows the learner to correct earlier mistakes and progressively improve their performance.

The procedure starts from an initial policy $\hat{\pi}_0$, which may be obtained from expert demonstrations or derived from a heuristic. This policy is then executed for $H$ simulated episodes using the decision rule, each consisting of $K$ decision epochs. At each epoch, the learner observes the current state and may query an expert up to $J$ times, as determined by the function $F_j$. Each query returns an expert action, potentially from different experts in $\mathcal{E}$, decided by the expert-selection function $F_e$.

We denote an expert action by $a^{\cdot}_{ijk}$, where $i$ indexes the learning iteration, $j$ the expert call within the epoch, and $k$ the decision epoch. For clarity, we omit the episode index in this notation. This may be task-optimal or approximate depending on the complexity of the underlying optimisation problem. The parameter $\kappa = [\kappa_1, \kappa_2, \dots]$ specifies the list of expert stopping criteria, thereby controlling the quality of the returned action. The resulting state–action pairs are stored in the dataset before the transition to the next state occurs.

After completing $H$ episodes, the dataset is augmented with the newly collected expert-labelled samples, and a new policy $\hat{\pi}_{i+1}$ is trained. This process is repeated until a stopping condition is satisfied, yielding progressively refined policies. The following paragraphs relate the components of the expert taxonomy to \Cref{gdagger}.

\begin{algorithm}[!ht]
\footnotesize
\SetKwInOut{Input}{Input}
\SetKwInOut{Output}{Output}
\SetKwFunction{DAgger}{DAgger}
\SetKwProg{Fn}{Function}{:}{end}
\caption{Adaptive DAgger Algorithm}
\label{gdagger}

\Fn{\DAgger{}}{
Initialize dataset $\mathcal{D} \leftarrow \emptyset$; initial policy $\hat{\pi}_0$;  iteration counter $i \leftarrow 0$\; learning routine $\textsc{Train}(\cdot)$;

\While{\texttt{STOP()}}{\tcc{Simulate $H$ episodes using current policy $\pi_i$}
    \For{$h \in \{1,\dots,H\}$}{
        \tcc{Sample a $K$-step trajectory}
        \For{$k \in \{1,\dots,K\}$}{
            Observe state $x_{k}$\; 
            $J \leftarrow F_j(x_{k})$\;  
            \tcc{Make $J$ expert calls}
            \For{$j \in \{1,\dots,J\}$}{
                $\pi^* \leftarrow F_e(x_{k}, \mathcal{E}, j)$\;
                ${a}^{\cdot}_{ijk} \leftarrow \pi^*(x_{k}, \kappa)$\;
            }
            
            $\hat{a}_{ik} \leftarrow 
            A\big(\{{a}^{\cdot}_{ijk}\}_{j=1}^J\big)$\;
             $\mathcal{D} \leftarrow \mathcal{D} \cup \{(x_{k}, \hat{a}_{ik})\}$\;
           
            $a_{k} \leftarrow 
            DR(x_{k}, \hat{a}_{ik}, \hat{\pi}_i)$\;
            
            $x_{k+1} \leftarrow 
            X^M(x_{k}, a_{k}, \xi_{k})$\;
        }
    }
    
    $i \leftarrow i + 1$\;
    $\hat{\pi}_{i+1} \leftarrow \textsc{Train}(\mathcal{D})$\;
}
}
\end{algorithm}

We characterize the imitation learning framework with respect to the taxonomy: the \emph{expert type} is given by the set \( \mathcal{E} \), which may include myopic, deterministic, full-information, or two-stage stochastic models; \emph{expert optimality}, governed by a stopping criterion \( \kappa \) that controls solution accuracy and computational effort; and \emph{expert interaction}, determined by the number of iterations \( I \), yielding a sequence of policies \( \{\hat{\pi}_1,\dots,\hat{\pi}_I\} \). Additional design choices include the \emph{labelling strategy} \( F_j \), which specifies when and how often expert queries are issued and how multiple actions \( a^*_{ijk} \) are aggregated via \( A \); the \emph{number of experts} \( |\mathcal{E}| \), with expert selection handled by \( F_e \); and \emph{trajectory control}, defined by a decision rule $DR$. In the experiments presented in \Cref{pdes}, we evaluate the treatment of 
uncertainty, level of optimality, and interaction mode dimensions explicitly;  the number of experts and conditional trajectory control dimensions are 
instantiated in \Cref{gdagger} but remain empirically unexplored and represent 
natural directions for future work.

%% file: Sec6-results.tex
\section{Application to Generalised Assignment Problem}\label{pdes}

This section presents the problem setting and experimental framework used to evaluate the proposed imitation learning methodology. We first introduce the dynamic physician-to-patient assignment problem, which serves as a representative sequential decision problem under uncertainty and allows for systematic instantiation of the expert taxonomy. We then describe the experimental design, including expert configurations, learning setups, and baseline policies, and report a comprehensive empirical evaluation of the proposed framework. The experiments are designed to assess how different expert types, levels of optimality, and interaction strategies influence learning performance, data efficiency, and computational cost.

\subsection{ Generalised Assignment Problem}

The classical Generalised Assignment Problem (GAP) has been extensively studied since the seminal work of \cite{balachandran1976integer}. The problem consists of assigning a set of items (or jobs) to a set of agents (or machines) such that each job is assigned to exactly one agent, agent capacity constraints are respected, and the total assignment cost or profit is optimised. GAP has found applications in a wide range of domains, including job scheduling \cite{balachandran1976integer}, routing \cite{fischetti2019machine}, and patient admission scheduling \cite{ceschia2011local}. A standard integer programming formulation of the GAP is given by \eqref{obj:GAP}-\eqref{cons:binary}. Here, $c_{ij} \in \mathbb{R}_+$ denotes the cost incurred by assigning job $j$ to agent $i$, $d_{ij} \in \mathbb{R}_+$ is the resource consumption of agent $i$ when processing job $j$, $L_i \in \mathbb{R}_+$ is the capacity of agent $i$, and $x_{ij}$ is a binary decision variable equal to $1$ if job $j$ is assigned to agent $i$ and $0$ otherwise.
\begin{align}
    \min \quad & \sum_{i=1}^{m} \sum_{j=1}^{n} c_{ij} x_{ij} \label{obj:GAP} \\[2pt]
    \text{s.t.} \quad 
    & \sum_{i=1}^{m} x_{ij} = 1, 
    \quad && \forall j = 1,\dots,n, \label{cons:assign} \\[2pt]
    & \sum_{j=1}^{n} d_{ij} x_{ij} \le L_i, 
    \quad && \forall i = 1,\dots,m, \label{cons:capacity} \\[2pt]
    & x_{ij} \in \{0,1\}, 
    \quad && \forall i = 1,\dots,m,\ \forall j = 1,\dots,n. \label{cons:binary}
\end{align}

\subsubsection{Physician-to-Patient Assignment}
We consider a dynamic physician-to-patient assignment (PPA) problem as a specific application of the GAP. In a clinical setting, appointment requests arrive sequentially over a booking horizon until a fixed cut-off time. The clinic is staffed by a set of physicians providing services during a given session. Let $\mathcal{P} = \{1, \dots, P\}$ denote the set of physicians. Each physician $p \in \mathcal{P}$ has a regular session duration $T$ (in minutes) and a maximum appointment capacity $L_p$.

Appointment requests arrive independently over time and are handled individually by a receptionist. At each arrival, the receptionist must decide whether to accept or reject the request based on the current state of the system. If the request is accepted, the receptionist must assign the patient to a physician. Each patient $k$ has a preferred physician $p_k$ among a set of eligible physicians $\mathcal{P}_k \subseteq \mathcal{P}$. Each patient is characterised by a service duration $d_k \in \mathbb{R}_+$, which is estimated at the time of the call, and a priority class $r_k \in \{1,2\}$. The arrival process is such that calls from priority 1 patients arrive near the end of the horizon.

Rejecting a patient of priority $r_k$ incurs a penalty $c^{\text{rej}}_{r_k}$, with $c^{\text{rej}}_{1} > c^{\text{rej}}_{2}$. In addition, assigning a patient to an undesired but eligible physician incurs a preference penalty $c^{\text{pref}}_{r_k}$. The primary source of uncertainty in the system arises from the random arrival times and sequences of patients within a session. This creates a trade-off between accepting lower-priority patients early and preserving capacity for potential high-priority patients who may arrive later. Furthermore, assigning patients to undesired physicians may reduce future capacity, potentially forcing rejections later in the booking horizon. The resulting problem is an SDP in which each decision epoch corresponds to a patient arrival, and the action is to assign the patient to an eligible physician or to reject the request. The offline mixed-integer programming formulation, denoted by $(\textbf{\textsc{MIP}}_{\textbf{PPA}})$, is provided in the appendix.

The classical GAP, while a natural starting point, is a simple setting in which heuristics are expected to perform well. The PPA problem instead presents a more challenging and representative instance of sequential decision making under uncertainty, making it better suited for systematically analysing expert design choices within imitation learning. The core difficulty of the PPA lies in the need to anticipate future patient arrivals and strategically reserve capacity for high-priority cases — a challenge that a greedy heuristic cannot adequately address. This structure naturally separates myopic, deterministic, and stochastic experts based on how each incorporates future information into its decisions. At the same time, the underlying combinatorial optimisation formulations remain sufficiently tractable to keep repeated expert queries during DAgger computationally feasible. Feasibility is straightforward to enforce through capacity constraints, enabling clean policy evaluation without extensive post-processing. The PPA further demands rapid decisions at each patient call-in, making it an apt testbed for assessing whether learnt policies can replace expensive online optimisation with fast, deployment-ready inference.

We stress that our aim is not to introduce a new solution method for the PPA itself, but rather to use it as a representative sequential decision problem for systematically studying how expert design choices affect imitation learning outcomes.

\subsubsection{MDP Formulation}\label{mdp}

We now formulate the PPA problem as an MDP.

\textbf{Decision Epochs} –  
Let $\mathcal{K} = \{0,1,\dots, K\}$ denote the set of decision epochs, where each epoch corresponds to the arrival of a patient call-in. Here, $K$ denotes the total number of requests within a session.

\textbf{State Variable} –  
At decision epoch $k$, the system state is given by $x_k = \big(x^k_{\text{patient}}, x^k_{\text{physician}}, x^k_{\text{eligibility}}\big)$. The component $x^k_{\text{patient}} = [d_k, r_k]$ represents the characteristics of the current patient. The component $ x^k_{\text{physician}} = [x^k_p]_{p \in \mathcal{P}}$ captures the status of all physicians, where $x^k_p = \big({1}_{p=p_k}, x^k_{pl}, x^k_{p1}, x^k_{pw}\big).$ Here, ${1}_{p=p_k}$ indicates whether physician $p$ is the patient preferred physician, $x^k_l$ denotes the remaining assignment capacity, $x^k_{p1}$ is the number of priority-1 patients already assigned, and $x^k_{pw}$ is the current workload. A list of indicators gives the eligibility vector functions $x^k_{\text{eligibility}} = [{1}_{p \in \mathcal{P}_k}]_{p \in \mathcal{P}}.$

\textbf{Decision Variable} –  
At each decision epoch $k$, the receptionist selects an action  $a_k \in \mathcal{A}_k = \{0\} \cup \mathcal{P}_k$, where $a_k = 0$ corresponds to rejecting the request and $a_k = p$ corresponds to assigning the patient to physician $p$. A physician $p$ is eligible only if $d_k + x^k_{pw} \le L_p$, ensuring feasibility of all actions in $\mathcal{A}_k$.

\textbf{Exogenous Information} –  
The exogenous information at epoch $k$ corresponds to the characteristics of the next arriving patient and is denoted by $\xi_k = [d_{k+1}, r_{k+1}, p_{k+1}, \mathcal{P}_{k+1}].$

\textbf{Transition Function} –  
The transition function $X^M(\cdot)$ maps the system from state $x_k$ to $x_{k+1}$ according to $
x_{k+1} = X^M(x_k, a_k, \xi_k).$ Specifically, for the selected physician $p = a_k$,
\begin{itemize}
    \item $x^{k+1}_{pl} = x^{k}_l - {1}$,
    \item $x^{k+1}_{p1} = x^{k}_{p1} + {1}_{r_k=1}$,
    \item $x^{k+1}_{pw} = x^{k}_{pw} + d_k$.
\end{itemize}

\textbf{Cost Function} –  
At each decision epoch, a cost $C_k(x_k,a_k)$ is incurred, defined as
\[
C_k(x_k,a_k) =
\begin{cases}
c^{\text{rej}}_{r_k}, 
& \text{if } a_k = 0, \\[6pt]
c^{\text{pref}}_{r_k}, 
& \text{if } a_k \neq 0 \text{ and } a_k \neq p_k, \\[6pt]
0,
& \text{otherwise.}
\end{cases}
\]

\subsection{Results}\label{results}

To demonstrate how different expert properties influence learning outcomes, we evaluate the performance of \Cref{gdagger} on the physician-to-patient assignment (PPA) problem. All expert optimisation models are solved using Gurobi. The full experimental codebase is publicly available at \href{https://github.com/prakashgawas/OCS}{GitHub}. All experiments are conducted on a machine equipped with an Intel Xeon Gold 6258R CPU running at 2.70\, GHz.
The remainder of this section is organised as follows. \Cref{exsetup} describes the experimental setup, including instance generation, evaluation metrics, baseline policies, and the computational environment. \Cref{perfeval} presents and discusses the experimental results in detail.

\subsubsection{Experimental Setup}\label{exsetup}

This subsection describes the instance generation procedure and the configuration of the learning models and baseline methods.

\subsubsection*{Data}\label{data} - 
We begin by describing the parameters used to generate instances of the PPA. The proposed approach is evaluated on synthetically generated instances constructed under controlled conditions. The total number of patient calls arriving at the clinic is denoted by $K$, as each patient corresponds to a decision epoch. Let $N_1$ and $N_2$ denote the number of patients that arrive with priority 1 and 2, respectively. 
Both $N_1$ and $N_2$ are sampled from a normal distribution where $N_1 \sim \mathcal{N}(25, 5)$ and $N_2 \sim \mathcal{N}(75, 12)$. 

To capture the tendency of high-priority (emergency) patients to call later in the day, we generate an \emph{arrival score} $s_k$ for each patient from a beta distribution, $\mathrm{Beta}(a, b)$. Specifically, arrival scores are drawn with parameters $(a=3, b=1)$ for priority~1 patients and $(a=1, b=1)$ for priority~2 patients. Patients are then ordered by these arrival scores, ensuring that high-priority patients are more likely to appear toward the end of the scheduling horizon.

Appointment durations are assumed to be known at the time of the call and are sampled from a lognormal distribution. For priority 1 patients, durations are generated with parameters $\mu = 3$ and $\sigma = 1.2$, while for priority 2 patients, the parameters are $\mu = 2.3$ and $\sigma = 0.8$. We consider a setting with $P = 4$ physicians, each having a duration capacity of $T = 300$ minutes and $20$ available appointment slots. To construct the eligibility set for each patient, physician preference weights are first sampled from a Dirichlet distribution with parameter vector $[0.4, 0.3, 0.15, 0.15]$. The top $k$ physicians with the largest weights are then selected, where $k$ is drawn uniformly from $\mathcal{U} \sim \{1, \dots, P\}$. A detailed description of this procedure is provided in the appendix.

Rejection penalties are set to $c^{\text{rej}}_{1} = 200$ for priority~1 patients and $c^{\text{rej}}_{2} = 50$ for priority~2 patients. Assigning a patient $i$ to a non-preferred but eligible physician incurs a penalty of $c^{\text{pref}}_{r_i} = 0.1 \times c^{\text{rej}}_{r_i}$. 

\subsubsection*{ML Models} - 
We construct a simulation model of the PPA using the data generation parameters described above. At each decision epoch, the learned policy is invoked to select an action. For each decision epoch, we generate a training example of the form $(x_k,\hat{a}_k)$, where $x_k$ denotes the concatenation of the state features and $\hat{a}_k$ is the target action provided by the expert at epoch~$k$. The learning model maps the current state features to a decision that either assigns the patient to an eligible physician or rejects the request.

As the learning architecture, we employ a multilayer perceptron (MLP) consisting of an input layer, a single hidden layer with 64 neurons, and an output layer. Normalised state features are provided as input, and the MLP outputs a probability distribution over the available actions, namely assignment to each eligible physician or rejection. The final action is selected based on this predicted probability vector.

\subsubsection*{Baselines} - 
We compare the learned policies against three baseline methods. The first is a greedy policy, denoted by $\pi_{\text{greedy}}$, which assigns an arriving patient to their preferred physician whenever possible. If the preferred physician is unavailable, the patient is randomly assigned to any eligible physician. This policy also serves as a myopic expert.

The second baseline, denoted by $\pi^{t}$, corresponds to a two-stage stochastic 
expert. At each decision epoch, this policy samples 30 future demand scenarios. 
These scenarios are generated by simulating future arrivals conditioned on the 
current system state via a Bayesian posterior sampling procedure that exploits 
the priority-specific Beta arrival distributions to form a posterior over unknown 
pool sizes at each epoch (see Appendix~\ref{app:bayesian_scenario}). The expert 
solver is then invoked to compute an action by solving the resulting two-stage 
stochastic optimisation problem over the generated scenarios. We add a stopping 
criterion of a time limit of 30 seconds and an MIP gap of 0.2\% for this method 
to keep the computation time in control.

The final baseline is the exact optimal full-information policy, denoted by $\pi^{*f}$, computed separately for each test instance. This policy represents an upper bound on achievable performance and serves as a benchmark for assessing the quality of learnt and heuristic policies.

\subsubsection{Performance Evaluation}\label{perfeval}

We evaluate multiple configurations of expert models using \Cref{gdagger} and compare their performance against the baseline policies. All methods are tested over another set of 1{,}000 instances. For each policy, we report the average objective value, the total number of rejected patients by priority class, and the total number of assignments to non-preferred physicians, aggregated across both priority classes. 

\subsubsection*{Baselines} - \Cref{tab:res8} reports the performance of the baseline policies on the test instances, including average cost, rejection rates by priority class, and the proportion of undesirable assignments. The primary driver of performance differences across policies is the rejection of priority-1 patients, who tend to arrive late in the booking horizon. The greedy policy $\pi_{\text{greedy}}$ yields the weakest performance, largely attributable to its high priority-1 rejection rate. The two-stage stochastic policy $\pi^{t}$ substantially outperforms the greedy baseline by curtailing priority-1 rejections, though this comes at the expense of a moderate increase in assignments to non-preferred physicians, reflecting a trade-off between the two cost components. Despite being subject to a time limit, $\pi^{t}$ remains computationally demanding: with 30 sampled scenarios and 10 episodes executed in parallel, it requires approximately 31 hours to complete. Finally, $\pi^{*f}$ establishes an upper bound on achievable performance, attaining the 
lowest priority-1 rejection rate among all policies considered. As this policy assumes 
perfect foresight over all patient arrivals, it is unattainable in practice and serves 
purely as a benchmark for assessing the optimality gap of learnt and heuristic policies.

\begin{table}[!ht] 
\centering
\caption{Average performance metrics for  $\pi_{greedy}$, $\pi^{t}$ and $\pi^{*f}$}
\label{tab:res8}
\resizebox{14cm}{!}{
\begin{tabular}{|c|c|c|c|c|}
\hline
\textbf{Policy}           & \textbf{ Cost} & \textbf{Priority 1 Rejected \%} & \textbf{Priority 2 Rejected \%} & \textbf{Undesirable Assignments (\%)} \\ \hline
\textbf{$\pi_{greedy}$ }    & 3380.1                & 46.5                       & 23.2                        & 21.2                            \\ \hline
\textbf{$\pi^{t}$}          & 1647.2                & 5.8                        & 30.2                        & 29.2                            \\ \hline
\textbf{$\pi^{*f}$}         & 1299.2                & 1.2                        & 28.1                        & 47.7                            \\ \hline
\end{tabular}}
\end{table}

\subsubsection*{Learning without interaction (Isolated Learning)} -  We begin by reporting results in the absence of expert interaction. The hyperparameter settings  for \Cref{gdagger} are summarised in \Cref{hypgdag} and define the base configuration of the 
algorithm; any deviations are explicitly noted in the corresponding experimental subsections. 
Setting $I = 1$ restricts the procedure to a single outer iteration, yielding one trained policy. 
Two stopping criteria are imposed on the expert: a time limit of 30 seconds, beyond which the expert must return the best solution found, and a MIP gap tolerance, which allows early termination once a solution of sufficient quality has been identified.
\begin{table}[!ht]
\centering
\caption{Hyper Parameters for \Cref{gdagger}}
\label{hypgdag}
\resizebox{5cm}{!}{
\begin{tabular}{|c|c|}
\hline
\textbf{Hyperparameters} &                \\ \hline
\textbf{\texttt{STOP()}}                   & $I = 1$    \\ \hline 
\textbf{$F_j(.)$}                     & 1              \\ \hline
\textbf{$\kappa_1$ - (Time Limit)}                    & 30 secs    \\ \hline
\textbf{$\kappa_2$ - (MIP Gap)}                    &  2\%       \\ \hline
{Decision Rule}            & Vanilla DAgger \\ \hline
\textbf{$\lambda$}                   & 1              \\ \hline
\textbf{$\beta$}                     & 0.8            \\ \hline
\textbf{$F_e(\cdot)$}                     & $\pi^{f}$, $\pi^{d}$, $\pi^{t}$        \\ \hline
\end{tabular}}
\end{table}

Three expert policies are considered as training targets: the full-information expert 
($\pi^{*f}$), the deterministic expert ($\pi^{*d}$), and the two-stage stochastic expert 
($\pi^{*t}$). The multi-stage expert $\pi^*$ is excluded, as it is computationally intractable 
even for small problem instances. Each expert is queried once per state, and the corresponding 
learned policies are denoted by $\hat{\pi}^{f}$, $\hat{\pi}^{d}$, and $\hat{\pi}^{t}$, 
respectively. A total of $H = 4{,}000$ episodes are simulated to construct the demonstration 
dataset, yielding approximately 400{,}000 expert-labelled state--action pairs per target. 
Due to the imposed stopping criteria---a time limit and a MIP gap tolerance---the resulting 
demonstrations represent near-optimal, rather than strictly optimal, solutions. We further 
note that employing the full-information expert $\pi^{f}$ within \Cref{gdagger} is equivalent 
to solve a deterministic instance of the PPA from scratch using an MIP solver at each decision epoch.

For the two-stage stochastic expert ($\pi^{t}$), the number of future scenarios sampled at 
each decision epoch must be specified. We adopt a fixed scenario count across all epochs, 
though adaptive designs---in which more scenarios are generated at later stages where the 
problem is easier to solve and capacity decisions are more consequential---represent a natural 
extension. Three scenario configurations are evaluated: 5, 15, and 30, yielding a total of 
five policy configurations across all expert types. Data collection is parallelised across 
10 threads, with 10 episodes executed simultaneously; all reported training times and results 
reflect this parallel setup.

\Cref{tab:res1} reports the performance of all learnt policies on the test instances. Policies 
trained using $\pi^{t}$ as the target achieve marked reductions in patient rejections relative 
to the other expert configurations. As expected, the 30-scenario configuration yields the 
strongest performance, coming closest to the two-stage stochastic baseline $\pi^{t}$. However, 
this performance gain entails a higher computational cost, as multiple scenarios must be sampled 
and the corresponding multi-scenario optimisation problem 
$(\textbf{\textsc{MIP}}_{\textbf{SPPA}})$ must be solved at each decision epoch. By contrast, 
the full-information and deterministic experts solve their respective subproblems near-instantly, 
resulting in substantially lower training times.

\begin{table}[!ht]
\centering
\caption{Average performance metrics for the best $\hat{\pi}^{\cdot}$ compared to two-stage stochastic baseline}
\label{tab:res1}
\resizebox{14cm}{!}{
\begin{tabular}{|c|c|c|c|c|c|c|}
\hline
\textbf{  Policy}       & \textbf{Scenarios}       & \textbf{ Cost} & \ \textbf{\begin{tabular}[c]{@{}c@{}}Priority 1 \\ Rejected \%\end{tabular}} &  \textbf{\begin{tabular}[c]{@{}c@{}}Priority 2 \\ Rejected\%\end{tabular}} &  \textbf{\begin{tabular}[c]{@{}c@{}}Undesirable \\ Assignments (\%)\end{tabular}}& \textbf{Train Time (hours)} \\ \hline
\textbf{Full-Information ($\hat{\pi}^{f}$)} & 1 & 2137.4              & 15.0                       & 31.9                    & 26.4      &      0.5                \\ \hline
\textbf{Deterministic ($\hat{\pi}^{d}$)}  & 1  & 1961.9              & 11.4                       & 31.3                        & 27.7            &     1           \\ \hline
\multirow{3}{*}{\textbf{Two-Stage ($\hat{\pi}^{t}$)}} & 5           & 1758.3           & 7.7           & 30.5          & 29.2          &     3.8               \\ \cline{2-7} 

                 & 15                 & 1695.1              & 6.1                     &  31.2                       & 28.9      &     12.1                \\ \cline{2-7} 
                        & 30                 & 1678.5              & 6.0                        & 30.9                       & 29.0         &    23.7               \\ \hline    
\textbf{Two Stage Baseline (${\pi}^{t}$)}  & 30  &  1647.2                & 5.8                        & 30.2                        & 29.2       &     -           \\ \hline
\end{tabular}}
\end{table}

\begin{figure}[!ht]
    \centering
        \caption{Average Gap and Average solve time across decision epoch}
    \label{fig:epochvsgapsolve}
    \includegraphics[width=0.5\linewidth]{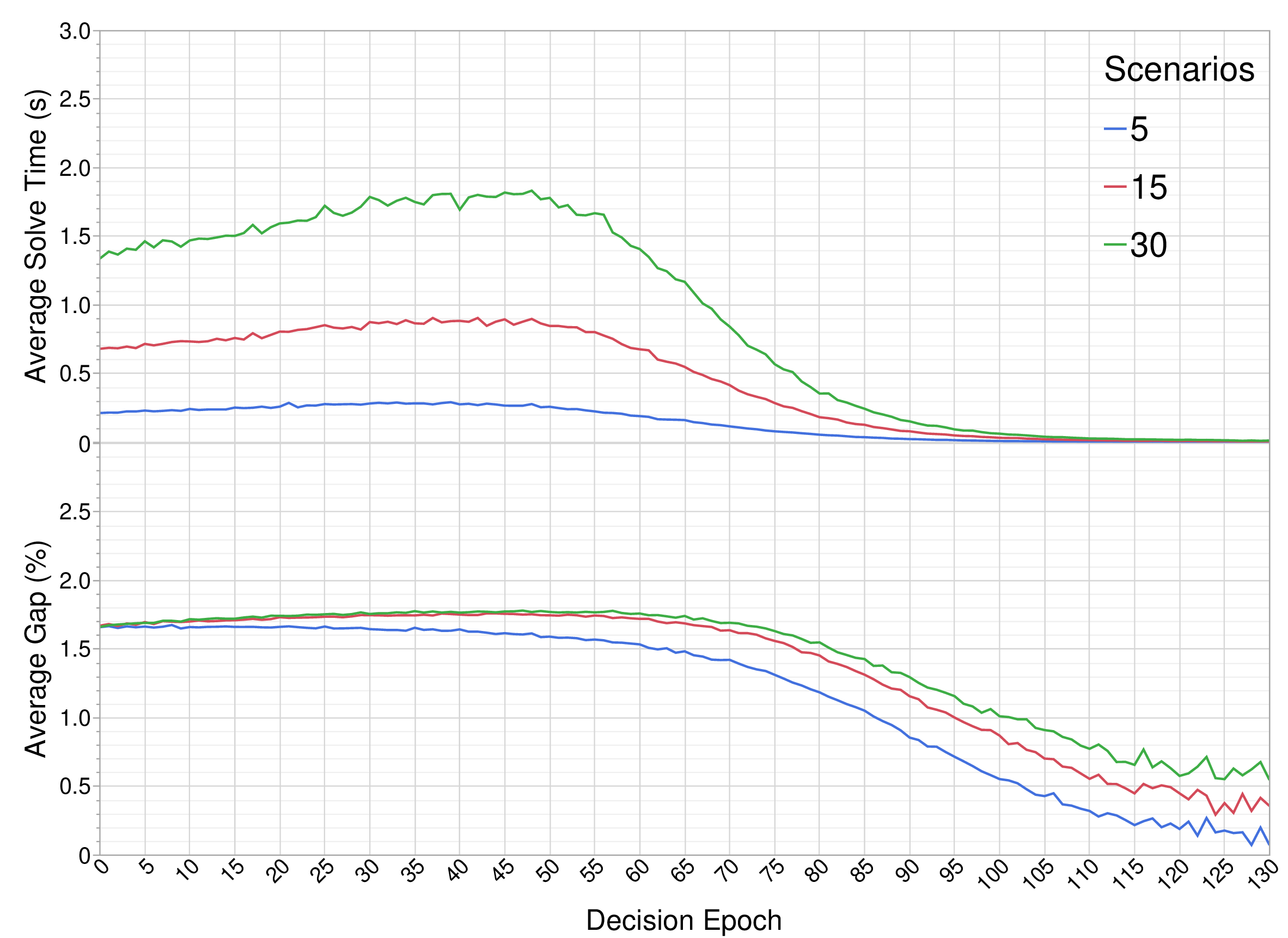}
\end{figure}

\Cref{fig:epochvsgapsolve} illustrates how solver time and MIP gap evolve across decision 
epochs when using the two-stage stochastic expert, with average values plotted per epoch. 
Solve times are longest in the early epochs, where the larger problem size at the start of 
an episode causes the MIP gap criterion to become active. In the absence of a gap limit, 
the expert may exhaust the full 30-second time budget without identifying an optimal solution, 
leading to substantially higher training times; preliminary experiments without this criterion 
showed training times at least three times longer than those reported here. For the PPA, the 
MIP gap tolerance therefore, emerges as the more consequential stopping criterion, offering a 
favourable trade-off between demonstration quality and computational cost. For more complex 
or large-scale problems, however, the time limit is expected to play an equally important 
role, particularly when even moderately sized instances cannot be solved to a prescribed 
optimality gap within the available budget. In contrast, solving 
$\textbf{\textsc{MIP}}_{\textbf{PPA}}$ under the full-information and deterministic experts 
requires an average of only 0.2 seconds per instance.

\begin{figure}[!ht]
    \centering
        \caption{Expert rejection trend across decision epoch}
    \label{agg_actions}
    \begin{subfigure}{0.48\linewidth}
           \caption{Priority 1}
           \label{fig:agg1}
        \includegraphics[width=\linewidth]{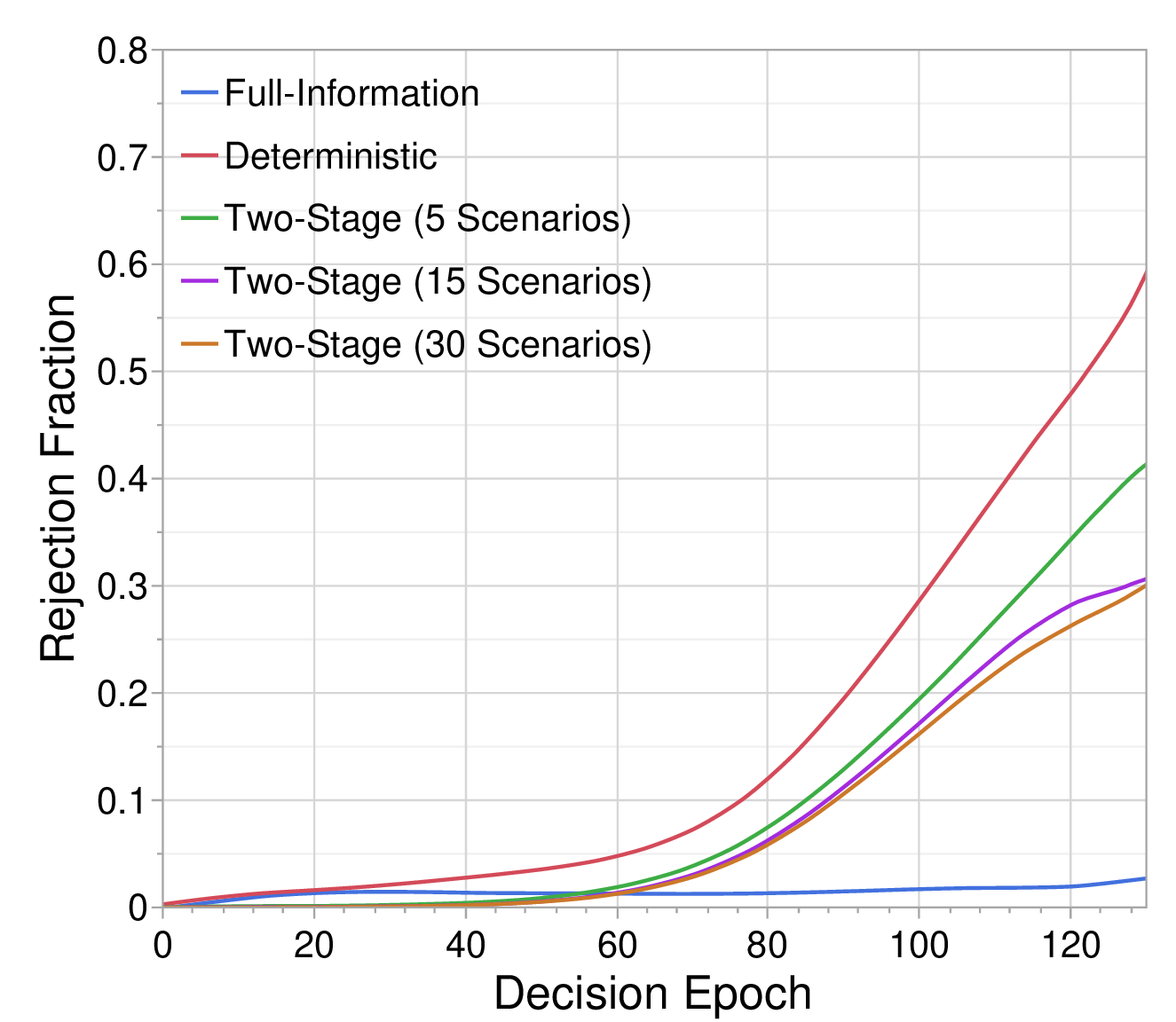}
    \end{subfigure}
    \hfill
    \begin{subfigure}{0.48\linewidth}
            \caption{Priority 2}
            \label{fig:agg2}
            \includegraphics[width=\linewidth]{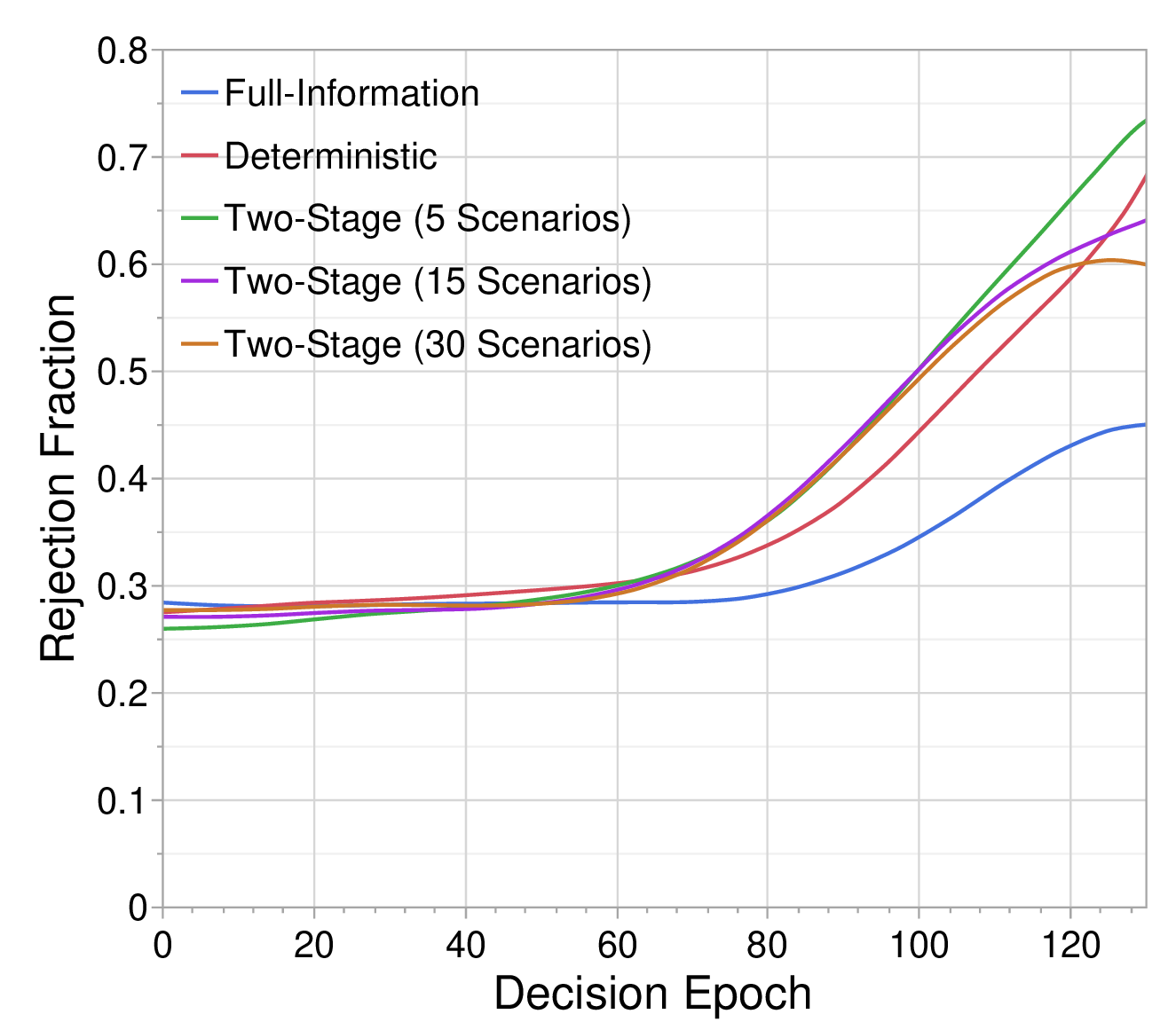}
    \end{subfigure}
    \vspace{2mm}
    \footnotesize
\end{figure}

Finally, we compare the expert actions obtained during the training in \Cref{agg_actions}. Since rejection incurs the highest cost, we plot the average rejection fraction for each decision epoch for both priority levels. The plot clearly differentiates the experts in terms of their demonstrations. The Full-Information expert tends to never reject Priority 1 patients, while the Deterministic expert rejects the most.  Note here that priority 1 patients arrive much later in the episode, and hence we see all experts not rejecting any priority 1 at the start. Realistically, imitating the Full-Information expert is fundamentally limited 
by the information asymmetry between training and deployment. The expert 
observes future patient arrivals when generating demonstrations, whereas the 
learned policy must act without this knowledge, creating a systematic mismatch 
between demonstrated and imitable behaviour that degrades policy performance 
regardless of learning quality. The Two-stage experts find a middle ground between the two other experts. For priority 2 patients, we see a constant rejection rate for all the experts for initial decision epochs. As the priority 1 patients start arriving, we see a higher percentage of priority 2 rejections. The Full-Information expert, however, rejects the fewest patients. This clearly explains the difference between experts in terms of the quality of demonstration.  

\subsubsection*{Learning with interaction}
We next consider the interactive learning setting, in which the expert is queried iteratively 
to refine the learned policy. The stopping criterion is set to $I = 400$ learning iterations, 
with the current policy $\pi_i$ executed for $H = 10$ episodes per iteration, yielding a 
total of 4{,}000 simulated episodes --- matching the episode budget used in the isolated 
learning setting.

\begin{figure}[!ht]
    \centering
        \caption{Validation performance of different learned models with interaction}
    \label{fig:epi_comparison2}
    \begin{subfigure}{0.48\linewidth}
           \caption{ With respect to learnt models}
           \label{fig:stoch}
        \includegraphics[width=\linewidth]{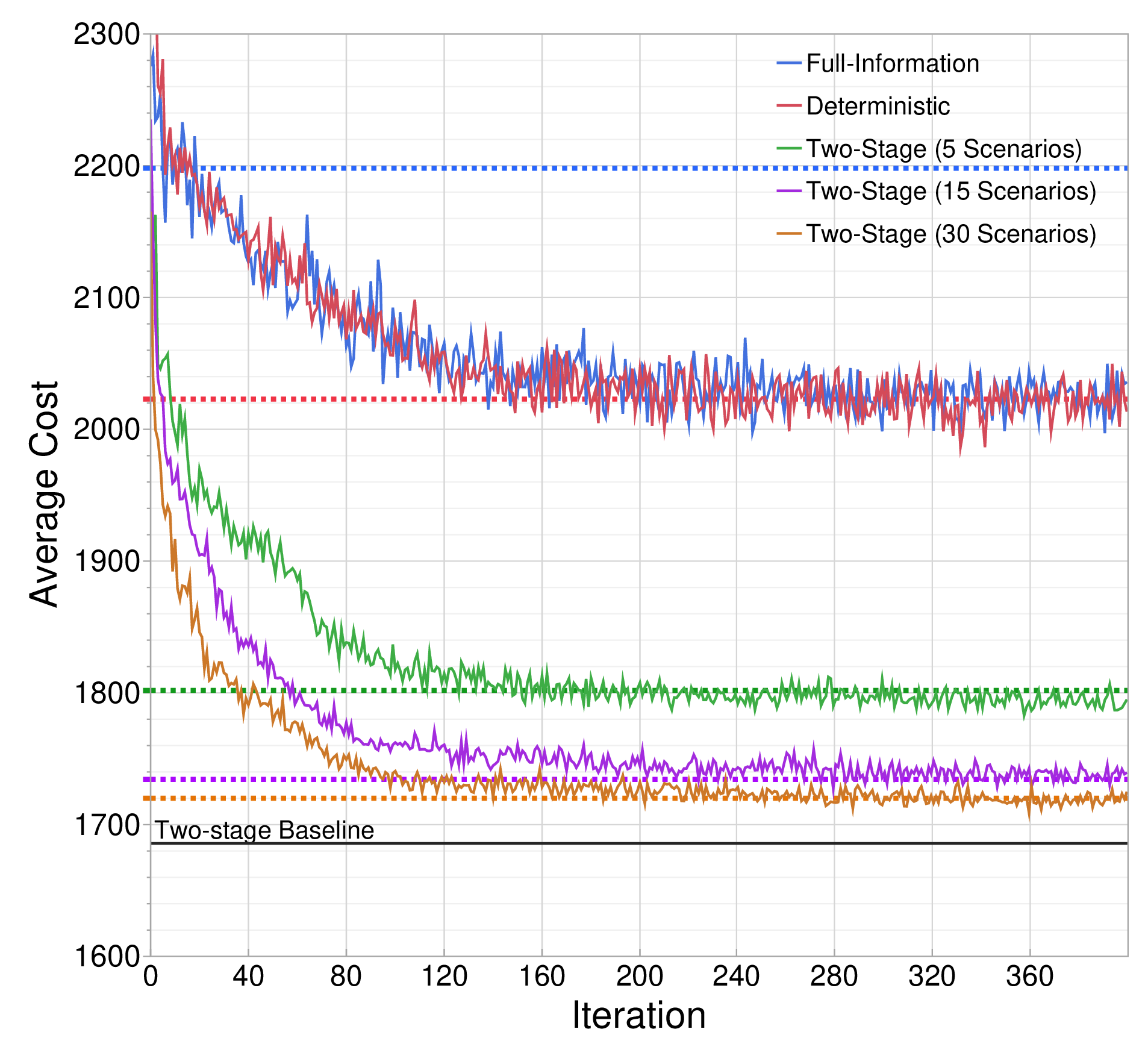}
    \end{subfigure}
    \hfill
    \begin{subfigure}{0.48\linewidth}
            \caption{With respect to time}
            \label{fig:stoch_time}
            \includegraphics[width=\linewidth]{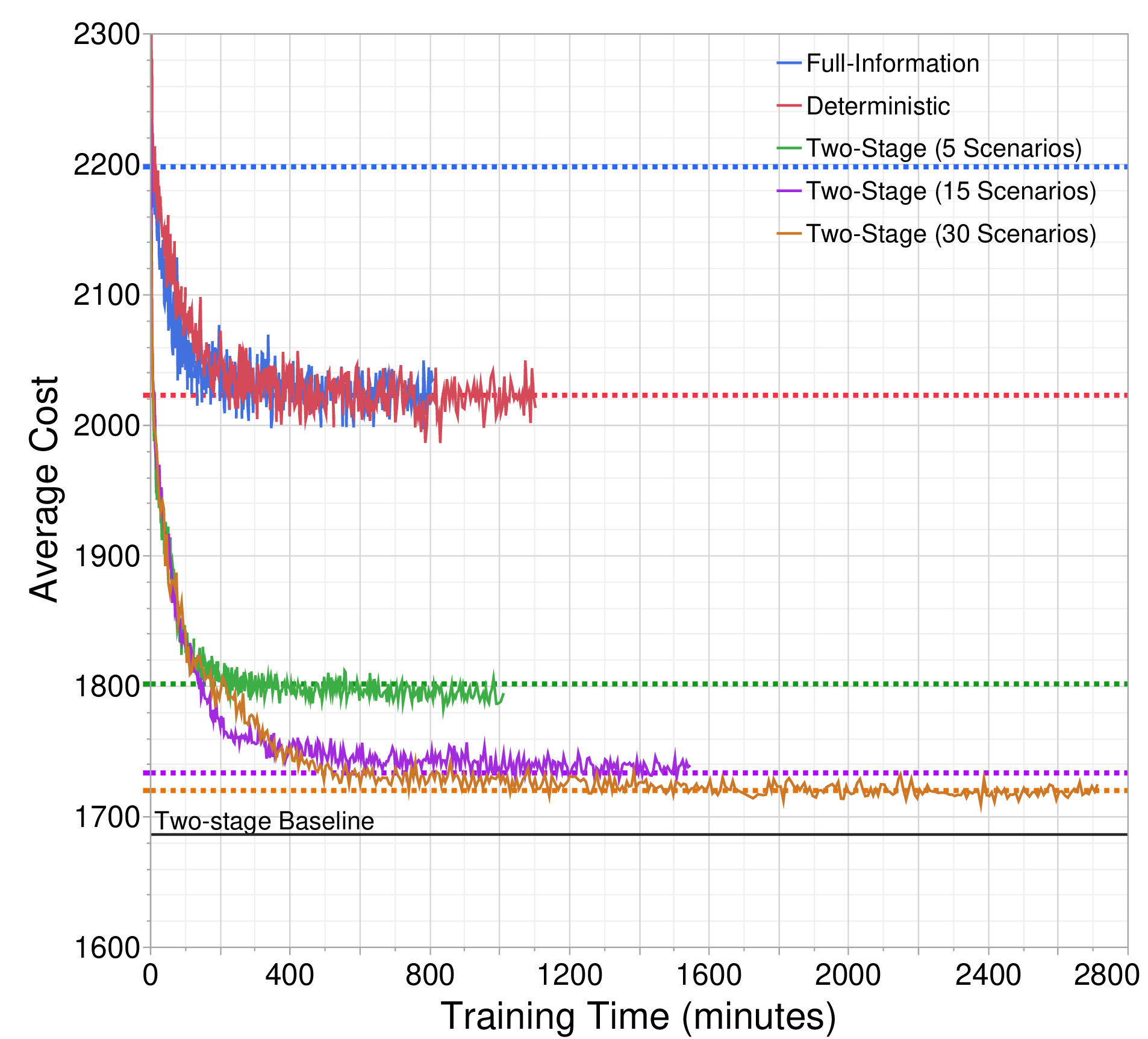}
    \end{subfigure}
    \vspace{0mm}
    \footnotesize
\textit{Note:} The dotted lines indicate models learnt without interaction from \Cref{tab:res1}.
\end{figure}

Upon completion of training, all learnt policies are evaluated on a separate validation set 
of 1{,}000 instances, and the policy achieving the lowest average cost is selected. 
\Cref{fig:epi_comparison2} reports the evaluation of all learnt policies using the average cost for all expert types for the validation instances. The horizontal axis corresponds to the learnt policy $\pi_i$ obtained after iteration $i$. We also show the average objectives of the best models obtained through isolated learning, indicated by dotted lines in the figure.  Across all expert configurations, the average cost decreases consistently as the training dataset grows through successive expert interactions. 
\Cref{tab:res5} summarises the performance of the best learnt policies on the test set. 
Across all expert configurations, interactive learning yields lower average costs than their 
isolated learning counterparts reported in \Cref{tab:res1}. The policy trained with the 
two-stage stochastic expert using 30 scenarios attains the strongest overall performance, 
approaching the two-stage stochastic baseline $\pi^{t}$. Gains in learning performance 
begin to plateau after approximately 250 iterations (2{,}500 instances), beyond which the 
best-performing models are identified. This 
saturation likely reflects sufficient coverage of the relevant state space, after which 
additional expert queries yield diminishing returns in terms of policy improvement. Notably, policies learned from the two-stage stochastic 
expert exhibit lower variance across consecutive iterations compared to those trained from 
other expert types. Crucially, once trained, all learned MLP policies produce decisions in 
milliseconds at deployment, in contrast to the two-stage stochastic baseline 
$\pi^{t}$, which must solve an MIP at each decision epoch, 
confirming that the computational cost of expert optimisation is effectively 
amortised over the training phase.

\begin{table}[!ht]
\centering
\caption{Average performance metrics for the best model with interaction on test instances}
\label{tab:res5}
\resizebox{14cm}{!}{
\begin{tabular}{|c|c|c|c|c|c|c|c|c|}
\hline
\textbf{Expert}        & \textbf{Scenarios} & \textbf{\begin{tabular}[c]{@{}c@{}}Num. of Episodes\\ to best model\end{tabular}} & \textbf{Cost} & \textbf{\begin{tabular}[c]{@{}c@{}}\% Improvement on  \\ Isolated Learning\end{tabular}} &  \textbf{\begin{tabular}[c]{@{}c@{}}Priority 1 \\ Rejected\%\end{tabular}} &  \textbf{\begin{tabular}[c]{@{}c@{}}Priority 2 \\ Rejected\%\end{tabular}} &  \textbf{\begin{tabular}[c]{@{}c@{}}Undesirable \\ Assignments (\%)\end{tabular}} & \textbf{Train Time (hours)} \\ \hline
\textbf{Full-Information } ($\hat{\pi}^{f}$) & 1                  &    3910                 & 1957.0   & 8.5      & 11.5                         & 30.9                         &         28.0            & 13.5              \\ \hline
\textbf{Deterministic}  ($\hat{\pi}^{d}$)  & 1                  &     3420              & 1943.3     & 0.9        & 11.5                        &      30.6                     &        28.0     & 18.3                      \\ \hline
\multirow{3}{*}{\textbf{\begin{tabular}[c]{@{}c@{}}Two-stage\\ Stochastic  ($\hat{\pi}^{t}$)\end{tabular}}} & 5                  &          3580           & 1748.2  & 0.5       & 7.3                        & 30.9                         &   29.2     & 17.0                            \\ \cline{2-9} 
                                             & 15                 &     2800                &     1693.5    & 0.1      &            6.5                  &     30.4                        &                 29.4    & 25.8              \\ \cline{2-9} 
                                             & 30                 &    3600                 &    1674.2    & 0.2       &        6.0                      &            30.8                  &              29.0     & 45.2                \\ \hline
\end{tabular}}
\end{table}

\Cref{tab:res5} also reports the percentage improvement relative to the isolated learning 
models. While average costs decrease consistently with additional iterations, the most 
notable finding is that meaningful gains from DAgger are observed primarily for 
the full-information expert; the deterministic and two-stage stochastic experts 
yield only marginal improvements over their isolated learning counterparts. 
This suggests a useful diagnostic: the benefit of interactive learning is 
greatest precisely when the  information advantage of the expert over the deployed 
policy is largest, as in the full-information case, and diminishes when the expert 
demonstrations already cover a representative portion of the deployment state 
distribution, as with deterministic and two-stage stochastic experts that 
operate without knowledge of future realisations.

The key practical implication is that interactive data collection via DAgger is most 
consequential when using the full-information expert, for which state diversity is 
critical to achieving strong policy performance. This finding is consistent with the 
taxonomy as the two-stage stochastic experts and the deterministic expert without the knowledge of the future, generate demonstrations across a richer 
and a more representative portion of the state space, thereby reducing the benefits 
of DAgger-based exploration relative to the full-information expert, whose demonstrations 
are confined to a narrower set of trajectories.

Training times are also reported in \Cref{tab:res5}. Under interactive learning, the training time for the deterministic expert increases by more than a factor of 2. For the full-information expert, training time increases substantially when using the vanilla DAgger decision rule. Under this rule, the states encountered during learning differ from those observed by a pure full-information expert, which necessitates solving the $\textbf{\textsc{MIP}}_{\textbf{PPA}}$ at each newly encountered state. Similarly, the two-stage stochastic expert also sees a considerable increase in training times. 

 \Cref{fig:stoch_time} shows the same plot with respect to training time in minutes. Within the first 3 hours of training, we see that the Two-Stage Stochastic Expert with 5 scenarios shows better test performance than those with 15 and 30.  Similarly, the one with 15 scenarios is better than 30 scenarios for the next 3 hours. This suggests that adding more states and making greater efforts in the short term is better. But eventually, if one allows more time, the higher-scenario model gives better results. 

\begin{figure}[!ht]
    \centering
        \caption{Expert rejection trend across decision epoch}
    \label{agg_actions2}
    \begin{subfigure}{0.48\linewidth}
           \caption{Priority 1}
           \label{fig2:agg1}
        \includegraphics[width=\linewidth]{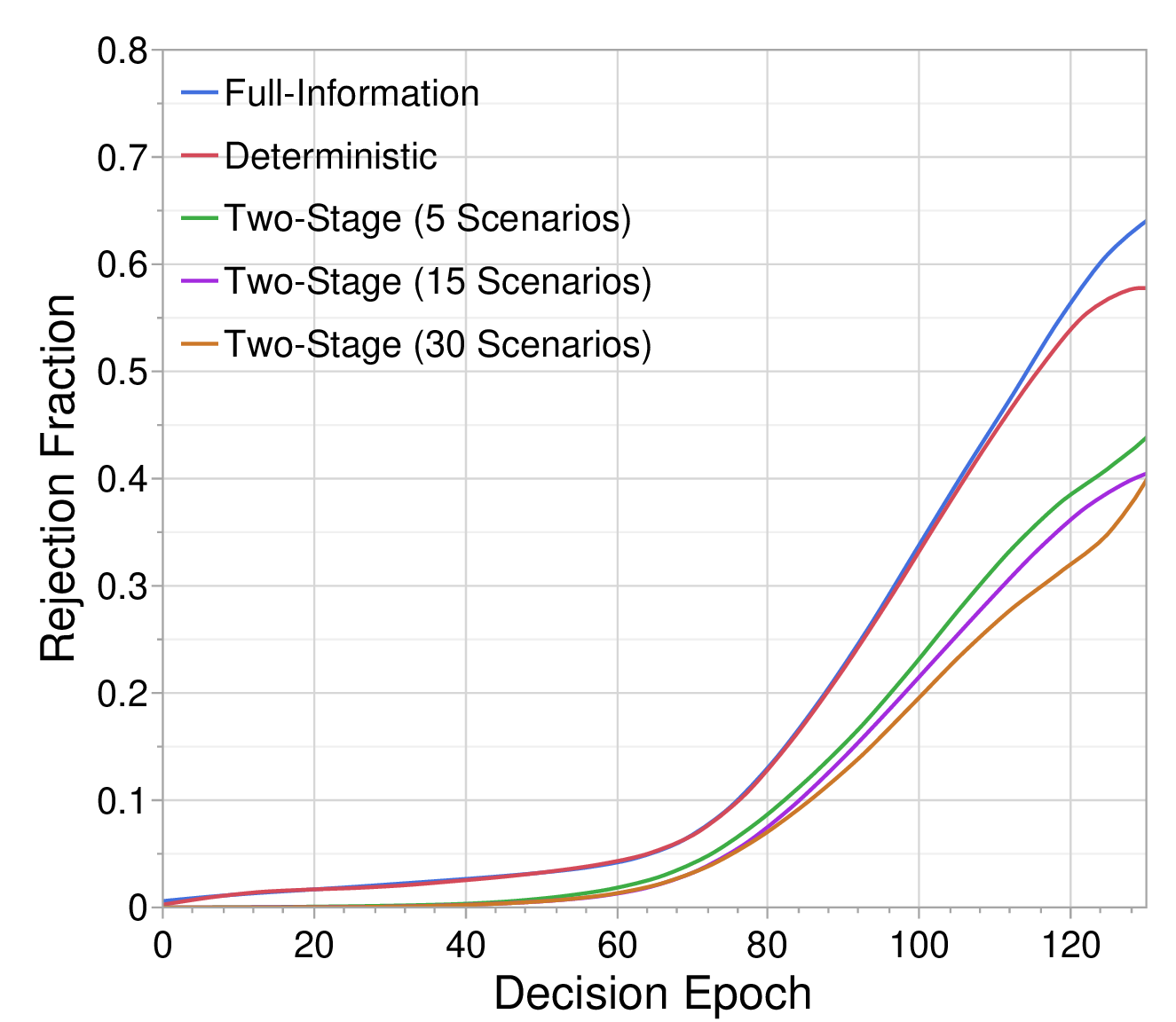}
    \end{subfigure}
    \hfill
    \begin{subfigure}{0.48\linewidth}
            \caption{Priority 2}
            \label{fig2:agg2}
            \includegraphics[width=\linewidth]{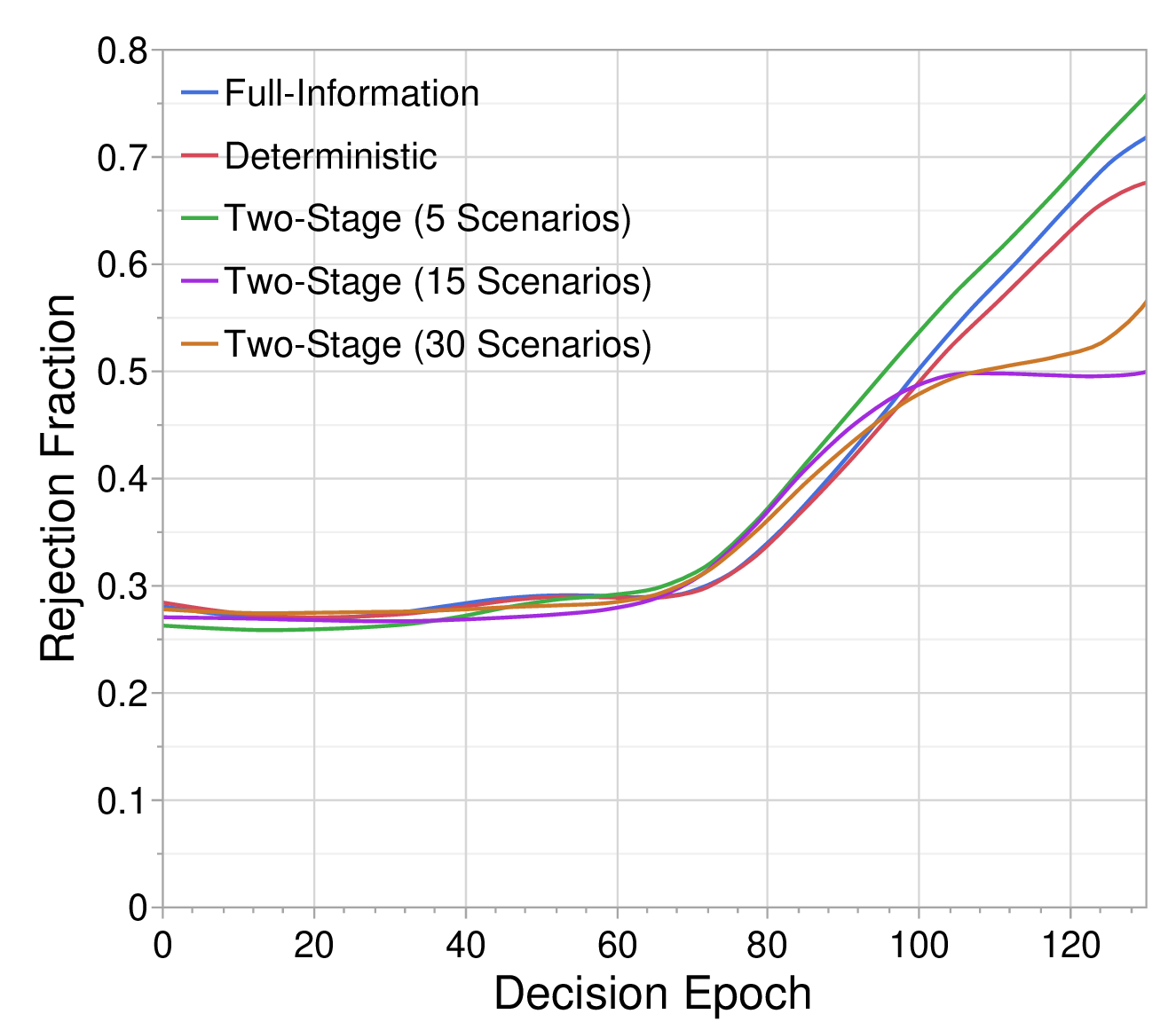}
    \end{subfigure}
    \vspace{2mm}
    \footnotesize
\end{figure}

\Cref{agg_actions2} shows the average rejection fraction in the final dataset after 400 DAgger iterations. Compared to \Cref{agg_actions}, we see a marked difference in the demonstrations of the Full-Information expert. As we are implementing the learned policy, if we reach new states, we have the opportunity to re-optimise by calling the expert again from this new state. This helps gather more diverse demonstrations. The overall trend is very similar to the Deterministic expert for both priority patients. However, for the other experts, by implementing the learned policy, we do not encounter states outside the distribution of the corresponding expert, and thus, we only see marginal gains. 


\subsubsection*{Compromising Optimality of Experts}
\begin{figure}[!ht]
    \centering
        \caption{Validation performance of different learned models with interaction for different MIP tolerances}
    \label{fig:mg_perf}
    \includegraphics[width=0.5\linewidth]{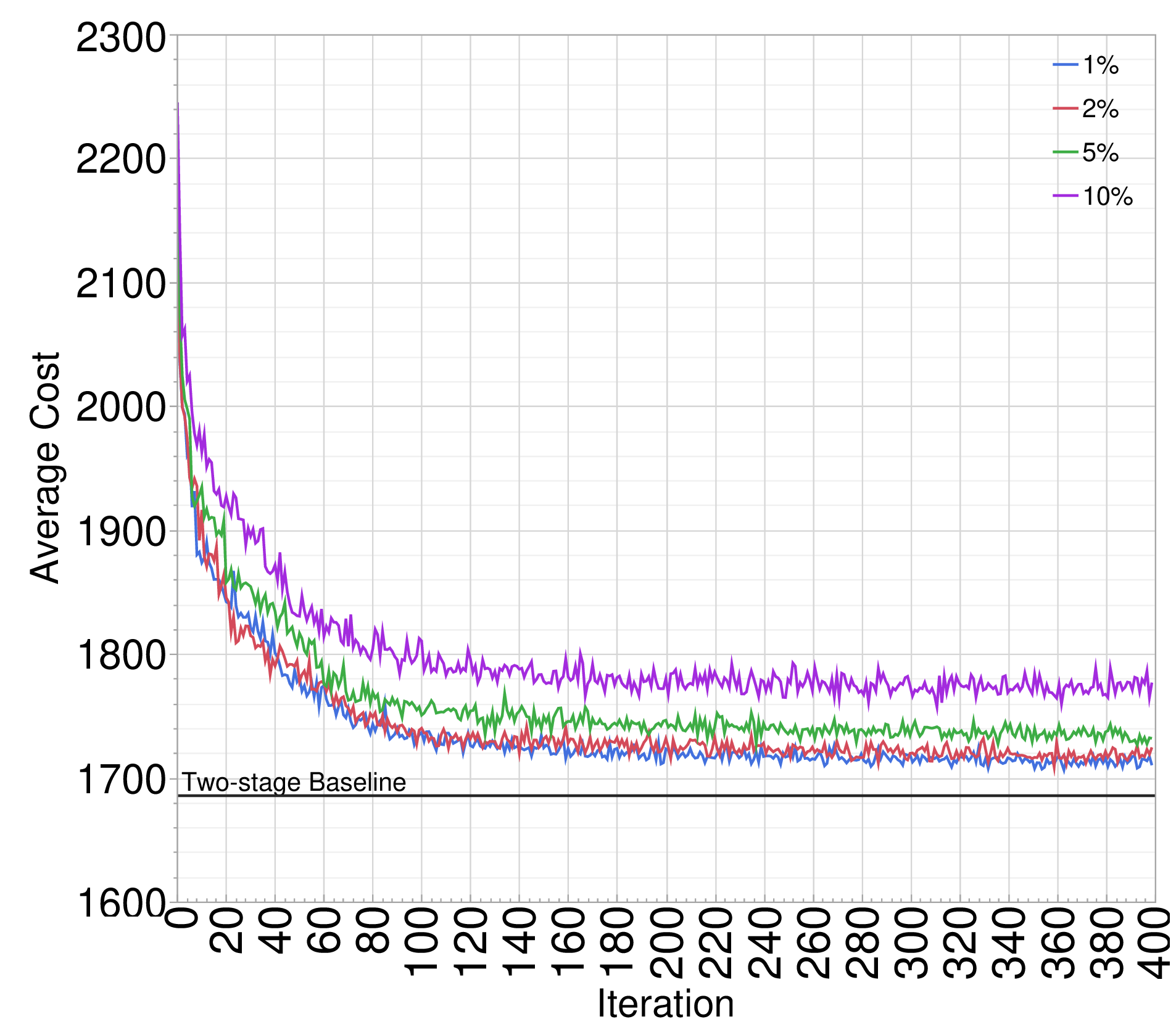}
\end{figure}

Having established in \Cref{fig:epi_comparison2} that the two-stage stochastic expert yields 
the strongest performance, we investigate whether its computational burden can be reduced by 
relaxing the optimality requirement of the underlying solver. To this end, we vary the MIP gap 
parameter while training with the two-stage stochastic expert using 30 scenarios, considering 
four values: $1\%$, $2\%$, $5\%$, and $10\%$. The corresponding validation set performance 
is reported in \Cref{fig:mg_perf}. Learning behaviour is broadly consistent across gap limits 
of $1\%$ and $2\%$, whereas the $5\%$ setting introduces a marginal degradation and the 
$10\%$ setting leads to a more pronounced deterioration in performance, indicative of 
lower-quality demonstrations at this level of approximation. Notably, despite yielding 
negligible differences in solution quality, tightening the gap to $1\%$ approximately doubles 
training time relative to the $2\%$ setting. The best-performing policy under each 
configuration is summarised in \Cref{tab:res12}. Collectively, these results identify a MIP 
gap of $2\%$ as a practical sweet spot: it achieves solution quality comparable to the 
tightest tolerance considered, at roughly half the computational cost, confirming that 
effective policies can be learnt without solving the two-stage stochastic problem to  
optimality.

\begin{table}[!ht]
\centering
\caption{Average performance metrics for the best model with interaction with 30 scenarios }
\label{tab:res12}
\resizebox{14cm}{!}{
\begin{tabular}{|c|c|c|c|c|c|c|c|}
\hline
\textbf{Expert}                                                                          & \textbf{MIP Gap \%} & \textbf{\begin{tabular}[c]{@{}c@{}}Num. of Episodes\\ to best model\end{tabular}} & \multicolumn{1}{l|}{\textbf{Cost}} & \textbf{\begin{tabular}[c]{@{}c@{}}Priority 1 \\ Rejected\end{tabular}} & \multicolumn{1}{l|}{\textbf{\begin{tabular}[c]{@{}l@{}}Priority 2 \\ Rejected\end{tabular}}} & \textbf{\begin{tabular}[c]{@{}c@{}}Undesirable \\ Assignments (\%)\end{tabular}} & \textbf{Train Time (hours)} \\ \hline
\multirow{4}{*}{\textbf{\begin{tabular}[c]{@{}c@{}}Two-stage\\ Stochastic ($\hat{\pi}^{t}$)\end{tabular}}} & 1             & 3640                 & 1670.8                             & 5.8                                                                    & 30.8                                                                                        & 29.5                                                                       & 99.4                        \\ \cline{2-8} 
                                    & 2        &     3600                 &    1674.2         &        6.0                      &            30.8                  &              29.0     & 45.2                            \\ \cline{2-8} 
             & 5             & 3940                 & 1686.8                             & 6.4                                                                    & 30.7                                                                                        & 29.7                                                                      & 29.8                         \\ \cline{2-8} 
                                & 10              & 3120                 & 1715.8                            & 6.3                                                                    & 31.5                                                                                         & 29.1                                                                       &    26.0                         \\ \hline
\end{tabular}}
\end{table}

%% file: Sec7-conclusion.tex
\section{Open Research Directions}\label{ord}

Despite rapid progress in imitation learning for CO under uncertainty, several important research directions remain open.

\begin{enumerate}

\item \textbf{Adaptive Expert Query Strategies} — Although DAgger-style interaction improves robustness by mitigating compounding errors, expert queries remain computationally expensive in many optimisation settings. Querying the expert in every state may be unnecessary 
and that principled mechanisms for deciding \emph{when} to query based on 
criteria such as model uncertainty, state novelty, or expected performance gain could substantially reduce training cost without sacrificing solution 
quality. Integrating selective sampling and confidence-aware querying 
\cite{zhang2016query, menda2019ensembledagger} from classical IL literature 
into stochastic optimisation frameworks is a particularly promising direction.

\item \textbf{Multi-Expert Supervision} — The taxonomy shows that experts offer complementary strengths: deterministic experts provide efficiency, whereas stochastic experts deliver stronger supervision. As illustrated in \Cref{gdagger}, heterogeneous experts can be combined adaptively based on the observed state. Learning meta-policies for state-dependent expert selection could reduce computational burden while maintaining solution quality. In the PPA, early decision epochs are especially demanding for two-stage stochastic models; employing a milder expert at these stages can yield substantial time savings. A similar idea appears in ambulance routing \cite{rautenstrauss2025optimization}.

\item \textbf{Learning under Structured Expert Imperfection} — Most studies assume bounded optimality gaps or heuristic approximations but offer limited theoretical insight into how expert suboptimality propagates through imitation learning. Establishing performance guarantees that relate expert approximation quality to learned policy performance remains an important open problem, particularly in stochastic settings.

\item \textbf{Sequential Prediction Architectures} — Decision spaces are large in CO problems, making accurate and feasible end-to-end prediction difficult. A promising approach is to decompose decisions into sequential stages, with separate models predicting different components. For example, in the Dynamic Inventory Routing Problem \cite{greif2024combinatorial}, one may first select customers and then determine their visiting order. Training coordinated models, each guided by a specialised expert, may improve scalability and predictive performance.

\end{enumerate}
The findings 
reported in this paper are illustrated through the PPA, which is a 
relatively compact sequential decision problem. Other CO problems such 
as vehicle routing or inventory routing involves larger and coupled action 
spaces, harder feasibility structures, spatially structured uncertainty 
over graph representations, and two-stage formulations that are 
considerably more expensive to solve. Whether the expert design choices 
identified here generalise to such settings remains an open empirical 
question.

\section{Conclusion}\label{conclusion}

This study examined the role of the expert in imitation learning for SDPs under uncertainty. Using the PPA case study, we compared expert configurations in terms of interaction, optimality, and computational effort. Policies trained with the two-stage stochastic expert using 30 scenarios 
achieved the strongest performance, both with and without interaction, while 
interaction further improved policy quality in the context of the PPA, where 
the late-arriving priority structure inherently rewards anticipatory decision 
making, a setting that may favour stochastic experts more strongly than 
problems with uniform uncertainty arrival. Deterministic and full-information 
experts enable rapid generation of demonstrations but tend to produce 
lower-quality supervision, whereas two-stage stochastic experts provide 
stronger guidance at a higher computational cost.

For the PPA, the two-stage formulation remains tractable with a moderate number of scenarios, but computational effort increases substantially in early decision epochs. Relaxing optimality requirements via an MIP gap limit significantly reduced training time without degrading learned policy performance, highlighting the trade-off between expert quality and scalability.

From a computational perspective, learned ML policies amortise the cost of expert computation, producing fast and stable decisions at deployment. Heuristic multi-scenario optimisation methods can reduce computation but may be biased towards sampled scenarios and perform poorly under rare events. In contrast, learnt policies aggregate information across instances, improving robustness and efficiency. The \emph{Aggregated Deterministic Expert}, which solves scenarios independently and aggregates decisions \cite{gawas2024ildagger, greif2024combinatorial}, offers a practical compromise: while not matching two-stage stochastic performance, it consistently outperforms single-scenario deterministic and full-information experts.

Overall, the findings emphasise the central role of expert design in imitation learning for sequential decision making. By formalising expert characteristics through a structured taxonomy and embedding them within a DAgger-based framework, this work provides a unified perspective for interpreting prior results and guiding expert selection. Future work may further explore adaptive expert-query strategies, richer stochastic expert models, and extensions to larger-scale optimisation problems.

%% file: references.tex
\providecommand{\newblock}{}

%% file: appendix.tex
\section*{Appendix}
\section{Background}
\subsection{MDP}
Given the core components from \Cref{background}, we further describe the value function and policy in detail here.
\begin{itemize}
        \item \textbf{Value function} -  
    The value function represents the cumulative cost over the planning horizon.
    For a cost-minimisation objective, the optimal value function from an initial
    state $x_0$ is given by
    \begin{align}
        V^*(x_0) =
        \min_{a} \mathbb{E}\Bigg\{ \sum_{k=0}^{K}
        C_k(x_k,a_k) \,\big|\, x_0 \Bigg\},
        \label{eq1.1}
    \end{align}
    We omit the discount factor for simplicity. 

    \item \textbf{Policy ($\pi$)} -  
    A policy $\pi$ is a mapping from states to admissible actions $a_k = \pi(x_k)$. Policies may be
    represented in tabular form or as parameterised functions, for example, with
    parameter $\theta$. The value of a policy $\pi$ is defined as
    \begin{align}
        V^\pi(x_0) &= \mathbb{E}\Bigg\{ \sum_{k=0}^{K}
        C_k(x_k,\pi(x_k)) \,\big|\, x_0 \Bigg\} 
    \end{align} 
    The optimal policy $\pi^*$ is then defined as
    \begin{align}
        \pi^* &= \argmin_{\pi} \mathbb{E}\Bigg\{ \sum_{k=0}^{K}
        C_k(x_k,\pi(x_k)) \,\big|\, x_0 \Bigg\} \\
              &= \argmin_{\pi} V^{\pi}(x_0).
        \label{eq1.2}
    \end{align}
In the MDP literature, $V^{\pi^*}(x_k)$ is referred to as the optimal cost-to-go from state $x_k$ at epoch $k$. 

\end{itemize}

Given a value function, the optimal policy can be recovered via one-step greedy
minimisation. Value functions for a fixed policy can be computed using dynamic
programming methods such as value iteration or policy iteration
\cite{puterman2014markov}. While these approaches yield exact solutions, they are
often computationally infeasible for large-scale problems due to the curse of
dimensionality \cite{powell2007approximate}.

In some applications, interest lies only in the optimal action at a specific epoch
$k$. The optimal action is given by
\begin{align}
    a_k^* = \argmin_{a_k} \Big\{
    C_k(x_k,a_k) + \mathbb{E}[V^*_{k+1}(x_{k+1})] \,\big|\, x_k \Big\},
    \label{eq1.2b}
\end{align}
where $V^*_{k+1}(x_{k+1})$ denotes the optimal value of the subsequent state.

\section{MIP formulations}
We first provide an MIP formulation for PPA for the static problem as well as the stochastic problem with $\Omega$ scenarios.
\subsection{Deterministic MIP formulation}
We give an MIP formulation for the PPA below. We denote the patients by $k$ and the physicians by $p$.

\begin{table}[!ht]
\caption{Model, Parameters, and Variables }
\label{tab:all_nots}
\centering
\renewcommand{\arraystretch}{1}
\begin{tabular}{llll}
\hline
\textbf{Sets} & & & \\
\hline
$\mathcal{P}$ & & & set of all physicians \\
$\mathcal{K}$ & & & set of all patients \\
$\mathcal{P}_k$ & & & set of all eligible physicians for patient $k$ \\
$\{1, 2\}$ & & & priority classes \\
\hline
\textbf{Parameters} & & & \\
\hline
$P$ & & & number of physicians \\
$K$ & & & number of patients \\
$t_k$ & & & duration of patient $k$ \\
$r_k$ & & & priority of patient $k$  \\
$p_k$ & & & preferred physician of patient $k$  \\
$T$ & & & total work capacity for physician $p$  \\
$L_p$ & & & total number of appointments allowed for physician $p$  \\
$c_{r_k}^{\text{rej}}$ & & & rejection cost for patient $k$ \\
$c_{r_k}^{\text{pref}}$ & & & penalty for assigning an undesired physician to patient $k$ \\
\hline
\textbf{Variables} & & & \\
\hline
$a_{kp} \in \{1,0\}$ & $k \in \mathcal{K}$, $p \in \mathcal{P}$ &  & 1 if patient $k$ is assigned to physician $p$ \\
$u_k \in \{1,0\}$ & $k \in \mathcal{K}$ & & 1 if patient $k$ is rejected \\
\hline
\end{tabular}
\end{table}
The objective function \eqref{eq1} is the sum of the total rejection cost and the penalty for all non-preferred assignments.
Constraint \eqref{eq:assign} ensures that all patients are either assigned to a physician or rejected.
Constraints \eqref{eq:cap} and \eqref{eq:overtime} enforce capacity limits on the number of appointments and total workload, respectively.

\resizebox{14cm}{!}{
\centering
\makebox[\textwidth]{\begin{minipage}{\dimexpr\textwidth-4\fboxsep-8\fboxrule\relax}
\begin{equation}
    \textbf{\textsc{MIP}}_{\textbf{PPA}} := \min \quad 
\underbrace{\sum_{k \in \mathcal{K}} c^{\text{rej}}_{r_k} u_k}_{\text{Rejection cost}}
\;+\;
\underbrace{\sum_{k \in \mathcal{K}} \sum_{p \in \mathcal{P}} c^{\text{pref}}_{r_k} a_{kp}}_{\text{Non-preferred assignment penalty}}
\label{eq1}
\end{equation}

\begin{align}
\textbf{s.t.} \quad 
    & \sum_{p \in \mathcal{P}_k} a_{kp} + u_k = 1 
    && \forall k \in \mathcal{K} \label{eq:assign} \\[4pt]
    & \sum_{k \in \mathcal{K}} a_{kp} \le L_p 
    && \forall p \in \mathcal{P} \label{eq:cap} \\[4pt]
    & \sum_{k \in \mathcal{K}: p \in \mathcal{P}_k} t_k\,a_{kp} \le T  
    && \forall p \in \mathcal{P} \label{eq:overtime} \\[4pt]
    & a_{kp} \in \{0,1\} 
    && \forall k \in \mathcal{K},\ \forall p \in \mathcal{P} \label{eq6} \\[4pt]
    & u_k \in \{0,1\} 
    && \forall k \in \mathcal{K} \label{eq7}
\end{align}
\end{minipage}}
}

\subsection{Two-stage Stochastic MIP formulation}

\Cref{tab:all_notsstoch} introduces additional notation for the two-stage formulation, where variables and parameters are augmented with a scenario index $\omega$. All other variables and sets retain the same meaning as in \Cref{tab:all_nots}. We use $\omega$ to denote a scenario. In a two-stage formulation, the first patient is common across all scenarios and is denoted by $k = 1$.

\begin{table}[!ht]
\caption{Model, Parameters, and Variables }
\label{tab:all_notsstoch}
\centering
\renewcommand{\arraystretch}{1}
\begin{tabular}{llll}
\hline
\textbf{Sets} & & & \\
\hline
$\Omega$ & & & set of all scenarios \\
$\mathcal{K}_\omega$ & & & set of all patients in scenario $\omega$ \\
$\mathcal{P}_{k\omega}$ & & & set of all eligible physicians for patient $k$ in scenario $\omega$ \\
\hline
\textbf{Parameters} & & & \\
\hline
$K_\omega$ & & & number of patients in scenario $\omega$ \\
$t_{k\omega}$ & & & duration of patient $k$ in scenario $\omega$ \\
$r_{k\omega}$ & & & priority of patient $k$ in scenario $\omega$ \\
$p_{k\omega}$ & & & preferred physician of patient $k$ in scenario $\omega$ \\
$c_{r_{k\omega}}^{\text{rej}}$ & & & rejection cost for patient $k$ in scenario $\omega$ \\
$c_{r_{k\omega}}^{\text{pref}}$ & & & penalty for assigning an undesired physician to patient $k$ in scenario $\omega$ \\
\hline
\textbf{Variables} & & & \\
\hline
$a_{kp\omega} \in \{1,0\}$ & & & 1 if patient $k$ is assigned to physician $p$ in scenario $\omega$ \\
$u_{k\omega} \in \{1,0\}$ & & & 1 if patient $k$ is rejected in scenario $\omega$ \\
\hline
\end{tabular}
\end{table}

\resizebox{14cm}{!}{
\centering
\makebox[\textwidth]{\begin{minipage}{\dimexpr\textwidth-4\fboxsep-8\fboxrule\relax}

\begin{equation}
\textbf{\textsc{MIP}}_{\textbf{SPPA}} 
:= \min \ \frac{1}{|\Omega|}
\left[
\sum_{\omega \in \Omega} 
\sum_{k \in \mathcal{K}_\omega} c^{\text{rej}}_{r_{k\omega}} u_{k\omega}
+
\sum_{\omega \in \Omega}
\sum_{k \in \mathcal{K}_\omega} \sum_{p \in \mathcal{P}_{k\omega}} 
c^{\text{pref}}_{r_{k\omega}} a_{kp\omega}
\right]
\label{eq:obj-offline}
\end{equation}

\begin{align}
\textbf{s.t.} \quad
& \sum_{p \in \mathcal{P}_{k\omega}} a_{kp\omega} + u_{k\omega} = 1 
&& \forall \omega \in \Omega,\ \forall k \in \mathcal{K}_\omega 
\label{eq:assign-offline}
\\[6pt]
& \sum_{k \in \mathcal{K}_\omega : p \in \mathcal{P}_{k\omega}} a_{kp\omega } \le L_p 
&& \forall \omega \in \Omega,\ \forall p \in \mathcal{P}
\label{eq:cap-offline}
\\[6pt]
& \sum_{k \in \mathcal{K}_\omega : p \in \mathcal{P}_{k\omega}} 
t_{k\omega}\,a_{kp\omega} \le T  
&& \forall \omega \in \Omega,\ \forall p \in \mathcal{P}
\label{eq:overtime-def}
\\[6pt]
& a_{1p\omega} = a_{1p\omega'}  
&& \forall \omega \neq \omega'
\label{eq:nonanticipate-def}
\\[6pt]
& u_{1\omega} = u_{1\omega'}  
&& \forall \omega \neq \omega'
\label{eq:nonanticipate-def2}
\\[6pt]
& a_{kp\omega} \in \{0,1\} 
&& \forall \omega \in \Omega,\ \forall k \in \mathcal{K}_\omega,\ \forall p \in \mathcal{P}
\label{eq:binary-x}
\\[6pt]
& u_{k\omega} \in \{0,1\} 
&& \forall \omega \in \Omega,\ \forall k \in \mathcal{K}_\omega
\label{eq:binary-u}
\end{align}

\end{minipage}}
}

\vspace{1cm}

The objective \eqref{eq:obj-offline} and constraints
\eqref{eq:assign-offline}--\eqref{eq:overtime-def} mirror those of
$\textbf{\textsc{MIP}}_{\textbf{PPA}}$.
Constraints \eqref{eq:nonanticipate-def} and \eqref{eq:nonanticipate-def2}
enforce non-anticipativity of the first-stage decisions across scenarios.

\section{Additional details on Instance Generation}
\subsection{Dirichlet-based eligibility construction}\label{app:dirichlet}

Eligible physicians are generated using a patient-level preference model based on the Dirichlet distribution. Let $\mathcal{P}=\{1,\dots,P\}$ denote the physician set. For each arriving patient, a probability vector
\[
\mathbf{w}=(w_1,\dots,w_P), \qquad w_p \ge 0,\ \sum_{p\in\mathcal{P}} w_p=1,
\]
is specified as a global baseline over physicians. In the experiments with $P=4$, we use $\mathbf{w}=[0.4,0.3,0.15,0.15]$. A concentration parameter $\alpha>0$ controls how strongly patient preferences concentrate around $\mathbf{w}$. We form the Dirichlet parameter vector
\[
\boldsymbol{\alpha}=\alpha\,\mathbf{w},
\]
and sample a patient-specific taste vector
\[
\mathbf{z}=(z_1,\dots,z_P)\sim \mathrm{Dirichlet}(\boldsymbol{\alpha}).
\]
Each component $z_p$ represents the relative affinity of the current patient for physician $p$, with $\sum_{p} z_p=1$. Larger values of $\alpha$ produce tastes that are closer to the baseline $\mathbf{w}$, whereas smaller values yield more dispersed and heterogeneous tastes across patients. In our implementation, $\alpha=25$ is used when the baseline weights are enabled, and $\alpha=1$ is used when all physicians are treated uniformly (that is, $w_p=1/P$ for all $p$).

Given $\mathbf{z}$, the eligible set size is drawn as
\[
K \sim \mathcal{U}\{k_{\min},\dots,k_{\max}\},
\]
and the eligible physician set for the patient is defined by selecting the $K$ physicians with the largest taste values:
\[
\mathcal{P}_k = \operatorname{TopK}(\mathbf{z},K).
\]
Equivalently, we construct an eligibility indicator $E_p\in\{0,1\}$ for each physician $p$, where $E_p=1$ if and only if $p\in\mathcal{P}_k$. A safety check enforces that at least one physician is eligible, which is guaranteed by construction for $K\ge 1$.

After constructing $\mathcal{P}_k$, the preferred physician is selected from within the eligible set using the taste vector restricted to $\mathcal{P}_k$. Specifically, the preferred physician is sampled with probabilities proportional to $\{z_p : p\in\mathcal{P}_k\}$, so that physicians with higher taste are more likely to be preferred while maintaining feasibility with respect to eligibility.

\subsection{Scenario Regeneration via Bayesian Posterior Sampling.}
\label{app:bayesian_scenario}
At each decision epoch $k$, when the policy must assign the arriving 
patient with score $s_k$, we regenerate a plausible future cohort as 
follows. Since arrival scores follow priority-specific Beta 
distributions---$\mathrm{Beta}(3,1)$ for priority-1 and 
$\mathrm{Beta}(1,1)$ for priority-2 patients---the score $s_k$ serves 
as a natural observation threshold: exactly $n_k^{(j)}$ patients of 
priority $j \in \{1,2\}$ have arrived with scores below $s_k$ by epoch 
$k$. We exploit this structure to form a Bayesian posterior over the 
unknown total pool sizes $N_1$ and $N_2$. Specifically, the number of 
priority-$j$ arrivals up to threshold $s_k$ follows a Binomial 
likelihood with success probability
\begin{equation}
    p_j \;=\; \mathcal{B}(s_k;\, a_j,\, b_j),
    \label{eq:cdf_threshold}
\end{equation}
where $\mathcal{B}(s;\, a, b)$ denotes the CDF of the 
$\mathrm{Beta}(a, b)$ distribution evaluated at $s$, 
with $(a_1, b_1) = (3, 1)$ for priority-1 and $(a_2, b_2) = (1, 1)$ 
for priority-2 patients. Combining the Normal prior 
$N_j \sim \mathcal{N}(\mu_j, \sigma_j^2)$, with $(\mu_1, \sigma_1) = (25, 5)$ 
and $(\mu_2, \sigma_2) = (75, 12)$, with the Binomial likelihood yields 
the unnormalized discrete posterior
\begin{equation}
    \mathbb{P}\!\left(N_j = n \mid n_k^{(j)}\right) 
    \;\propto\; 
    \binom{n}{n_k^{(j)}} 
    \, p_j^{\,n_k^{(j)}} 
    \,(1-p_j)^{n - n_k^{(j)}}
    \cdot
    \phi\!\left(\frac{n - \mu_j}{\sigma_j}\right),
    \qquad n \geq n_k^{(j)},
    \label{eq:posterior}
\end{equation}
where $\phi(\cdot)$ is the standard normal density. We draw 
$\tilde{N}_j$ from this posterior and generate 
$\tilde{N}_j - n_k^{(j)}$ future patients of priority $j$, whose 
arrival scores are sampled from $\mathrm{Beta}(a_j, b_j)$ truncated 
below at $s_k$, preserving the correct conditional ordering of future 
arrivals. For each regenerated patient, appointment duration, eligible  physician set, and non-preferred penalty $c^{\text{pref}}_{r_k}$ are 
sampled independently from their respective distributions as described 
in \Cref{data}. This procedure ensures that each regenerated  scenario is statistically consistent with both the prior demand 
distribution and the partial information revealed by the $k$ arrivals 
observed so far.

\section{Results}
We present some results on additional experiments.
\subsection{Aggregated Deterministic Expert}  The aggregated deterministic expert \cite{gawas2024ildagger}, \cite{greif2024combinatorial} is motivated by settings in which solving a full multi-scenario stochastic optimisation problem is computationally expensive or unstable under tight time limits. Instead of jointly optimising across scenarios, this approach solves multiple deterministic instances independently—each corresponding to a sampled future scenario—and then aggregates the resulting actions. This decomposition reduces solver complexity, enables parallelisation, and provides robustness by smoothing over scenario-specific noise. Aggregated deterministic experts therefore offer a practical compromise between purely deterministic and fully stochastic experts, especially when high-quality stochastic solutions are difficult to obtain within the available computational budget.

\subsubsection{Learning without interaction}
For a fixed number of scenarios, the deterministic expert, with $F_e(\cdot)$ set to $\pi^{*f}$, can be used to solve each scenario instance independently, after which the resulting solutions are aggregated to produce a single target action. Instance generation follows the same procedure as for the two-stage stochastic expert. Under this strategy, aggregation effectively serves as a heuristic approximation of a two-stage stochastic problem. Relative to the base configuration of \Cref{gdagger}, the primary modification is that the number of expert actions returned by $F_j(\cdot)$ is equal to the number of scenarios considered. Like the two-stage stochastic expert, we consider 5, 15, and 30 scenarios.

For aggregation, we convert the optimal actions into frequency vectors and use these as training targets for the ML model. This yields a smoother, continuous target representation, as opposed to the binary action targets used with the other expert configurations. Across all scenario counts, performance in terms of total cost remains broadly similar, indicating that increasing the number of scenarios offers limited additional benefit. Compared with policies learned from the two-stage stochastic expert, the aggregated deterministic policies perform significantly worse overall.  The training times are similar to the two-stage stochastic expert.
\begin{table}[!ht]
\centering
\caption{Average performance metrics for Aggregated Deterministic Expert}
\label{tab:res4}
\resizebox{14cm}{!}{
\begin{tabular}{|c|c|c|c|c|c|c|}
\hline
\textbf{Learnt Policy}                         & \textbf{Scenarios} & \textbf{Average Cost} &  \textbf{\begin{tabular}[c]{@{}c@{}}Priority 1 \\ Rejected\%\end{tabular}} &  \textbf{\begin{tabular}[c]{@{}c@{}}Priority 2 \\ Rejected\%\end{tabular}} &  \textbf{\begin{tabular}[c]{@{}c@{}}Undesirable \\ Assignments\%\end{tabular}}& \textbf{Train Time (hours)} \\ \hline
\multirow{3}{*}{\textbf{Deterministic ($\pi^{*d}$)}} & 5                  & 1948.4               & 11.4                        & 30.7                        & 28.0    &               3.9                                                 \\ \cline{2-7} 
                                        & 15                 & 1946.9               &    11.5                  &      30.7                   &       27.7     &        11.1         \\ \cline{2-7} 
                                        & 30                 &  1942.2                &    11.2                    &    31.0                    &     27.6              &    23.3        \\ \hline
\end{tabular}}
\end{table}

\subsubsection{Learning with interaction} 
\begin{figure}[!ht]
    \centering
    \caption{Aggregated Deterministic Expert}
         \label{fig:epi_adetI}
        \includegraphics[scale=0.25]{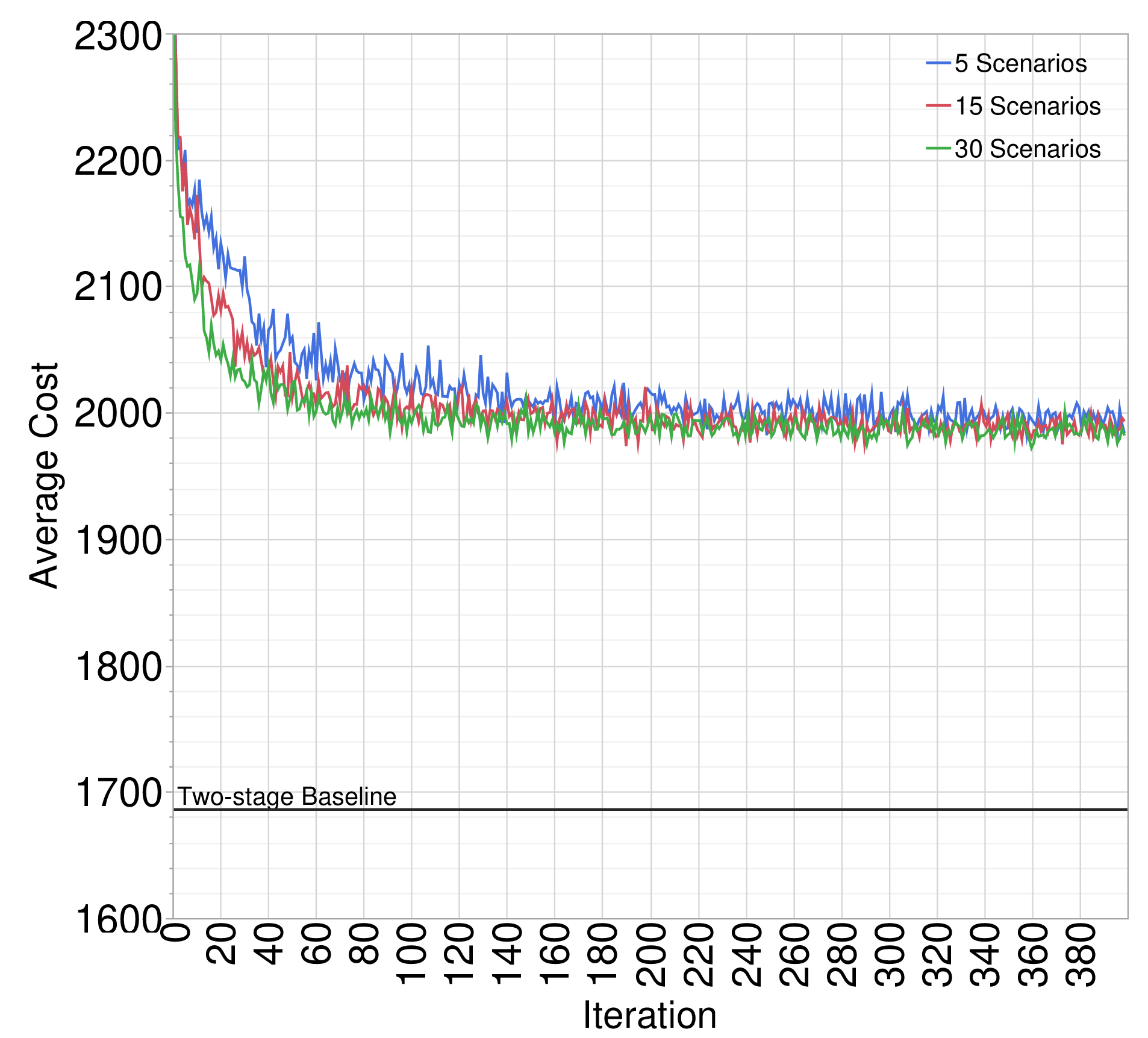}
    \end{figure}

-  \Cref{fig:epi_adetI} illustrates the performance of learned policies obtained using different numbers of aggregated scenarios, while \Cref{tab:res7} reports the performance of the best-performing policies. No significant performance differences are observed across the scenario counts considered. 

    \begin{table}[!ht]
\centering
\caption{Average performance metrics for the best model using the Aggregated Deterministic Expert with interaction}
\label{tab:res7}
\resizebox{14cm}{!}{
\begin{tabular}{|c|c|c|c|c|c|c|c|c|}
\hline
\textbf{Expert}                              & \textbf{Scenarios} & \textbf{\begin{tabular}[c]{@{}c@{}}Num. of Episodes \\ to best model\end{tabular} } & \textbf{Cost} & \textbf{\% Improvement} &  \textbf{\begin{tabular}[c]{@{}c@{}}Priority 1 \\ Rejected\end{tabular}} &  \textbf{\begin{tabular}[c]{@{}c@{}}Priority 2 \\ Rejected\end{tabular}} &  \textbf{\begin{tabular}[c]{@{}c@{}}Undesirable \\ Assignments\end{tabular}} & \textbf{Train Time(hours)} \\ \hline
\multirow{3}{*}{{\textbf{\begin{tabular}[c]{@{}c@{}}Aggregated\\ Deterministic\end{tabular}}}} & 5                  &           3960          & 1934.8   & 0.7      & 11.0                         & 31.0                        &    27.6       &        19.0                \\ \cline{2-9} 
                                             & 15                 &    1900                 &    1937.2   &0.5        &       11.3                       &       30.8                       &    27.6               &    25.2            \\ \cline{2-9} 
                                             & 30                 &    360      & 1939.1    & 0.2            &      11.4         &  30.8                            &       27.6    &    39.5                               \\ \hline
\end{tabular}}
\end{table}

Overall, the learned policies marginally outperform those trained using the full-information and deterministic experts. However, they do not achieve the same level of performance as policies learned from the two-stage stochastic expert. In addition, increasing the number of aggregated scenarios to 15 or 30 does not yield meaningful performance improvements relative to the isolated learning results reported in \Cref{tab:res4}.

\subsubsection{Inference}
Overall, the aggregated deterministic expert provides an alternative to the two-stage stochastic expert, but for the PPA, it does not yield superior policies. Moreover, aggregation requires repeated expert calls, significantly increasing training time. Empirical results show that, for the same number of scenarios, \Cref{gdagger} with an aggregated deterministic expert incurs a higher computational cost than the two-stage stochastic expert. That said, as demonstrated in \cite{gawas2024ildagger}, aggregated deterministic experts can outperform two-stage stochastic experts in settings where the latter are difficult to solve to high quality within practical time limits.

%% file: arxiv.bbl
\begin{thebibliography}{134}
\providecommand{\natexlab}[1]{#1}
\providecommand{\url}[1]{\texttt{#1}}
\providecommand{\urlprefix}{URL }

\bibitem[{Abbasi et~al.(2020)}]{abbasi2020predicting}
Abbasi B, Babaei T, Hosseinifard Z, Smith-Miles K, Dehghani M (2020)
  Predicting solutions of large-scale optimization problems via machine
  learning: A case study in blood supply chain management. \emph{Computers \&
  Operations Research} 119:104941.

\bibitem[{Alvarez et~al.(2016)}]{alvarez2016online}
Alvarez AM, Wehenkel L, Louveaux Q (2016) Online learning for strong branching
  approximation in branch-and-bound.

\bibitem[{Alvarez et~al.(2017)}]{alvarez2017machine}
Alvarez AM, Louveaux Q, Wehenkel L (2017) A machine learning-based
  approximation of strong branching. \emph{INFORMS Journal on Computing}
  29(1):185--195.

\bibitem[{Balachandran(1976)}]{balachandran1976integer}
Balachandran V (1976) An integer generalized transportation model for optimal
  job assignment in computer networks. \emph{Operations Research} 24(4):742--
  759.

\bibitem[{Balcan et~al.(2018)}]{balcan2018learning}
Balcan MF, Dick T, Sandholm T, Vitercik E (2018) Learning to branch. In
  \emph{International conference on machine learning}, 344--353.

\bibitem[{Barnhart et~al.(1998)}]{barnhart1998branch}
Barnhart C, Johnson EL, Nemhauser GL, Savelsbergh MW, Vance PH (1998) Branch-
  and-price: Column generation for solving huge integer programs.
  \emph{Operations research} 46(3):316--329.


\bibitem[{Baty et~al.(2024)}]{baty2024combinatorial}
Baty L, Jungel K, Klein PS, Parmentier A, Schiffer M (2024) Combinatorial
  optimization-enriched machine learning to solve the dynamic vehicle routing
  problem with time windows. \emph{Transportation Science}

\bibitem[{Bengio et~al.(2021)}]{bengio2021machine}
Bengio Y, Lodi A, Prouvost A (2021) Machine learning for combinatorial
  optimization: a methodological tour d’horizon. \emph{European Journal of
  Operational Research} 290(2):405--421.

\bibitem[{Berthold et~al.(n.d.)}]{berthold2206learning}
Berthold T, Francobaldi M, Hendel G (n.d.) Learning to use local cuts.(2022).
  \emph{arXiv preprint arXiv:2206.11618}

\bibitem[{Berthold and Hendel(2021)}]{berthold2021learning}
Berthold T, Hendel G (2021) Learning to scale mixed-integer programs. In
  \emph{Proceedings of the AAAI Conference on Artificial Intelligence}, 3661--
  3668.
\bibitem[{Bishop(2006)}]{bishop2006pattern}
Bishop CM (2006)
\emph{Pattern Recognition and Machine Learning}
(Springer).



\bibitem[{Bonami et~al.(2018)}]{bonami2018learning}
Bonami P, Lodi A, Zarpellon G (2018) Learning a classification of mixed-
  integer quadratic programming problems. In \emph{International conference on
  the integration of constraint programming, artificial intelligence, and
  operations research}, 595--604.


\bibitem[{Cai et~al.(2024)}]{cai2024multi}
Cai J, Huang T, Dilkina B (2024) Multi-task Representation Learning for Mixed
  Integer Linear Programming. \emph{arXiv preprint arXiv:2412.14409}

\bibitem[{Ceschia and Schaerf(2011)}]{ceschia2011local}
Ceschia S, Schaerf A (2011) Local search and lower bounds for the patient
  admission scheduling problem. \emph{Computers \& Operations Research}
  38(10):1452--1463.

\bibitem[{Chen et~al.(2023)}]{chen2023two}
Chen W, Khir R, Van Hentenryck P (2023) Two-stage learning for the flexible
  job shop scheduling problem. \emph{arXiv preprint arXiv:2301.09703}

\bibitem[{Chu et~al.(2023)}]{chu2023data}
Chu H, Zhang W, Bai P, Chen Y (2023) Data-driven optimisation for last-mile
  delivery. \emph{Complex \& Intelligent Systems} 9(3):2271--2284.

\bibitem[{Demirovic et~al.(2020)}]{demirovic2020dynamic}
Demirovic E, Stuckey PJ, Guns T, Bailey J, Leckie C, Ramamohanarao K, Chan J,
  others (2020) Dynamic Programming for Predict+ Optimise. In \emph{AAAI},
  1444--1451.

\bibitem[{Detassis et~al.(2021)}]{detassis2021teaching}
Detassis F, Lombardi M, Milano M (2021) Teaching the old dog new tricks:
  Supervised learning with constraints. In \emph{Proceedings of the AAAI
  Conference on Artificial Intelligence}, 3742--3749.

\bibitem[{Deza and Khalil(2023)}]{deza2023machine}
Deza A, Khalil EB (2023) Machine learning for cutting planes in integer
  programming: A survey. \emph{arXiv preprint arXiv:2302.09166}

\bibitem[{Ding et~al.(2020)}]{ding2020accelerating}
Ding JY, Zhang C, Shen L, Li S, Wang B, Xu Y, Song L (2020) Accelerating
  primal solution findings for mixed integer programs based on solution
  prediction. In \emph{Proceedings of the AAAI Conference on Artificial
  Intelligence}, 1452--1459.

\bibitem[{Donti et~al.(2017)}]{donti2017task}
Donti P, Amos B, Kolter JZ (2017) Task-based end-to-end model learning in
  stochastic optimization. \emph{Advances in neural information processing
  systems} 30.

\bibitem[{Elmachtoub and Grigas(2022)}]{elmachtoub2022smart}
Elmachtoub AN, Grigas P (2022) Smart “predict, then optimize”.
  \emph{Management Science} 68(1):9--26.


\bibitem[{Fischetti and Fraccaro(2019)}]{fischetti2019machine}
Fischetti M, Fraccaro M (2019) Machine learning meets mathematical
  optimisation to predict the optimal production of offshore wind parks.
  \emph{Computers \& Operations Research} 106:289--297.


\bibitem[{Fitzpatrick et~al.(2021)}]{fitzpatrick2021learning}
Fitzpatrick J, Ajwani D, Carroll P (2021) Learning to sparsify travelling
  salesman problem instances. In \emph{International Conference on Integration
  of Constraint Programming, Artificial Intelligence, and Operations
  Research}, 410--426.

\bibitem[{Furian et~al.(2021)}]{furian2021machine}
Furian N, O’sullivan M, Walker C, {\c{C}}ela E (2021) A machine learning-based
  branch and price algorithm for a sampled vehicle routing problem. \emph{Or
  Spectrum} 43(3):693--732.

\bibitem[{Gasse et~al.(2019)}]{gasse2019exact}
Gasse M, Ch{\'e}telat D, Ferroni N, Charlin L, Lodi A (2019) Exact
  combinatorial optimisation with graph convolutional neural networks.
  \emph{Advances in neural information processing systems} 32.

\bibitem[{Gawas et~al.(2025)}]{gawas2024ildagger}
Gawas P, Legrain A, Rousseau LM (2025) An Imitation-Based Learning Approach
  Using DAgger for the Casual Employee Call Timing Problem. In \emph{Learning
  and Intelligent Optimisation}, 153--168.


\bibitem[{Gerbaux et~al.(2025)}]{gerbaux2025machine}
Gerbaux J, Desaulniers G, Cappart Q (2025) A machine-learning-based column
  generation heuristic for electric bus scheduling. \emph{Computers \&
  Operations Research} 173:106848.

\bibitem[{Grassia et~al.(2019)}]{grassia2019learning}
Grassia M, Lauri J, Dutta S, Ajwani D (2019) Learning multi-stage
  sparsification for maximum clique enumeration. \emph{arXiv preprint
  arXiv:1910.00517}

\bibitem[{Greif et~al.(2024)}]{greif2024combinatorial}
Greif T, Bouvier L, Flath CM, Parmentier A, Rohmer SU, Vidal T (2024)
  Combinatorial Optimization and Machine Learning for Dynamic Inventory
  Routing. \emph{arXiv preprint arXiv:2402.04463}

\bibitem[{Guaje et~al.(2024)}]{guaje2024machine}
Guaje O, Deza A, Kazachkov AM, Khalil EB (2024) Machine learning for
  optimization-based separation: the case of mixed-integer rounding cuts.
  \emph{arXiv preprint arXiv:2408.08449}

\bibitem[{Gupta et~al.(2022)}]{gupta2022lookback}
Gupta P, Khalil EB, Chet{\'e}lat D, Gasse M, Bengio Y, Lodi A, Kumar MP (2022)
  Lookback for learning to branch. \emph{arXiv preprint arXiv:2206.14987}

\bibitem[{Han et~al.(2023)}]{han2023gnn}
Han Q, Yang L, Chen Q, Zhou X, Zhang D, Wang A, Sun R, Luo X (2023) A gnn-
  guided predict-and-search framework for mixed-integer linear programming.
  \emph{arXiv preprint arXiv:2302.05636}
\bibitem[{Hastie et~al.(2009)}]{hastie2009elements}
Hastie T, Tibshirani R, Friedman J (2009)
\emph{The Elements of Statistical Learning: Data Mining, Inference, and Prediction}
(Springer).


\bibitem[{He et~al.(2014)}]{he2014learning}
He H, Daum{\'e} III H, Eisner J (2014) Learning to search in branch and bound
  algorithms. \emph{Advances in neural information processing systems} 27.

\bibitem[{Hoque et~al.(2021)}]{hoque2021lazydagger}
Hoque R, Balakrishna A, Putterman C, Luo M, Brown DS, Seita D, Thananjeyan B,
  Novoseller E, Goldberg K (2021) Lazydagger: Reducing context switching in
  interactive imitation learning. In \emph{2021 IEEE 17th international
  conference on automation science and engineering (case)}, 502--509.

\bibitem[{Hottung et~al.(2020)}]{hottung2020deep}
Hottung A, Tanaka S, Tierney K (2020) Deep learning assisted heuristic tree
  search for the container pre-marshalling problem. \emph{Computers \&
  Operations Research} 113:104781.

  \bibitem[{Hu et~al.(2023)}]{hu2023predict+}
Hu X, Lee JCH, Lee JHM (2023)
Predict+ optimize for packing and covering LPs with unknown parameters in constraints.
\emph{Proceedings of the AAAI Conference on Artificial Intelligence} 37(4):3987--3995.


\bibitem[{Huang et~al.(2022)}]{huang2022learning}
Huang Z, Wang K, Liu F, Zhen HL, Zhang W, Yuan M, Hao J, Yu Y, Wang J (2022)
  Learning to select cuts for efficient mixed-integer programming.
  \emph{Pattern Recognition} 123:108353.

\bibitem[{Huang et~al.(2023)}]{huang2023searching}
Huang T, Ferber AM, Tian Y, Dilkina B, Steiner B (2023) Searching large
  neighbourhoods for integer linear programs with contrastive learning. In
  \emph{International conference on machine learning}, 13869--13890.

\bibitem[{Hussein et~al.(2017)}]{hussein2017imitation}
Hussein A, Gaber MM, Elyan E, Jayne C (2017) Imitation learning: A survey of
  learning methods. \emph{ACM Computing Surveys (CSUR)} 50(2):1--35.

\bibitem[{H{\i}z{\i}r et~al.(2025)}]{hizir2025large}
H{\i}z{\i}r AE, Barnhart C, Vaze V (2025) Large-scale airline crew recovery
  using mixed-integer optimization and supervised machine learning.
  \emph{Transportation Science}

\bibitem[{Joshi et~al.(2019)}]{joshi2019efficient}
Joshi CK, Laurent T, Bresson X (2019) An efficient graph convolutional network
  technique for the travelling salesman problem. \emph{arXiv preprint
  arXiv:1906.01227}

\bibitem[{Julien et~al.(2024)}]{julien2024machine}
Julien E, Postek K, Birbil \\ (2024) Machine learning for k-adaptability in
  two-stage robust optimisation. \emph{INFORMS Journal on Computing}

\bibitem[{Jungel et~al.(2025)}]{jungel2025learning}
Jungel K, Parmentier A, Schiffer M, Vidal T (2025) Learning-based online
  optimization for autonomous mobility-on-demand fleet control. \emph{INFORMS
  Journal on Computing}

\bibitem[{Kaempfer and Wolf(2018)}]{kaempfer2018learning}
Kaempfer Y, Wolf L (2018) Learning the multiple traveling salesmen problem
  with permutation invariant pooling networks. \emph{arXiv preprint
  arXiv:1803.09621}

\bibitem[{Khalil et~al.(2016)}]{khalil2016learning}
Khalil E, Le Bodic P, Song L, Nemhauser G, Dilkina B (2016) Learning to branch
  in mixed integer programming. In \emph{Proceedings of the AAAI conference on
  artificial intelligence}.

\bibitem[{Khalil et~al.(2017)}]{khalil2017learning}
Khalil EB, Dilkina B, Nemhauser GL, Ahmed S, Shao Y (2017) Learning to Run
  Heuristics in Tree Search. In \emph{Ijcai}, 659--666.

\bibitem[{Khalil et~al.(2022)}]{khalil2022mip}
Khalil EB, Morris C, Lodi A (2022) Mip-gnn: A data-driven framework for
  guiding combinatorial solvers. In \emph{Proceedings of the AAAI Conference
  on Artificial Intelligence}, 10219--10227.

\bibitem[{Kim et~al.(2021)}]{kim2021demodice}
Kim GH, Seo S, Lee J, Jeon W, Hwang H, Yang H, Kim KE (2021) Demodice: Offline
  imitation learning with supplementary imperfect demonstrations. In
  \emph{International Conference on Learning Representations}.


\bibitem[{Kotary et~al.(2021)}]{kotary2021end}
Kotary J, Fioretto F, Van Hentenryck P, Wilder B (2021) End-to-End Constrained
  Optimization Learning: A Survey. In \emph{Proceedings of the Thirtieth
  International Joint Conference on                Artificial Intelligence,
  {IJCAI-21}}, 4475--4482.

\bibitem[{Kotary et~al.(2024)}]{kotary2024learning}
Kotary J, Di Vito V, Christopher J, Van Hentenryck P, Fioretto F (2024)
  Learning Joint Models of Prediction and Optimisation. In \emph{ECAI 2024},
  2476--2483 (IOS Press).

\bibitem[{Kraul et~al.(2023)}]{kraul2023machine}
Kraul S, Seizinger M, Brunner JO (2023) Machine learning--supported prediction
  of dual variables for the cutting stock problem with an application in
  stabilized column generation. \emph{INFORMS Journal on Computing} 35(3):692
  --709.

\bibitem[{Kruber et~al.(2017)}]{kruber2017learning}
Kruber M, L{\"u}bbecke ME, Parmentier A (2017) Learning when to use a
  decomposition. In \emph{International conference on AI and OR techniques in
  constraint programming for combinatorial optimisation problems}, 202--210.

\bibitem[{La et~al.(2024)}]{larocca2024combining}
La Rocca CR, Cordeau JF, Frejinger E (2024) Combining supervised learning and
  local search for the multicommodity capacitated fixed-charge network design
  problem. \emph{Transportation Research Part E: Logistics and Transportation
  Review} 192:103805.

\bibitem[{Labassi et~al.(2022)}]{labassi2022learning}
Labassi AG, Ch{\'e}telat D, Lodi A (2022) Learning to compare nodes in branch
  and bound with graph neural networks. \emph{Advances in neural information
  processing systems} 35:32000--32010.

\bibitem[{Larsen et~al.(2018)}]{larsen2018predicting}
Larsen E, Lachapelle S, Bengio Y, Frejinger E, Lacoste-Julien S, Lodi A (2018)
  Predicting solution summaries to integer linear programs under imperfect
  information with machine learning. \emph{arXiv preprint arXiv:1807.11876}
  2(4):14.

\bibitem[{Lauri and Dutta(2019)}]{lauri2019fine}
Lauri J, Dutta S (2019) Fine-grained search space classification for hard
  enumeration variants of subset problems. In \emph{Proceedings of the AAAI
  Conference on Artificial Intelligence}, 2314--2321.

\bibitem[{Li et~al.(2018)}]{li2018combinatorial}
Li Z, Chen Q, Koltun V (2018) Combinatorial optimization with graph
  convolutional networks and guided tree search. \emph{Advances in neural
  information processing systems} 31.


\bibitem[{Li et~al.(2025)}]{li2025learning}
Li S, Ouyang W, Ma Y, Wu C (2025) Learning-guided rolling horizon optimization
  for long-horizon flexible job-shop scheduling. \emph{arXiv preprint
  arXiv:2502.15791}

\bibitem[{Lin et~al.(2022)}]{lin2022learning}
Lin J, Zhu J, Wang H, Zhang T (2022) Learning to branch with tree-aware
  branching transformers. \emph{Knowledge-Based Systems} 252:109455.


\bibitem[{Liu et~al.(2022)}]{liu2022learning}
Liu D, Fischetti M, Lodi A (2022) Learning to search in local branching. In
  \emph{Proceedings of the AAAI Conference on Artificial Intelligence}, 3796--
  3803.

\bibitem[{Lodi et~al.(2020)}]{lodi2020learning}
Lodi A, Mossina L, Rachelson E (2020) Learning to handle parameter
  perturbations in combinatorial optimization: an application to facility
  location. \emph{EURO Journal on Transportation and Logistics} 9(4):100023.

\bibitem[{Lu and Kumar(2020)}]{lu-2020}
Lu J, Kumar MP (2020) {Neural network branching for neural network
  verification}. \emph{International Conference on Learning Representations}

\bibitem[{Mandi et~al.(2024)}]{mandi2024decision}
Mandi J, Kotary J, Berden S, Mulamba M, Bucarey V, Guns T, Fioretto F (2024)
  Decision-focused learning: Foundations, state of the art, benchmark and
  future opportunities. \emph{Journal of Artificial Intelligence Research}
  80:1623--1701.

\bibitem[{Mandi et~al.(2025)}]{mandi2025feasibility}
Mandi J, Defresne M, Berden S, Guns T (2025)
Feasibility-aware decision-focused learning for predicting parameters in the constraints.
\emph{arXiv preprint} arXiv:2510.04951.


\bibitem[{Matsuoka et~al.(2019)}]{matsuoka2019machine}
Matsuoka Y, Nishi T, Tiemey K (2019) Machine learning approach for
  identification of objective function in production scheduling problems. In
  \emph{2019 IEEE 15th international conference on automation science and
  engineering (CASE)}, 679--684.

\bibitem[{Menda et~al.(2019)}]{menda2019ensembledagger}
Menda K, Driggs-Campbell K, Kochenderfer MJ (2019) Ensembledagger: A bayesian
  approach to safe imitation learning. In \emph{2019 IEEE/RSJ International
  Conference on Intelligent Robots and Systems (IROS)}, 5041--5048.

\bibitem[{Morabit et~al.(2021)}]{morabit2021machine}
Morabit M, Desaulniers G, Lodi A (2021) Machine-learning-based column
  selection for column generation. \emph{Transportation Science} 55(4):815--
  831.

\bibitem[{Morabit et~al.(2022)}]{morabit2022machine}
Morabit M, Desaulniers G, Lodi A (2022) Machine-learning-based arc selection
  for constrained shortest path problems in column generation. INFORMS J Optim
  5 (2).


\bibitem[{Mossina et~al.(2019)}]{mossina2019multi}
Mossina L, Rachelson E, Delahaye D (2019) Multi-label Classification for the
  Generation of Sub-problems in Time-constrained Combinatorial Optimization.
  In \emph{ICORES 2019, 8th International Conference on Operations Research
  and Enterprise Systems}, pp--133.


\bibitem[{Nair et~al.(2020)}]{Nair2020SolvingMI}
Nair V, Bartunov S, Gimeno F, Glehn IV, Lichocki P, Lobov I, O'Donoghue B,
  Sonnerat N, Tjandraatmadja C, Wang P, Addanki R, Hapuarachchi T, Keck T,
  Keeling J, Kohli P, Ktena I, Li Y, Vinyals O, Zwols Y (2020) Solving Mixed
  Integer Programs Using Neural Networks. \emph{ArXiv} abs/2012.13349.

\bibitem[{Ng et~al.(2000)}]{ng2000algorithms}
Ng AY, Russell S, others (2000) Algorithms for inverse reinforcement learning.
  In \emph{Icml}, 2.

\bibitem[{Niroumandrad et~al.(2024)}]{niroumandrad2024learning}
Niroumandrad N, Lahrichi N, Lodi A (2024) Learning tabu search algorithms: A
  scheduling application. \emph{Computers \& Operations Research} 170:106751.

\bibitem[{Nowozin et~al.(2011)}]{nowozin2011structured}
Nowozin S, Lampert CH, others (2011) Structured learning and prediction in
  computer vision. \emph{Foundations and Trends{\textregistered} in Computer
  Graphics and Vision} 6(3--4):185--365.

\bibitem[{Ojha et~al.(2023)}]{ojha2023optimization}
Ojha R, Chen W, Zhang H, Khir R, Erera A, Van Hentenryck P (2023)
  Optimization-based learning for dynamic load planning in trucking service
  networks. \emph{arXiv preprint arXiv:2307.04050}

\bibitem[{Parmentier and t'Kindt(2021)}]{parmentier2021learning}
Parmentier A, t'Kindt V (2021) Learning to solve the single machine scheduling
  problem with release times and sum of completion times. \emph{arXiv preprint
  arXiv:2101.01082}

\bibitem[{Paul et~al.(2025)}]{paul2025data}
Paul A, Levin MW, Waller ST, Rey D (2025) Data-driven optimization for drone
  delivery service planning with online demand. \emph{Transportation Research
  Part E: Logistics and Transportation Review} 198:104095.

\bibitem[{Paulus et~al.(2022)}]{paulus2022learning}
Paulus MB, Zarpellon G, Krause A, Charlin L, Maddison C (2022) Learning to cut
  by looking ahead: Cutting plane selection via imitation learning. In
  \emph{International conference on machine learning}, 17584--17600.

\bibitem[{Pereira et~al.(2022)}]{pereira2022learning}
Pereira P, Courtade E, Aloise D, Quesnel F, Soumis F, Yaakoubi Y (2022)
  Learning to branch for the crew pairing problem. \emph{Les Cahiers du GERAD
  ISSN} 711:2440.

\bibitem[{Pham et~al.(2023)}]{pham2023prediction}
Pham TS, Legrain A, De Causmaecker P, Rousseau LM (2023) A prediction-based
  approach for online dynamic appointment scheduling: A case study in
  radiotherapy treatment. \emph{INFORMS Journal on Computing} 35(4):844--868.

\bibitem[{Pomerleau(1988)}]{pomerleau1988alvinn}
Pomerleau DA (1988) Alvinn: An autonomous land vehicle in a neural network.
  \emph{Advances in neural information processing systems} 1.

\bibitem[{Powell(2007)}]{powell2007approximate}
Powell WB (2007) \emph{Approximate Dynamic Programming: Solving the curses of
  dimensionality} (John Wiley \& Sons).

\bibitem[{Powell(2022)}]{powell2022reinforcement}
Powell WB (2022) \emph{Reinforcement Learning and Stochastic Optimisation: A
  Unified Framework for Sequential Decisions} (John Wiley \& Sons).

\bibitem[{Puterman(2014)}]{puterman2014markov}
Puterman ML (2014) \emph{Markov decision processes: discrete stochastic
  dynamic programming} (John Wiley \& Sons).

\bibitem[{Qi et~al.(2023)}]{qi2023practical}
Qi M, Shi Y, Qi Y, Ma C, Yuan R, Wu D, Shen ZJ (2023) A practical end-to-end
  inventory management model with deep learning. \emph{Management Science}
  69(2):759--773.

\bibitem[{Rautenstrau{\ss} and Schiffer(2025)}]{rautenstrauss2025optimization}
Rautenstrau{\ss} M, Schiffer M (2025) Optimisation-Augmented Machine Learning
  for Vehicle Operations in Emergency Medical Services. \emph{arXiv preprint
  arXiv:2503.11848}

\bibitem[{Raykar et~al.(2009)}]{raykar2009supervised}
Raykar VC, Yu S, Zhao LH, Jerebko A, Florin C, Valadez GH, Bogoni L, Moy L
  (2009) Supervised learning from multiple experts: whom to trust when
  everyone lies a bit. In \emph{Proceedings of the 26th Annual international
  conference on machine learning}, 889--896.

\bibitem[{Raykar et~al.(2010)}]{raykar2010learning}
Raykar VC, Yu S, Zhao LH, Valadez GH, Florin C, Bogoni L, Moy L (2010)
  Learning from crowds. \emph{Journal of machine learning research} 11(4).

\bibitem[{Ross et~al.(2011)}]{ross2011reduction}
Ross S, Gordon G, Bagnell D (2011) A reduction of imitation learning and
  structured prediction to no-regret online learning. In \emph{Proceedings of
  the fourteenth international conference on artificial intelligence and
  statistics}, 627--635.

\bibitem[{Russell(1998)}]{russell1998learning}
Russell S (1998) Learning agents for uncertain environments. In
  \emph{Proceedings of the eleventh annual conference on Computational
  learning theory}, 101--103.

\bibitem[{Sadana et~al.(2025)}]{sadana2025survey}
Sadana U, Chenreddy A, Delage E, Forel A, Frejinger E, Vidal T (2025) A survey
  of contextual optimization methods for decision-making under uncertainty.
  \emph{European Journal of Operational Research} 320(2):271--289.

\bibitem[{Sang et~al.(2022)}]{sang2022electricity}
Sang L, Xu Y, Long H, Hu Q, Sun H (2022) Electricity price prediction for
  energy storage system arbitrage: A decision-focused approach. \emph{IEEE
  Transactions on Smart Grid} 13(4):2822--2832.

\bibitem[{Sasaki and Yamashina(2020)}]{sasaki2020behavioral}
Sasaki F, Yamashina R (2020) Behavioural cloning from noisy demonstrations. In
  \emph{International Conference on Learning Representations}.

\bibitem[{Scavuzzo et~al.(2024)}]{scavuzzo2024machine}
Scavuzzo L, Aardal K, Lodi A, Yorke-Smith N (2024) Machine learning augmented
  branch and bound for mixed integer linear programming. \emph{Mathematical
  Programming} 1--44.

\bibitem[{Sekhari et~al.(2024)}]{sekhari2024selective}
Sekhari A, Sridharan K, Sun W, Wu R (2024) Selective sampling and imitation
  learning via online regression. \emph{Advances in Neural Information
  Processing Systems} 36.

\bibitem[{Selsam et~al.(2018)}]{selsam2018learning}
Selsam D, Lamm M, B{\"u}nz B, Liang P, de Moura L, Dill DL (2018) Learning a
  SAT solver from single-bit supervision. \emph{arXiv preprint
  arXiv:1802.03685}

\bibitem[{Shen et~al.(2022)}]{shen2022enhancing}
Shen Y, Sun Y, Li X, Eberhard A, Ernst A (2022) Enhancing column generation by
  a machine-learning-based pricing heuristic for graph coloring. In
  \emph{Proceedings of the AAAI conference on artificial intelligence}, 9926--
  9934.

\bibitem[{Sinclair et~al.(2023)}]{sinclair2023hindsight}
Sinclair SR, Frujeri FV, Cheng CA, Marshall L, Barbalho HDO, Li J, Neville J,
  Menache I, Swaminathan A (2023) Hindsight learning for mdps with exogenous
  inputs. In \emph{International Conference on Machine Learning}, 31877--
  31914.

\bibitem[{Song et~al.(2020)}]{song2020general}
Song J, Yue Y, Dilkina B, others (2020) A general large neighbourhood search
  framework for solving integer linear programs. \emph{Advances in Neural
  Information Processing Systems} 33:20012--20023.

\bibitem[{Sonnerat et~al.(2021)}]{sonnerat2021learning}
Sonnerat N, Wang P, Ktena I, Bartunov S, Nair V (2021) Learning a large
  neighbourhood search algorithm for mixed integer programs. \emph{arXiv
  preprint arXiv:2107.10201}

\bibitem[{Spieckermann et~al.(2025)}]{spieckermann2025reduce}
Spieckermann C, Minner S, Schiffer M (2025) Reduce-then-Optimise for the
  Fixed-Charge Transportation Problem. \emph{Transportation Science} 59(3):540
  --564.

\bibitem[{Sun et~al.(2021)}]{sun2021generalization}
Sun Y, Ernst A, Li X, Weiner J (2021) Generalisation of machine learning for
  problem reduction: a case study on travelling salesman problems. \emph{Or
  Spectrum} 43(3):607--633.

\bibitem[{Sun et~al.(2022)}]{sun2022learning}
Sun Y, Ernst AT, Li X, Weiner J (2022) Learning to generate columns with
  application to vertex coloring. In \emph{The Eleventh International
  Conference on Learning Representations}.

\bibitem[{Sun et~al.(2023)}]{sun2023mega}
Sun X, Yang S, Mangharam R (2023) Mega-dagger: Imitation learning with
  multiple imperfect experts. \emph{arXiv preprint arXiv:2303.00638}

\bibitem[{Sun et~al.(2024)}]{sun2024enhancing}
Sun Y, Nguyen S, Thiruvady D, Li X, Ernst AT, Aickelin U (2024) Enhancing
  constraint programming via supervised learning for job shop scheduling.
  \emph{Knowledge-Based Systems} 293:111698.

\bibitem[{Sutton et~al.(1998)}]{sutton1998reinforcement}
Sutton RS, Barto AG, others (1998) \emph{Reinforcement learning: An
  introduction} (MIT press Cambridge).

\bibitem[{Tahir et~al.(2021)}]{tahir2021improved}
Tahir A, Quesnel F, Desaulniers G, El Hallaoui I, Yaakoubi Y (2021) An
  improved integral column generation algorithm using machine learning for
  aircrew pairing. \emph{Transportation Science} 55(6):1411--1429.

\bibitem[{Tayebi et~al.(2022)}]{tayebi2022learning}
Tayebi D, Ray S, Ajwani D (2022) Learning to prune instances of k-median and
  related problems. In \emph{2022 Proceedings of the Symposium on Algorithm
  Engineering and Experiments (ALENEX)}, 184--194.

\bibitem[{Tian et~al.(2023)}]{tian2023smart}
Tian X, Yan R, Liu Y, Wang S (2023) A smart predict-then-optimise method for
  targeted and cost-effective maritime transportation. \emph{Transportation
  Research Part B: Methodological} 172:32--52.

\bibitem[{Vinyals et~al.(2015)}]{vinyals2015pointer}
Vinyals O, Fortunato M, Jaitly N (2015) Pointer networks. \emph{Advances in
  neural information processing systems} 28.

\bibitem[{V{\'a}clav{\'\i}k et~al.(2018)}]{vaclavik2018accelerating}
V{\'a}clav{\'\i}k R, Nov{\'a}k A, {\v{S}}{\r{u}}cha P, Hanz{\'a}lek Z (2018)
  Accelerating the branch-and-price algorithm using machine learning.
  \emph{European Journal of Operational Research} 271(3):1055--1069.

\bibitem[{Wang et~al.(2021)}]{wang2021learning}
Wang Y, Xu C, Du B, Lee H (2021) Learning to weight imperfect demonstrations.
  In \emph{International Conference on Machine Learning}, 10961--10970.

\bibitem[{Wang et~al.(2023)}]{wang2023imitation}
Wang Y, Dong M, Du B, Xu C (2023) Imitation learning from purified
  demonstration. \emph{arXiv preprint arXiv:2310.07143}

\bibitem[{Wu et~al.(2019)}]{wu2019imitation}
Wu YH, Charoenphakdee N, Bao H, Tangkaratt V, Sugiyama M (2019) Imitation
  learning from imperfect demonstration. In \emph{International Conference on
  Machine Learning}, 6818--6827.

\bibitem[{Xavier et~al.(2021)}]{xavier2021learning}
Xavier \S, Qiu F, Ahmed S (2021) Learning to solve large-scale security-
  constrained unit commitment problems. \emph{INFORMS Journal on Computing}
  33(2):739--756.

\bibitem[{Xu et~al.(2018)}]{xu2018towards}
Xu H, Koenig S, Kumar TS (2018) Towards effective deep learning for constraint
  satisfaction problems. In \emph{International Conference on Principles and
  Practice of Constraint Programming}, 588--597.

\bibitem[{Xu et~al.(2022)}]{xu2022discriminator}
Xu H, Zhan X, Yin H, Qin H (2022) Discriminator-weighted offline imitation
  learning from suboptimal demonstrations. In \emph{International Conference
  on Machine Learning}, 24725--24742.

\bibitem[{Yaakoubi et~al.(2020)}]{yaakoubi2020flight}
Yaakoubi Y, Soumis F, Lacoste-Julien S (2020) Flight-connection prediction for
  airline crew scheduling to construct initial clusters for OR optimizer.
  \emph{arXiv preprint arXiv:2009.12501}

\bibitem[{Yang et~al.(2023)}]{yang2023learning}
Yang Z, Ishay A, Lee J (2023) Learning to solve constraint satisfaction
  problems with recurrent transformer. \emph{arXiv preprint arXiv:2307.04895}

\bibitem[{Yilmaz and Yorke-Smith(2021)}]{yilmaz2021study}
Yilmaz K, Yorke-Smith N (2021) A study of learning search approximation in
  mixed integer branch and bound: Node selection in scip. \emph{Ai} 2(2):150--
  178.

\bibitem[{Zare et~al.(2024)}]{zare2024survey}
Zare M, Kebria PM, Khosravi A, Nahavandi S (2024) A survey of imitation
  learning: Algorithms, recent developments, and challenges. \emph{IEEE
  Transactions on Cybernetics}

\bibitem[{Zarpellon et~al.(2021)}]{zarpellon2021parameterizing}
Zarpellon G, Jo J, Lodi A, Bengio Y (2021) Parameterising branch-and-bound
  search trees to learn branching policies. In \emph{Proceedings of the AAAI
  Conference on Artificial Intelligence}, 3931--3939.

\bibitem[{Zhang and Cho(2016)}]{zhang2016query}
Zhang J, Cho K (2016) Query-efficient imitation learning for end-to-end
  autonomous driving. \emph{arXiv preprint arXiv:1605.06450}

\bibitem[{Zheng et~al.(2022)}]{zheng2022imitation}
Zheng B, Verma S, Zhou J, Tsang IW, Chen F (2022) Imitation learning:
  Progress, taxonomies and challenges. \emph{IEEE Transactions on Neural
  Networks and Learning Systems} 35(5):6322--6337.

\bibitem[{{\"O}zar{\i}k et~al.(2024)}]{ozarik2024machine}
{\"O}zar{\i}k SS, da Costa P, Florio AM (2024) Machine learning for data-
  driven last-mile delivery optimisation. \emph{Transportation Science}
  58(1):27--44.

\end{thebibliography}
